\documentclass[acmtog,screen,nonacm]{acmart}
\AtBeginDocument{%
  }

\usepackage[utf8]{inputenc}
\usepackage[T1]{fontenc}
\usepackage{booktabs}
\usepackage{colortbl}
\usepackage{multirow}
\usepackage{xspace}
\usepackage{enumitem}

\ifx\figurename\undefined \def\figurename{Figure}\fi
\renewcommand{\figurename}{Fig.}
\newcommand{\Sect}[1]{Sec.~\ref{#1}}
\newcommand{\Fig}[1]{Fig.~\ref{#1}}
\newcommand{\Tbl}[1]{Tbl.~\ref{#1}}
\newcommand{\Eqn}[1]{Eqn.~\ref{#1}}
\newcommand{\Apx}[1]{Supp.~\ref{#1}}
\newcommand{\proj}{\textsc{ControlGS}\xspace}
\newcommand{\mode}[1]{\underline{\textsc{#1}}\xspace}


\begin{document}

\title{\proj: Conditioning Neural Gaussians for Downstream-Processing--Aware XR Rendering}

\author{Weikai Lin}
\orcid{0000-0003-3537-4857}
\affiliation{%
  \institution{University of Rochester}
  \city{Rochester}
  \country{USA}}
\email{wlin33@ur.rochester.edu}

\author{Junjie Zhao}
\orcid{0009-0005-6251-8483}
\affiliation{%
  \institution{University of Rochester}
  \city{Rochester}
  \country{USA}}
\email{jzhao58@u.rochester.edu}

\author{Carl Marshall}
\orcid{0009-0001-7288-5341}
\affiliation{%
  \institution{Reality Labs, Meta}
  \city{Redmond}
  \country{USA}}
\email{csmarshall@meta.com}

\author{Sushant Kondguli}
\orcid{0000-0002-7295-4626}
\affiliation{%
  \institution{Reality Labs, Meta}
  \city{Rochester}
  \country{USA}}
\email{sushant.kondguli@gmail.com}

\author{Yuhao Zhu}
\orcid{0000-0002-2802-0578}
\affiliation{%
  \institution{University of Rochester}
  \city{Rochester}
  \country{USA}}
\email{yzhu@rochester.edu}

\renewcommand{\shortauthors}{Lin et al.}

\begin{abstract}
Extended Reality (XR) users do not directly perceive the output of a rendering engine.
Instead, rendered images pass through a post-processing pipeline and the physical display-optics path before reaching the eye.
Critically, the exact downstream processing can vary significantly at run time, influenced by, for instance, camera pose and display power budget.
Traditional 3DGS methods either implicitly assume that this downstream pipeline preserves image quality or cannot adapt to downstream processing changes.
To bridge this gap, we present \proj, an XR Gaussian rendering pipeline that optimizes end-to-end visual quality.
\proj models and integrates the entire downstream processing, between the rendering output and the human eye, into the optimization objective.
To adapt to downstream processing at run time, \proj dynamically generates Gaussian primitives conditioned upon the downstream processing parameters.
Experiments show that \proj consistently improves end-to-end post-optics XR quality across different neural Gaussian backbones and datasets, with minimal  overhead.
Code is available at \url{https://horizon-lab.org/controlgs/}.

\end{abstract}

\begin{teaserfigure}
  \centering
  \includegraphics[width=0.94\textwidth]{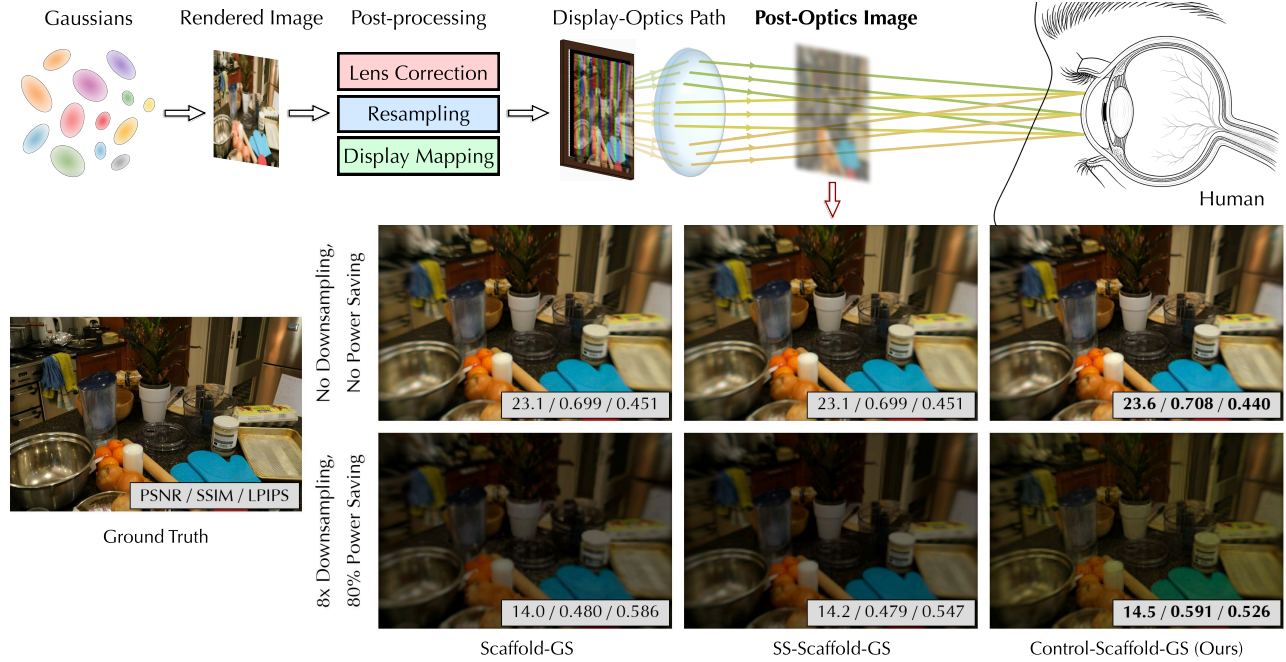}
  \caption{
    \textbf{ControlGS enables downstream-processing--aware XR rendering.}
  Top: a Gaussian-rendered image passes through post-processing and the display-optics path before forming the post-optics virtual image.
  Bottom: compared with Scaffold-GS and its supersampling-enhanced variant (SS-Scaffold-GS), ControlGS produces higher-quality virtual images under both native resolution with no power saving and $8{\times}$ downsampling with $80\%$ power saving.
  }
  \label{fig:teaser}
\end{teaserfigure}

\maketitle
\begingroup
\renewcommand{\thefootnote}{}
\footnotetext{All artifacts were accessed/generated by the University of Rochester.}
\endgroup

\section{Introduction}
\label{sec:intro}

Extended Reality (XR) is poised to become the next-generation human-computer interface.
3D Gaussian Splatting (3DGS)~\citep{kerbl20233dgs} is emerging as a promising rendering paradigm for XR due to its ability to produce photorealistic results at real-time frame rates while massively simplifying content authoring.

Directly using a Gaussian renderer in XR, however, creates a gap between the rendered image and the final image perceived by humans.
This is because there is a sequence of signal processing, both computationally and in hardware, that takes place to transform the rendered image to photons entering the user's eye.
Computationally, a rendered image must pass through three main processing stages, i.e., lens correction, anti-aliasing resampling, and display mapping under a power constraint~\citep{duinkharjav2022color,chen2024pea,lin2026lowpowar}.
These computational stages, followed by the physical display-optics path, transform the rendered image before it is perceived by humans.

Critically, the exact downstream processing can vary significantly at run time, influenced by, for instance, camera pose and display power budget.
For instance, as the camera/user moves farther away from a scene, the effective display sampling rate of the scene is lower (equivalently, the spatial frequency of the scene on the display plane is higher).
Traditional 3DGS-for-XR methods either implicitly assume that this downstream pipeline preserves image quality and, thus, optimize only for the renderer output~\cite{vrsplatting, lin2024metasapiens, lin2025powergs} or cannot adapt to downstream processing changes.

We introduce \proj, a framework for downstream-processing--aware Gaussian rendering.
We model the entire downstream processing pipeline between the renderer output and the human eyes, and integrate such a model into the 3DGS training objective.
Importantly, we observe that traditional 3DGS is fundamentally ill-suited to adapt to the variability in downstream processing, because its Gaussian primitives are fixed after training.
Instead, our idea is to dynamically generate, from a set of learned latent vectors, the Gaussian primitives at run time.
This generation process is, importantly, conditioned upon various downstream-specific parameters such as display sampling rate and the power budget to enable adaptation.

Specifically, \proj combines a condition-injection module with an MLP.
For each computational stage, we learn a condition vector and a projection layer that embed the downstream parameters as a latent condition, which is injected into the MLP to control the attributes of the Gaussian primitives.
The injected conditions can be individually and independently toggled depending on the exact downstream processing pipeline, broadening its applicability.

We evaluate \proj with diverse MLP-based Gaussian decoders under a typical post-processing pipeline covering lens correction, anti-aliasing resampling, and display mapping.
Experiments show that \proj consistently improves the end-to-end image quality compared to baselines that do not consider downstream processing.
Our contributions are summarized as follows:
\begin{itemize}[leftmargin=1.5em]
\item We formulate XR Gaussian rendering as end-to-end optimization across both Gaussian splatting and post-rendering processing.
\item We propose \proj, a Gaussian splatting method that dynamically generates Gaussian primitives conditioned on downstream processing parameters.
\item \proj supports various XR post-processing modules, achieving consistent end-to-end quality gains.
\end{itemize}

\section{Related Work}
\label{sec:related}

\paragraph{Gaussian Splatting for XR Rendering.}
3DGS~\cite{kerbl20233dgs} represents scenes with explicit anisotropic Gaussians and supports real-time photorealistic rendering.
Neural Gaussian methods~\cite{scaffoldgs,contextgs, compgs, hacpp, ren2024octreegs} further replace explicit per-Gaussian attributes with  features and neural decoders, which generate Gaussian attributes at render time.
Several works improve Gaussian rendering for XR or multi-resolution settings, including anti-aliased Gaussian rendering~\cite{yu2024mipsplatting}, foveated rendering~\cite{lin2024metasapiens,vrsplatting, fan2024fovgs}, and display-power-aware rendering~\cite{lin2025powergs}.
However, these methods  optimize renderer-side quality and ignore quality changes introduced by post-rendering modules.
In contrast, \proj optimizes for the end-to-end XR quality.

\paragraph{XR Post-Rendering Pipeline.}
After rendering, XR systems apply a post-rendering pipeline before the image reaches the user's  eye.
Lens correction pre-warps images to compensate for distortion and chromatic aberrations~\citep{openxr_distortion,pohl2013improved,martschinke2019gaze}.
Display mapping converts rendered colors to display intensities and can reduce display power through color or luminance remapping~\cite{duinkharjav2022color,chen2024pea, lin2026lowpowar}.
Differentiable lens models have also been used for end-to-end camera imaging optimization~\citep{sun2021differentiable,zhang2024end}.
\proj uses differentiable post-rendering modules, enabling an end-to-end loss path from the eye-space image to our condition-injection module.

\paragraph{Neural Gaussians.}
Our method shares similarities and key differences with neural gaussian models, which also use MLPs to dynamically decode Gaussian primitives during rendering~\cite{scaffoldgs, hacpp, ren2024octreegs, compgs, contextgs}.
However, the main objective of neural Gaussian models is to compress the model (through the MLP).
\proj proposes \textit{conditionally} dynamic decoding, where the MLP is conditioned upon downstream processing parameters.
In addition, \proj's use of the MLP is less concerned with compression (although we do inherent the benefits of a small model size), but focuses on adapting the Gaussian primitives to fit the downstream processing.

\section{Problem Formulation}
\label{sec:prelim}

\begin{figure*}[t]
  \centering
  \includegraphics[width=\linewidth]{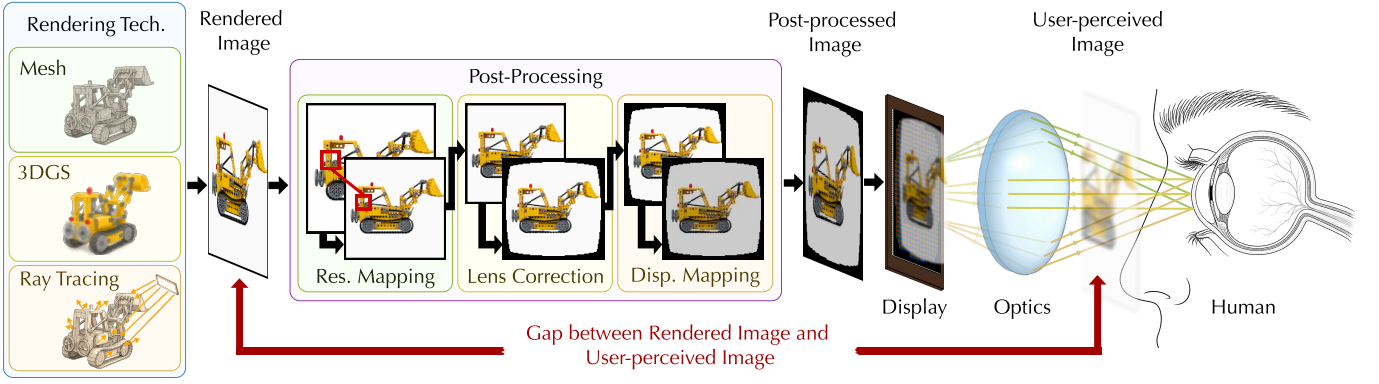}
\caption{
 XR pipeline for Gaussian rendering.
The rendered image from a Gaussian renderer passes through post-processing modules and the physical display-optics path before reaching the user's eye.
Thus, the user perceives the final eye-space image rather than the raw rendered image.
For clarity, the figure places the perceived image on the right side of the optics, although an actual near-eye display forms a virtual image on the opposite side.
}
  \label{fig:bg}
\end{figure*}

\Fig{fig:bg} illustrates the XR pipeline for 3D Gaussian rendering.
A 3D Gaussian renderer first produces a rendered image, which passes through a sequence of computational post-processing stages (\Sect{sec:prelim:post}), followed by the physical display-optics path (\Sect{sec:prelim:physical-path}).
Only after these stages does the image reach the user's eye.
Our rendering objective integrates these two processes (\Sect{sec:prelim:prob}).

\subsection{Post-Processing Pipeline in XR}
\label{sec:prelim:post}
Given a rendered image $I$, an XR system applies a sequence of post-processing modules.
In this paper, we focus on three common modules: lens correction $\mathcal{P}_{\mathrm{lens}}$, anti-aliasing resampling $\mathcal{P}_{\mathrm{res}}$, and display mapping $\mathcal{P}_{\mathrm{disp}}$.
Each stage is parameterized by a corresponding run-time parameter: rendering-time sampling rate $s_{\mathrm{res}}$, lens model $s_{\mathrm{lens}}$, and display power target $s_{\mathrm{disp}}$.
These stages collectively produce the post-processed image $I_{\mathrm{post}}$ given an image $I$ generated by a 3DGS model:
\begin{equation}
  I_{\mathrm{post}}
  = \mathcal{P}_{\mathrm{disp}}\!\left(
      \mathcal{P}_{\mathrm{res}}\!\left(
        \mathcal{P}_{\mathrm{lens}}(I; s_{\mathrm{lens}});
        s_{\mathrm{res}}
      \right);
      s_{\mathrm{disp}}
    \right).
\label{eq:xr-post-pipeline}
\end{equation}

\paragraph{Lens Correction.}
The projection lens introduces spatially varying aberrations (\Eqn{eq:lens-conv}).
Following prior work~\citep{openxr_distortion,pohl2013improved,martschinke2019gaze,hiroi2022neuraldistortion}, we assume that the post-processing pipeline uses a color-dependent warping stage $\mathcal{P}_{\mathrm{lens}}$ to computationally compensate for the lens aberrations (\Eqn{eq:lens-conv}).
As is commonly done, we model $\mathcal{P}_{\mathrm{lens}}$ as a per-channel distortion map between each input pixel $I_{x, y, c}$ and output pixel $I_{x', y', c}$, and the one-to-one mapping between the input pixel location $(x, y, c)$ and output pixel location $(x', y', c)$ is given by the lens model $s_{\mathrm{lens}}$:
\begin{equation}
    \mathcal{P}_{\mathrm{lens}}(I_{x, y, c}; s_{\mathrm{lens}}): I_{x, y, c} \mapsto I_{x', y', c},
    \;\;
    (x', y', c) = s_{\mathrm{lens}}(x, y, c).
\label{eq:lens-module}
\end{equation}

\paragraph{Anti-Aliasing Resampling.}
From an abstract signal-processing perspective, and analogous to traditional surface and volume splatting~\cite{zwicker2001ewa, zwicker2001surface}, a 3DGS model is best thought of as a pre-computed texture stored in the scene space.
The scene sampling rate of this texture is determined by the image resolution and camera poses at the training time.
Rendering the 3DGS model can be thought of as sampling the texture (or re-sampling the scene).

However, the rendering-time sampling rate and training-time sampling rate might differ --- for two main reasons.
First, the display resolution might be different from the resolution of the training images.
Second, the rendering-time camera pose could be closer (upsampling) or farther (downsampling) than the training poses;
similarly, the rendering lens might support zoom and/or focal-length adjustment.
As a result, aliasing could occur.

A typical anti-aliasing strategy is to first super-sample the texture at a rate higher than the display sampling rate, followed by a reconstruction filter to reconstruct the continuous scene signal, followed by a pre-filter that band-limits the scene signal before re-sampling the band-limited signal at the display sampling rate.
The reconstruction filter and the pre-filter are usually combined together and called the \textit{resampling} filter.

In our formulation, $\mathcal{P}_{\mathrm{res}}$ is expressed as:
\begin{equation}
    \mathcal{P}_{\mathrm{res}}(I_S; s_{\mathrm{res}}) : I_S \mapsto  \text{III}_{s_{\mathrm{res}}}\!\left(I_S \otimes h\right),
\label{eq:res-down-signal}
\end{equation}
where $I_S$ is the image rendered by a 3DGS model and lens-corrected at a sampling rate $S$ higher than the display sampling rate $s_{\mathrm{res}}$, $h$ represents the resampling filter, $\otimes$ is convolution, and $\text{III}_{s_{\mathrm{res}}}$ is a Dirac-comb function representing sampling at a rate of $s_{\mathrm{res}}$.

\paragraph{Display Mapping.}
After anti-aliasing resampling, the image resolution matches the target display resolution, but the image values still need to be mapped to the pixel values for the physical display, a step we call display mapping $\mathcal{P}_{\mathrm{disp}}$.
Apart from color space conversion (CSC) and gamut mapping (GM), an important consideration of display mapping is to conserve display power, which has become increasingly important for XR devices~\citep{duinkharjav2022color,chen2024pea}.
\citet{chen2024pea} shows that uniform dimming (UD) is a particularly effective method for saving display power while preserving perceptual quality.
UD applies a single global scaling factor to all image pixels to meet a given display power target.
Under UD, $\mathcal{P}_{\mathrm{disp}}$ is expressed as (omitting CSC and GM):
\begin{equation}
    \mathcal{P}_{\mathrm{disp}}(I; s_{\mathrm{disp}}) : I \mapsto \frac{s_{\mathrm{disp}}}{P_{\mathrm{disp}}(I)} I,
\label{eq:map-module}
\end{equation}
\noindent where $P_{\mathrm{disp}}(\cdot)$ represents the display power of an image, and $s_{\mathrm{disp}}$ represents the target display power and can vary at run time.

\subsection{XR Physical Display-Optics Path}
\label{sec:prelim:physical-path}

After post-processing, $I_{\mathrm{post}}$ is sent to the physical display $\mathcal{D}_{\mathrm{phys}}$ and then transformed by the headset optics $\mathcal{O}_{\mathrm{phys}}$ before it is perceived by the user.
The post-optics virtual image $I_{\mathrm{eye}}$ is
\begin{equation}
    I_{\mathrm{eye}}
    = \mathcal{O}_{\mathrm{phys}}\!\left(
        \mathcal{D}_{\mathrm{phys}}(I_{\mathrm{post}})
      \right).
\label{eq:physical-path}
\end{equation}

\paragraph{Physical Display.}
The display $\mathcal{D}_{\mathrm{phys}}$ reconstructs a continuous optical signal from discrete image pixel values.
This signal reconstruction is modeled by treating each display pixel as a box filter:
\begin{equation}
    \mathcal{D}_{\mathrm{phys}}(I_{\mathrm{post}}): I_{\mathrm{post}} \mapsto I_{\mathrm{disp}}, \qquad I_{\mathrm{disp}} = I_{\mathrm{post}} \otimes b,
  \label{eq:disp-recon}
\end{equation}

\noindent where $I_{\mathrm{disp}}$ is the optical image emitted from the display, and $b$ is a box filter.
The resulting display power is modeled using a common linear model~\citep{duinkharjav2022color,chen2024pea,lin2025powergs}, where the dynamic display power is modeled as a weighted sum of the linear RGB values:
\begin{equation}
  P_{\mathrm{disp}}(I)
  = \sum_{x,y} \big(\alpha I_R(x,y) + \beta I_G(x,y) + \gamma I_B(x,y)\big),
  \label{eq:disp-power}
\end{equation}
where $(\alpha,\beta,\gamma)$ are panel-dependent coefficients.

\paragraph{Projective Lens.}
The projective optics $\mathcal{O}_{\mathrm{phys}}$ projects the displayed image to a larger virtual image at a comfortable viewing distance.
At the same time, the optics introduce aberrations.
The optical propagation can be modeled with a wavelength-dependent shift-variant point-spread function (PSF):

\begin{equation}
\begin{aligned}
\mathcal{O}_{\mathrm{phys}}(I_{\mathrm{disp}})
: I_{\mathrm{disp}} \mapsto I_{\mathrm{eye}},& \\
 I_{\mathrm{eye}}^{\lambda}(x,y)
= \iint \mathrm{PSF}_{\lambda}(x,y;u,v)\,
I_{\mathrm{disp}}^{\lambda}(x-u,y-v)\,du\,dv.&
\end{aligned}
\label{eq:lens-conv}
\end{equation}

\noindent where $\lambda \in \Lambda$ denotes the visible wavelength range, and the PSF depends on both field position and wavelength.
This model captures spatially varying, wavelength-dependent (chromatic) aberrations.

\subsection{Problem Formulation}
\label{sec:prelim:prob}

To bridge the gap between a 3DGS-rendered image and the image that humans see, we optimize $I_{\mathrm{eye}}$ after the optics (\Eqn{eq:physical-path}), rather than the intermediate rendered image.
The end-to-end XR operator $\mathcal{E}_{s}$ includes post-processing, the physical display, and optics:
\begin{equation}
  \mathcal{E}_{s}(I)
  =
  \mathcal{O}_{\mathrm{phys}}\!\left(
    \mathcal{D}_{\mathrm{phys}}\!\left(
      \mathcal{P}_{\mathrm{disp}}\!\left(
        \mathcal{P}_{\mathrm{res}}\!\left(
          \mathcal{P}_{\mathrm{lens}}(I; s_{\mathrm{lens}});
          s_{\mathrm{res}}
        \right);
        s_{\mathrm{disp}}
      \right)
    \right)
  \right).
  \label{eq:method-end-to-end-operator}
\end{equation}
Let $q$ be the camera view and $R_{\theta}$ be a Gaussian renderer with parameters $\theta$.
The objective is:
\begin{equation}
  \min_{\theta}\;
  \mathbb{E}_{q,s}\,
  \mathcal{L}_{\mathrm{img}}\!\left(
    \mathcal{E}_{s}\!\left(R_{\theta}(q)\right),
    I_{\mathrm{gt}}(q)
  \right).
  \label{eq:post-aware-objective}
\end{equation}
Here $R_{\theta}(q)$ is the image rendered by the Gaussian renderer from view $q$, and $I_{\mathrm{gt}}(q)$ is the target image for view $q$.
$\mathcal{L}_{\mathrm{img}}$ is an image loss that compares the difference between two images.

Importantly, the three downstream processing parameters ($s_\mathrm{lens}$, $s_\mathrm{res}$, and $s_\mathrm{disp}$) might vary at the rendering time.
Therefore, the renderer must adapt to the specific run-time states of these parameters in order to maximize the end-to-end visual quality.
\begin{itemize}[leftmargin=2.2em]
    \item As the camera moves farther from the scene, or when an XR application simulates zooming out by reducing the lens focal length, the scene occupies a smaller region on the image plane, thereby reducing the effective sampling rate $s_\mathrm{res}$.
    \item The display power budget and, accordingly, $s_\mathrm{disp}$, might also change depending on the user preference or battery status.
    \item Finally, although the physical lens is fixed for a given XR device, its aberrations are spatially varying (i.e., $s_\mathrm{disp}(\cdot)$ depends on both $x$ and $y$, as seen in \Eqn{eq:lens-module}).
    As the camera pose changes, the same Gaussian primitive may project to different image-plane locations and, thus, experience different aberrations.
\end{itemize}

We provide concrete examples of dynamic sampling-rate changes and view-dependent aberration-state changes in \Apx{sec:appendix:runtime-state-examples}.

\section{\proj}
\label{sec:method}

\begin{figure*}[t]
  \centering
  \includegraphics[width=\linewidth]{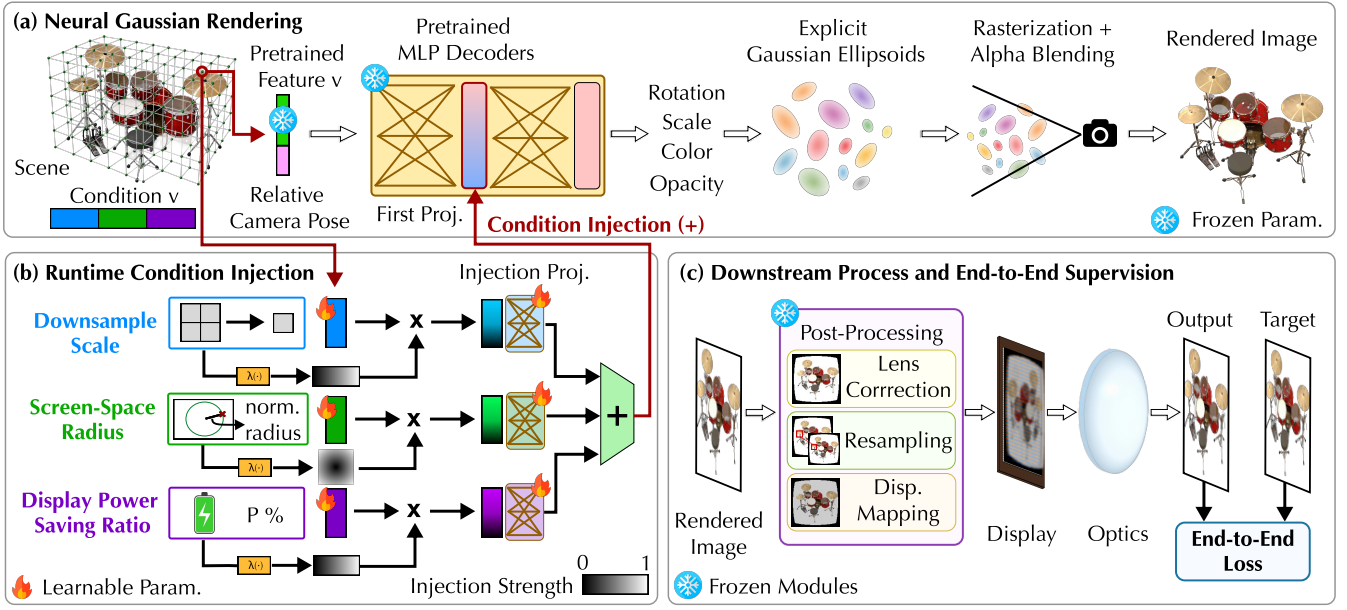}
  \caption{Overview of \proj.
  (a) A neural Gaussian renderer decodes Gaussian attributes from pretrained feature anchors and rasterizes them into an image.
  (b) For each feature and downstream module, \proj learns a condition vector, projects it into the decoder hidden space, and scales it based on injection strength before injection (\Eqn{eq:condition-injection}).
  (c) The conditioned rendering result is passed through fixed XR downstream modules and the plug-in is trained with end-to-end supervision on the final output image (\Eqn{eq:plugin-training-objective}).
  }
  \label{fig:arch}
\end{figure*}

\proj provides a solution to the problem in \Eqn{eq:post-aware-objective}.
\Fig{fig:arch} gives an overview of \proj.
The key idea is to dynamically generate the Gaussian primitives conditioned upon the downstream processing parameters (\Sect{sec:method:neural-gs}).
The conditioning strength is controlled based on the exact parameter magnitude (\Sect{sec:method:strength}).
We discuss how to train \proj in a light-weight fashion (\Sect{sec:method:train}).

\subsection{Conditional Dynamic Gaussians}
\label{sec:method:neural-gs}

\paragraph{Neural Gaussian Backbone.}
The formulation in \Eqn{eq:post-aware-objective} requires the Gaussian renderer to adapt to run-time  XR states.
Explicit 3DGS methods~\cite{kerbl20233dgs, fan2023lightgaussian, yu2024mipsplatting} use fixed Gaussian attributes at run time and cannot meet this requirement.
Therefore, we must dynamically generate the Gaussian primitives based on the downstream processing parameters ($s_\mathrm{lens}$, $s_\mathrm{res}$, and $s_\mathrm{disp}$).
Inspired by the neural Gaussian-family methods~\cite{scaffoldgs}, we use an MLP to generate the Gaussian primitives from learned latent features at rendering time.

As shown in \Fig{fig:arch} (a), we represent the scene with a set of sparse anchor features.
Each feature $v$ has a 3D position $x_v$, an associated feature $\hat f_v$, and a local scale factor $l_v$.
At run time, anchor $v$ emits $k$ Gaussian primitives through learned offsets $\{\delta_m\}_{m=0}^{k-1}$:
\begin{equation}
  \{\mu_0,\ldots,\mu_{k-1}\}
  = x_v + \{{\delta}_0,\ldots,{\delta}_{k-1}\}\, l_v,
  \label{eq:scaffold-offset}
\end{equation}
where $\mu_m$ denotes the center of the $m$-th emitted Gaussian.
The attributes of these Gaussians are decoded by an MLP at render time:
\begin{equation}
  \{o, S, R, c\}
  = F_{\theta}\!\left(\hat f_v, \Delta_{vc}, \tilde d_{vc}\right),
  \label{eq:scaffold-decoder}
\end{equation}
where $F_{\theta}$ is the MLP attribute decoder, and $\{o, S, R, c\}$ are opacity, scale, rotation, and color.
The inputs $\Delta_{vc}$ and $\tilde d_{vc}$ encode the relative position and viewing direction between anchor $v$ and the camera $c$.
Thus, the decoded Gaussian attributes depend on the camera pose.

\paragraph{Conditional Injection.}
While the MLP can dynamically generate the Gaussian primitives, it must do so based on the specific downstream processing parameters.
The MLP decoder in neural Gaussian rendering provides a natural place to inject downstream conditions.
For each post-processing stage $m$, \proj learns per-feature $N$-dimensional conditions $\{\mathcal{C}_{v,m}\}$ and a shared projection $\mathcal{A}_m:\mathbb{R}^{N}\!\to\!\mathbb{R}^{D_1}$, where $D_1$ is the first hidden-layer width.
The projected condition is added to this hidden layer to modulate decoded Gaussian attributes without changing the pretrained model.

Formally, let $\mathrm{FC}_1$ be the first linear layer of the decoder and let $F_{\theta,\mathrm{rest}}$ be the remaining decoder layers.
For anchor feature $\hat f_v$, view inputs $(\Delta_{vc},\tilde d_{vc})$, and active modules $\mathcal{M}$, we decode Gaussian attributes as
\begin{equation}
  \{o,S,R,c\}
  = F_{\theta,\mathrm{rest}}\!\left(
      \mathrm{FC}_1(\hat f_v,\Delta_{vc},\tilde d_{vc})
      + \sum_{m\in\mathcal{M}}\lambda_{v,m}(s_{v,m})\,\mathcal{A}_m(\mathcal{C}_{v,m})
    \right).
\label{eq:condition-injection}
\end{equation}
The decoded attributes are then combined with the pretrained Gaussian positions (\Eqn{eq:scaffold-offset}) for rasterization.
Here $\mathcal{C}_{v,m}$ is the condition vector for feature $v$ and module $m$, while $\mathcal{A}_m$ is the projection layer shared across all features for module $m$.
The function $\lambda_{v,m}(s_{v,m})$ maps the run-time  state of feature $v$ under module $m$ to an injection strength, as defined in \Sect{sec:method:strength}.
As shown in \Fig{fig:arch}(b), different module conditions are summed before injection, making module-specific conditions composable through the same decoder interface.

\subsection{Controlling Run-time Injection Strength}
\label{sec:method:strength}

The learned condition defines an adaptation direction for a post-processing stage.
At the run time, the downstream parameter can change as discussed in \Sect{sec:prelim:prob}.
The decoding therefore also needs to know how strongly this condition should be injected for each Gaussian feature.
We control this with a feature-specific injection strength $\lambda_{v,m}(s_{v,m})\in[0,1]$.
When $\lambda_{v,m}(s_{v,m})=1$, the condition for module $m$ is fully injected for feature $v$.
When $\lambda_{v,m}(s_{v,m})=0$, the condition is removed from \Eqn{eq:condition-injection} for feature $v$.
In this case, the method degenerates to the frozen neural Gaussian backbone.

To compute the strength, each module first converts its run-time  state to a scalar control value $\alpha_{v,m}$ using a module-specific function $f_m(\cdot)$.
We then define two endpoints, $\alpha_m^{\min}$ and $\alpha_m^{\max}$, over the supported run-time  range.
The minimum endpoint maps to $\lambda_{v,m}=0$, and the maximum endpoint maps to $\lambda_{v,m}=1$.
States between them are handled by linear interpolation:
\begin{equation}
  \lambda_{v,m}(s_{v,m})
  =
  \frac{\alpha_{v,m}-\alpha_m^{\min}}{\alpha_m^{\max}-\alpha_m^{\min}}
  =
  \frac{f_m(s_{v,m})-f_m(s_m^{\min})}{f_m(s_m^{\max})-f_m(s_m^{\min})}.
  \label{eq:injection-strength}
\end{equation}

Thus the minimum supported state produces no injection for feature $v$, and the maximum supported state produces full injection.
We choose these endpoints to cover the XR run-time  range specified by the application.
If a run-time  state falls outside this range, we  clamp the interpolated strength to $[0,1]$ before injection.

We now discuss, for each of the three post-processing stages, how to design the run-time  state $s_{v,m}$, its endpoints $s_m^{\min}$ and $s_m^{\max}$, and $f_m(\cdot)$ used to compute injection strength (\Eqn{eq:injection-strength}).

\paragraph{Lens Correction.}
Most projective lenses in XR are radially symmetric about the optical center.
Therefore, the distance to the optical center determines the amount of aberrations (although not the exact PSF) introduced by the lens.
For each Gaussian feature $v$, the run-time  state is its projected screen-space location,
$s_{v,\mathrm{lens}}=(x_v,y_v)$, as defined in \Eqn{eq:lens-module}.
We normalize this location to $p_v=\mathrm{Norm}(x_v,y_v)\in[-1,1]^2$ and use its radius as the control value:
$\alpha_{v,\mathrm{lens}}=f_{\mathrm{lens}}(s_{v,\mathrm{lens}})=\|p_v\|_2$.
Thus, $\alpha_{\mathrm{lens}}^{\min}=0$ at the optical center and $\alpha_{\mathrm{lens}}^{\max}=\sqrt{2}$ at the screen corner.
The injection strength $\lambda_{v, \mathrm{lens}}(s_{v,\mathrm{lens}})$ is computed by \Eqn{eq:module-strengths}.

\paragraph{Anti-Aliasing Resampling.}
The run-time state $s_{\mathrm{res}}$ is the display sampling rate and is shared by all Gaussian features $v$, i.e., $s_{v,\mathrm{res}}=s_{\mathrm{res}}$ for all $v$.
We convert it to a downsampling scale $d_{\mathrm{res}}=S/s_{\mathrm{res}}$, where $S$ is the Gaussian training resolution.
We support $d_{\mathrm{res}}\in[1,8]$: $d_{\mathrm{res}}=1$ is native resolution with no injection, and $d_{\mathrm{res}}=8$ is the strongest downsampling with full injection.
We use $\alpha_{v,\mathrm{res}}=f_{\mathrm{res}}(s_{v,\mathrm{res}})=\log_2 d_{\mathrm{res}}$, so $\alpha_{\mathrm{res}}^{\min}=0$ and $\alpha_{\mathrm{res}}^{\max}=\log_2 8$.
$\lambda_{v, \mathrm{res}}$ is computed by \Eqn{eq:module-strengths}.

\paragraph{Display Mapping.}
The run-time state $s_{\mathrm{disp}}$ is the target display power $P_{\mathrm{target}}$ and is shared by all Gaussian features $v$, i.e., $s_{v,\mathrm{disp}}=s_{\mathrm{disp}}$ for all $v$.
We convert it to a power-saving ratio using the reference display power $P_{\mathrm{ref}}$, which is the display power consumed by the ground truth image:
$\rho=f_{\mathrm{disp}}(s_{\mathrm{disp}})=1-s_{\mathrm{disp}}/P_{\mathrm{ref}}=1-P_{\mathrm{target}}/P_{\mathrm{ref}}$.
We use $\alpha_{v,\mathrm{disp}}=\rho$, with $\alpha_{\mathrm{disp}}^{\min}=0$ and $\alpha_{\mathrm{disp}}^{\max}=0.8$.
Thus, a larger power-saving ratio produces stronger condition injection.
The injection strength $\lambda_{v,\mathrm{map}}$ is computed by \Eqn{eq:module-strengths}.

With these choices, \Eqn{eq:injection-strength} for module $m$ and feature $v$ becomes
\begin{equation}
  \lambda_{v, \mathrm{res}}=\frac{\log_2 d_{\mathrm{res}}}{\log_2 8}, \quad
  \lambda_{v, \mathrm{lens}}=\frac{\|p_v\|_2}{\sqrt{2}}, \quad
  \lambda_{v, \mathrm{disp}}=\frac{\rho}{0.8}.
  \label{eq:module-strengths}
\end{equation}

\subsection{Training and Inference}
\label{sec:method:train}

To minimize training overhead, we use pre-trained neural Gaussian MLPs and train only the run-time injection module.
Let $\theta$ be the fixed pretrained neural Gaussian parameters, and let
$\phi=\{\mathcal{C}_{v,m},\mathcal{A}_m\}_{v,m}$ be the trainable condition-injection parameters.
We denote the conditioned renderer by $R_{\theta,\phi}(q;s)$.
It renders view $q$ while modulating Gaussian attribute decoding with run-time  states $s$ through \Eqn{eq:condition-injection}.
Because the pretrained renderer parameters $\theta$ are fixed, we optimize only the condition-injection parameters $\phi$.
Thus, the formulation in \Eqn{eq:post-aware-objective} becomes the training objective:
\begin{equation}
  \min_{\phi}\;
  \mathbb{E}_{q,s}\,
  \mathcal{L}_{\mathrm{img}}\!\left(
    \mathcal{E}_{s}\!\left(R_{\theta,\phi}(q;s)\right),
    I_{\mathrm{gt}}(q)
  \right).
  \label{eq:plugin-training-objective}
\end{equation}
$\mathcal{E}_{s}$ applies the downstream modules (\Eqn{eq:method-end-to-end-operator}).
$I_{\mathrm{gt}}(q)$ is the ground-truth image for view $q$.
We use the image loss
\begin{equation}
  \mathcal{L}_{\mathrm{img}}
  = \alpha\,\mathcal{L}_{1}
  + \beta\,\mathcal{L}_{\mathrm{SSIM}}
  + \gamma\,\mathcal{L}_{\mathrm{MSE}}
  + \delta\,\mathcal{L}_{\mathrm{LPIPS}},
  \label{eq:image-loss}
\end{equation}
where $\alpha$, $\beta$, $\gamma$, and $\delta$ are scalar loss weights.
We follow 3DGS~\cite{kerbl20233dgs} for $\mathcal{L}_{1}$ and $\mathcal{L}_{\mathrm{SSIM}}$, and add MSE and LPIPS~\cite{zhang2018unreasonable} to improve optimization through the display-optics pipeline.

Inspired by ControlNet~\cite{zhang2023adding}, we zero-initialize the condition vectors so the conditioned renderer starts from the original model.
During training, we sample display sampling rates and target powers from their supported ranges, while aberration states are determined by each feature's projected screen-space position and is naturally sampled by the camera view.

\section{Experiments}
\label{sec:experiments}

\subsection{Experimental Setup}
\label{sec:experiments:setup}

\paragraph{Datasets.}
We evaluate on Mip-NeRF~360~\citep{barron2022mipnerf360} and NeRF Synthetic~\citep{mildenhall2020nerf}.
Mip-NeRF~360 contains real unbounded scenes with background, representing immersive VR rendering.
NeRF Synthetic contains object-centric scenes with only foreground objects, representing AR scenes.

\paragraph{Backbones.}
\proj uses \mode{Scaffold-GS}~\citep{scaffoldgs}, a canonical neural Gaussian splatting, as the main MLP backbone.
To show broad applicability, we also experiment with other common neural Gaussian methods, including \mode{Octree-GS}~\citep{ren2024octreegs}, \mode{HAC{++}}~\citep{hacpp}, \mode{ContextGS}~\citep{contextgs}, and \mode{CompGS}~\citep{compgs}.
All backbones are pretrained and frozen; only the condition-injection parameters are learned (\Sect{sec:method}).

\paragraph{Implementation Details.}
For each downstream module, we set the dimension of per-feature condition $\mathcal{C}_{v,m}$ to $8$.
The full plug-in therefore uses a $24$-dimensional condition for three modules.
We train the plug-in for $10{,}000$ iterations with a learning rate of $3\times10^{-3}$.
For the loss in \Eqn{eq:image-loss}, we set $(\alpha,\beta,\gamma,\delta)$ to $(0.8,0.2,10,0.1)$ for Mip-NeRF~360 and $(0.8,0.2,100,1)$ for NeRF Synthetic.
We use larger MSE and LPIPS weights on NeRF Synthetic to compensate for its smaller residual errors.
Additional details are provided in \Apx{sec:appendix:implementation}.

\paragraph{Downstream Modules.}
For post-processing, we use the box filter for anti-aliasing resampling $\mathcal{P}_{\mathrm{res}}$, pre-computed pixel-wise warping for lens correction $\mathcal{P}_{\mathrm{lens}}$, and uniform dimming (UD)~\cite{chen2024pea} for display mapping $\mathcal{P}_{\mathrm{map}}$.
For the physical display processes, we model the OLED display pixels as box filters and use the an OLED display power model from prior work~\cite{duinkharjav2022color}.
For optics, we use a convex lens design based on Google Cardboard~\cite{googlecardboard}, and generate the spatially-variant PSFs using Zemax.
The implementation details and supporting validation are provided in \Apx{sec:appendix:post-rendering-module-details}.

\paragraph{Metrics.}
We compare the final virtual image from the XR pipeline
(\Eqn{eq:method-end-to-end-operator}) with the ground-truth image
using PSNR~\cite{huynh2008scope}, SSIM~\citep{wang2004ssim}, and LPIPS~\citep{zhang2018unreasonable}.
We evaluate four effective display sampling rates or, equivalently, downsampling factors
$d_{\mathrm{res}}\in\{1,2,4,8\}$ ($d_{\mathrm{res}} = 1$ is equivalent to training-time resolution), and five display-power-saving ratios
$\{0\%,20\%,40\%,60\%,80\%\}$, yielding  $20$ run-time settings.
Note that the downsampling factor here is not intended to model different physical display resolutions.
Rather, it emulates changes in effective sampling rate under camera pose and lens zoom changes, as discussed in \Sect{sec:prelim:prob}.

\begin{figure*}[t]
\centering
\includegraphics[width=\linewidth]{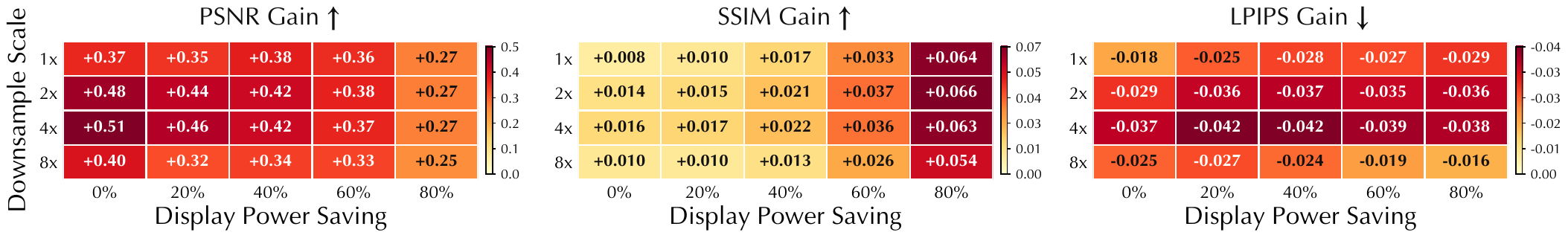}
\caption{
Average quality improvement of ControlGS over SS-Scaffold-GS on Mip-NeRF~360.
Heatmaps show PSNR, SSIM, and LPIPS gains across downsample scale and display-power-saving settings, averaged over all scenes.
ControlGS consistently improves end-to-end quality .
}
\label{fig:supp:runtime-state-heatmaps}
\end{figure*}

\begin{figure*}[t]
\centering
\includegraphics[width=\linewidth]{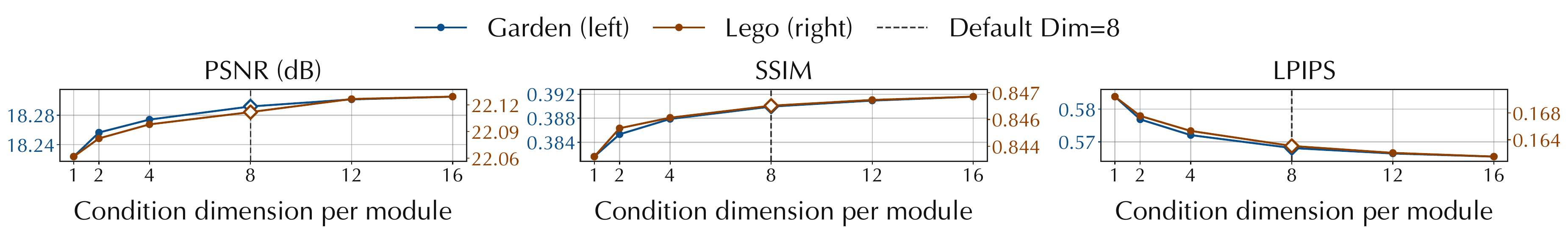}
\caption{Effect of the condition dimension on end-to-end quality.
The x-axis is the total condition dimension across anti-aliasing resampling, lens correction, and display mapping.
The y-axis reports PSNR, SSIM, and LPIPS on the Garden and Lego scenes.}
\label{fig:condition-dimension}
\end{figure*}

\begin{table}[t]
\centering
\caption{
End-to-end quality after the XR post-processing pipeline, averaged over all scenes and $20$ resolution-power run-time  settings.
\proj here uses the MLP in SS-Scaffold-GS; results on other MLPs are in \Apx{sec:appendix:gen}.
}
\label{tab:xrgs-end-to-end-quality-summary}
\definecolor{xrgsGroupShade}{RGB}{235,241,248}
\definecolor{xrgsSSShade}{RGB}{255,249,224}
\definecolor{xrgsControlShade}{RGB}{255,239,224}
\setlength{\tabcolsep}{3.5pt}
\renewcommand{\arraystretch}{1.08}
\resizebox{0.9\columnwidth}{!}{%
\begin{tabular}{lcccccc}
\toprule
Method / Dataset & \multicolumn{3}{c}{{Mip-NeRF~360}} & \multicolumn{3}{c}{{NeRF Synthetic}} \\
\cmidrule(lr){2-4}\cmidrule(lr){5-7}
& PSNR$\uparrow$ & SSIM$\uparrow$ & LPIPS$\downarrow$ & PSNR$\uparrow$ & SSIM$\uparrow$ & LPIPS$\downarrow$ \\
\midrule
\rowcolor{xrgsGroupShade}
\multicolumn{7}{l}{\textbf{\textsc{Run-Time--Fixed Gaussian Splatting}}} \\
3DGS & 18.15 & 0.478 & 0.579 & 21.96 & 0.840 & 0.166 \\
\rowcolor{xrgsSSShade}
SS-3DGS & \textbf{18.93} & \textbf{0.496} & \textbf{0.558} & \textbf{23.36} & \textbf{0.869} & \textbf{0.146} \\
\specialrule{0.25pt}{1.0pt}{1.0pt}
Mip-Splatting & 18.87 & 0.493 & 0.564 & 23.17 & 0.862 & 0.152 \\
\rowcolor{xrgsSSShade}
SS-Mip-Splatting & \textbf{18.93} & \textbf{0.497} & \textbf{0.557} & \textbf{23.37} & \textbf{0.870} & \textbf{0.145} \\
\midrule
\rowcolor{xrgsGroupShade}
\multicolumn{7}{l}{\textbf{\textsc{Neural Gaussian Splatting}}} \\
Scaffold-GS & 18.19 & 0.479 & 0.578 & 21.94 & 0.839 & 0.167 \\
\rowcolor{xrgsSSShade}
SS-Scaffold-GS & 18.96 & 0.496 & 0.558 & 23.35 & 0.869 & 0.147 \\
\rowcolor{xrgsControlShade}
ControlGS (Ours) & \textbf{19.33} & \textbf{0.524} & \textbf{0.528} & \textbf{23.59} & \textbf{0.871} & \textbf{0.137} \\
\bottomrule
\end{tabular}%
}
\end{table}

\paragraph{Baselines.}
We use pretrained neural Gaussian backbones as baselines.
We also include two explicit 3DGS-based baselines with fixed primitives: \mode{3DGS}~\citep{kerbl20233dgs} and \mode{Mip-Splatting}~\citep{yu2024mipsplatting}.
The baselines directly render at the target display sampling rate.
Mip-Splatting has its own set of anti-aliasing methods.
Each baseline also has a variant that uses supersampling (SS) for anti-aliasing, where we always render at the training-time resolution before resampling to the display sampling rate.
We discuss the details in \Apx{sec:appendix:display-resolution-matching}.
For lens correction and display mapping, all baselines use the same modules as \proj.

\subsection{Main Results}
\label{sec:experiments:main}

\paragraph{End-to-End Quality.}
\Tbl{tab:xrgs-end-to-end-quality-summary} compares end-to-end quality averaged over all scenes and all $20$ run-time  settings.
Using \mode{Scaffold-GS} as backbones, \proj achieves the best quality among baselines.
We also enhance each baseline with the same supersampling (SS) strategy used by \proj.
SS improves all baselines, confirming the importance of supersampling.
Compared with these SS-enhanced baselines, \proj still consistently achieves the best end-to-end quality, showing the benefit of adapting to run-time  downstream states through condition injection.
\proj generalizes to other neural Gaussian backbones, with results provided in \Apx{sec:appendix:gen}.
Per-scene results are in \Apx{sec:appendix:per-scene-main-results}.

We further evaluate post-optics image quality using four perceptual metrics: ColorVideoVDP~\cite{mantiuk2024colorvideovdp}, HDR-VDP-3~\cite{mantiuk2023hdr}, HDR-FLIP~\cite{andersson2021visualizing}, and FovVideoVDP~\cite{Mantiuk2021FovVideoVDP}.
These metrics serve as proxies for the human visual system and complement the standard image-quality metrics above.
\Tbl{tab:hvs-aware-metrics} reports results on Mip-NeRF~360.
For all metrics, we use the \texttt{sRGB\_display} setting in HDR-VDP-3 and compute pixels per degree from our optical design. \proj consistently improves over SS-Scaffold-GS across all four metrics.

\begin{table}[t]
  \centering
  \caption{Perceptual quality on Mip-NeRF~360.}
  \label{tab:hvs-aware-metrics}
  \setlength{\tabcolsep}{3pt}
  \resizebox{\linewidth}{!}{%
  \begin{tabular}{lcccc}
  \toprule
  Method & ColorVideoVDP $\uparrow$ & HDR-VDP-3 $\uparrow$ & HDR-FLIP $\downarrow$ & FovVideoVDP $\uparrow$ \\
  \midrule
  SS-Scaffold-GS & 7.179 & 6.971 & 0.455 & 7.001 \\
  ControlGS & \textbf{7.232} & \textbf{7.111} & \textbf{0.418} & \textbf{7.079} \\
  \bottomrule
  \end{tabular}%
  }
\end{table}

\paragraph{Per-Setting Analysis.}
To show that \proj improves quality across run-time  states rather than only on average, we visualize its gains over the SS-enhanced baseline on the full $4{\times}5$ run-time  grid.
Using Scaffold-GS as the backbone, \Fig{fig:supp:runtime-state-heatmaps} reports PSNR, SSIM, and LPIPS gains under all resolution and power-saving settings.
\proj consistently outperforms the SS-enhanced baseline.
Even at $1{\times}$ downsampling and $0\%$ power saving, \proj still improves quality through the lens condition, which adapts Gaussian decoding based on each feature's projected screen-space location.
Additional heatmaps for {NeRF Synthetic} are provided in \Apx{sec:appendix:run-time-state-heatmaps}.

\begin{table}[t]
\centering
\caption{FPS and model storage.}
\vspace{-10pt}
\label{tab:xrgs-run-time -storage-overhead}
\setlength{\tabcolsep}{4.0pt}
\renewcommand{\arraystretch}{1.08}
\resizebox{0.8\columnwidth}{!}{%
\begin{tabular}{lcccc}
\toprule
Method / Dataset & \multicolumn{2}{c}{{Mip-NeRF~360}} & \multicolumn{2}{c}{{NeRF Synthetic}} \\
\cmidrule(lr){2-3}\cmidrule(lr){4-5}
& FPS$\uparrow$ & Storage (MB)$\downarrow$ & FPS$\uparrow$ & Storage (MB)$\downarrow$ \\
\midrule
SS-3DGS & 214.9 & 647.51 & 764.0 & 58.51 \\
SS-Scaffold-GS & 337.9 & 180.53 & 920.5 & 15.13 \\
ControlGS & 268.9 & 235.36 & 702.7 & 19.72 \\
\bottomrule
\end{tabular}%
}
\end{table}

\begin{figure*}[t]
\centering
\includegraphics[width=0.85\linewidth]{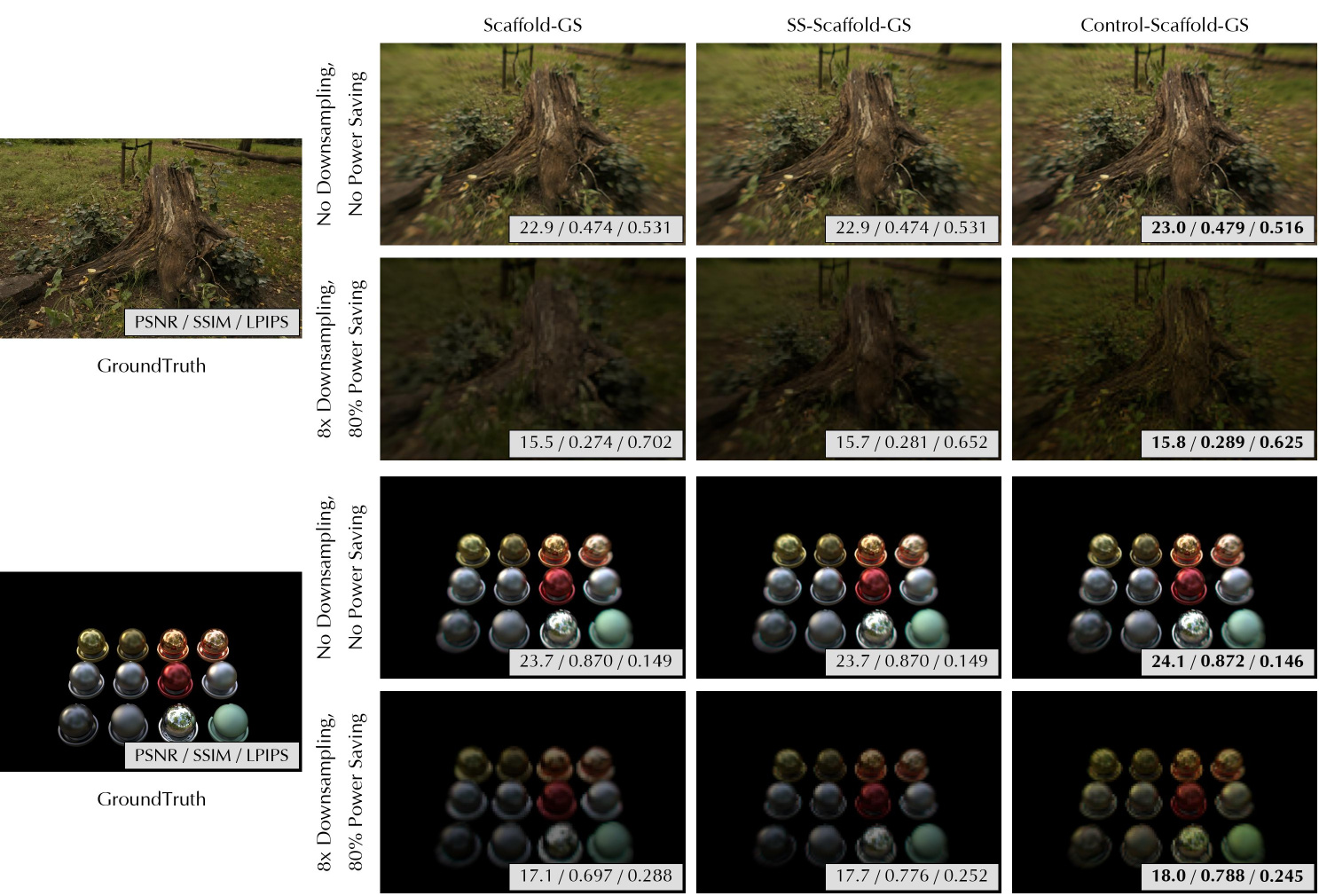}
\caption{
Qualitative end-to-end comparison on two representative scenes.
We compare Scaffold-GS, its supersampled variant, and \proj under no downsampling/no power saving and under $8{\times}$ downsampling with $80\%$ display-power saving.
Numbers report PSNR / SSIM / LPIPS for the shown view; bold indicates the best value among the three methods.
}
\Description{Qualitative comparison of Scaffold-GS, SS-Scaffold-GS, and ControlGS on the Stump and Materials scenes under two XR run-time  settings.}
\label{fig:qualitative-comparison}
\end{figure*}

\paragraph{Qualitative Comparison.}
Beyond objective metrics, \Fig{fig:qualitative-comparison} compares the end-to-end visual results of the original baseline, the SS-enhanced baseline, and \proj after optics simulation.
At the highest resolution without display-power saving, \proj produces sharper peripheral details by compensating for lens-induced blur.
Under aggressive downsampling and display-power saving, \proj adapts to the downsampling ratio to preserve sharper structures while shifting colors toward lower-power yellowish tones for display mapping.
This preserves more details that would otherwise be suppressed by dimming.
Additional qualitative results across more scenes are provided in \Apx{sec:appendix:additional-qualitative}.

\paragraph{Overhead Analysis.}
Efficiency remains important for XR rendering.
We evaluate \proj in a split-rendering setting, where a tethered server or cloud GPU renders Gaussian frames before streaming them to the XR device.
Using Scaffold-GS, \Tbl{tab:xrgs-run-time -storage-overhead} reports averaged render-side FPS and storage on Mip-NeRF~360 and {NeRF Synthetic}.
\proj achieves $268.9$ FPS on Mip-NeRF~360 and $702.7$ FPS on {NeRF Synthetic}, remaining well above typical XR frame-rate requirements.
Compared with SS-Scaffold-GS, \proj increases storage by about $30\%$ due to the conditional injection module, while remaining substantially smaller than the original 3DGS.
We also discussed impact of supersampling in \Apx{sec:appendix:ss-overhead}.

\begin{figure*}[t]
\centering
\includegraphics[width=0.9\linewidth,height=0.94\textheight,keepaspectratio]{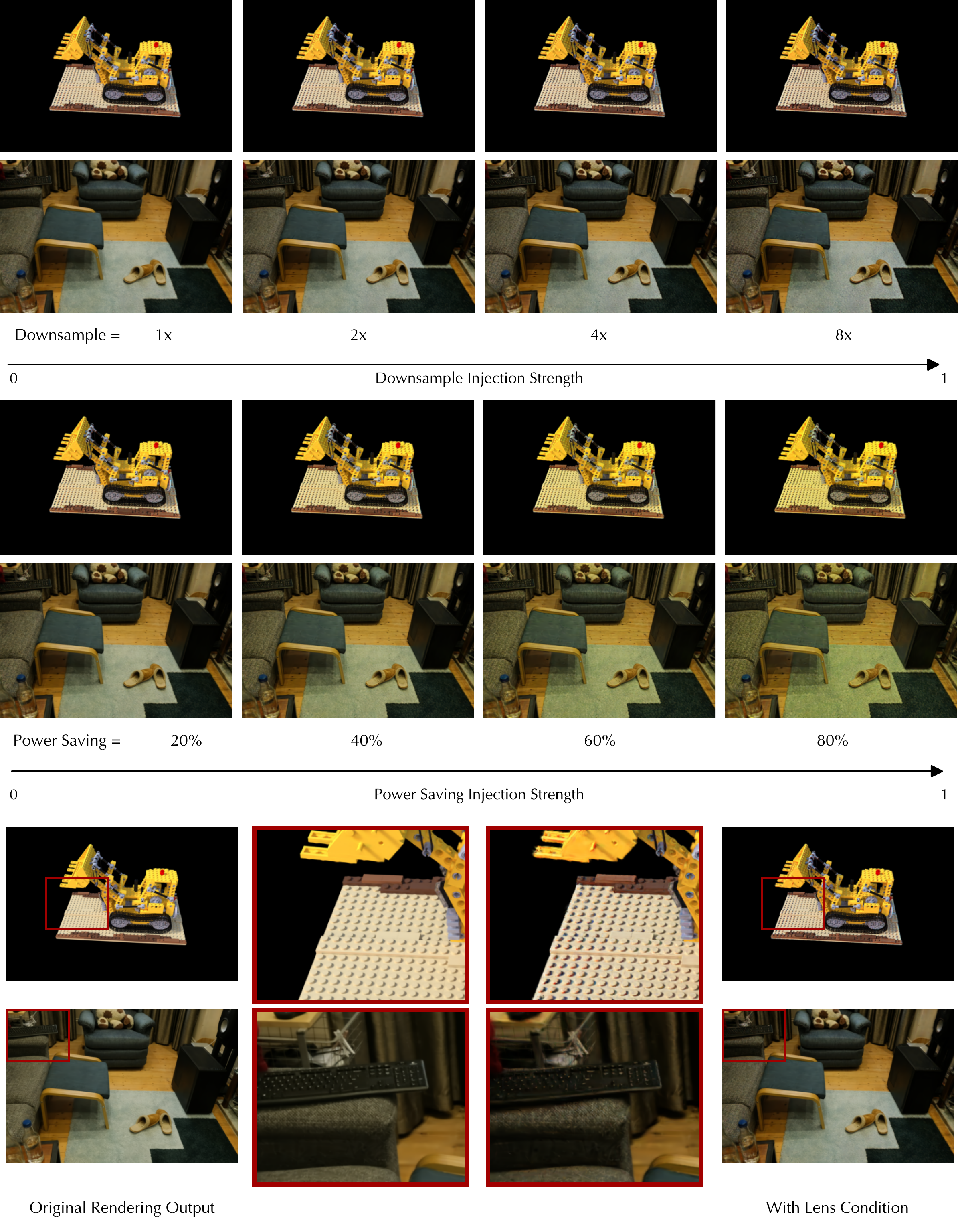}
\caption{
Condition-response visualization on Lego and Room.
We vary one condition at a time and show the raw rendered outputs before the downstream pipeline.
From top to bottom, the groups show the effects of anti-aliasing resampling, display mapping, and lens correction.
For lens correction, we compare the original Scaffold-GS rendering with the \proj-conditioned rendering at the peripheral crops.
}
\label{fig:condition-response}
\end{figure*}

\paragraph{Effect of Injection Strength.}
\Fig{fig:condition-response} visualizes how injection strength changes the raw rendered output before the downstream pipeline.
Using Scaffold-GS as the backbone, we vary one condition at a time while disabling the others.
More examples are provided in \Apx{sec:appendix:injection-strength}.

\begin{figure}[t]
\centering
\includegraphics[width=0.85\linewidth]{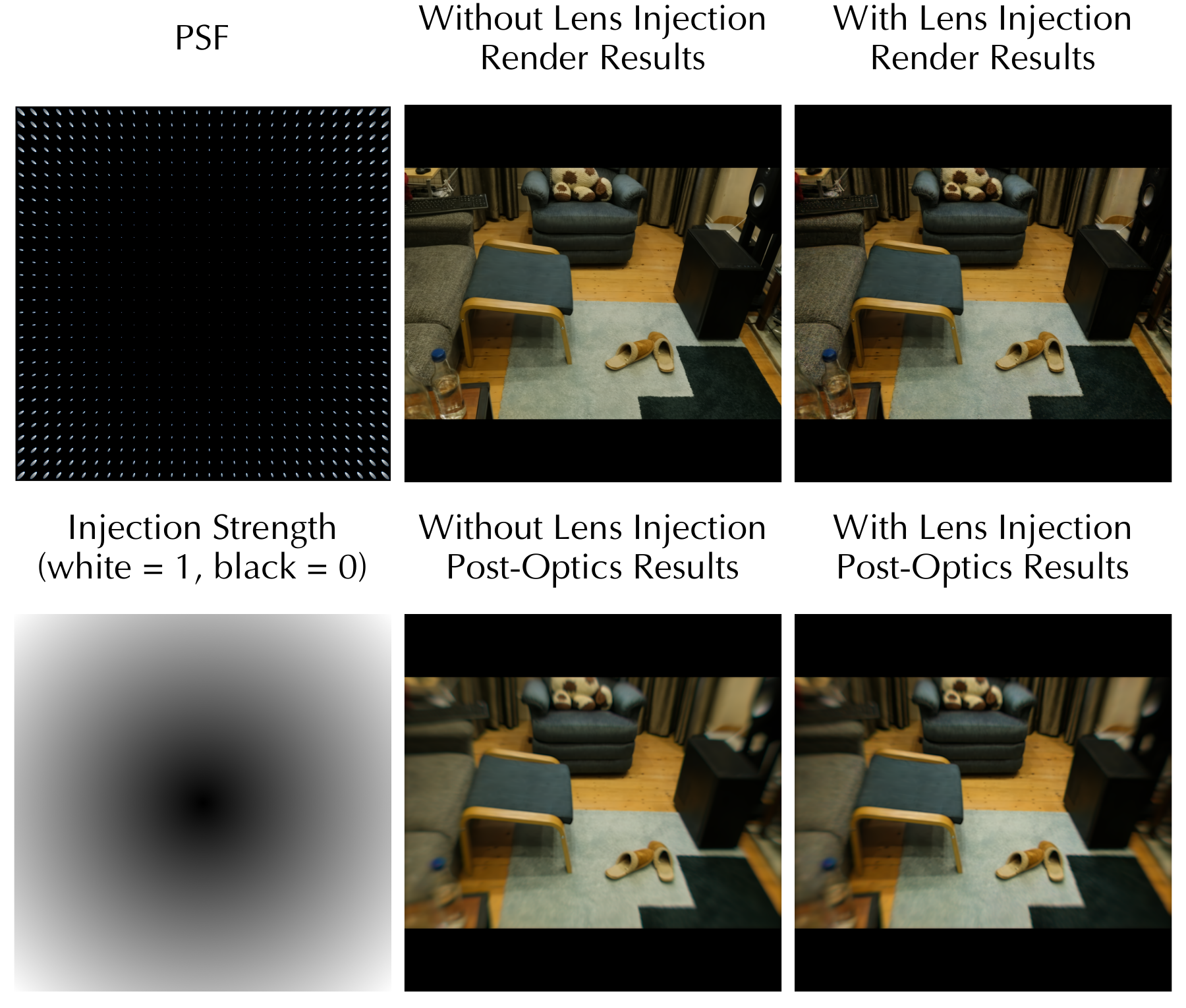 }
\caption{Lens-aware condition injection under spatially varying aberrations.
Left: the spatially varying PSFs (top) and injection-strength map (bottom).
Middle and right: results without and with lens injection, respectively; the top and bottom rows show the rendered and post-optics images.}
\label{fig:psf-optics-correction-comparison}
\end{figure}

From top to bottom, we show the results for anti-aliasing resampling, display mapping, and lens correction.
Increasing the display-resolution condition produces sharper outputs to compensate for stronger downsampling.
Increasing the display-mapping condition shifts colors toward lower-power yellowish tones.
For lens correction, we highlight peripheral crops where optical blur is stronger.
The lens-correction condition, whose strength grows from the optical center to the periphery, enhances peripheral details to compensate for the optical distortion.
\Fig{fig:psf-optics-correction-comparison} further visualizes the relationship across the spatially varying PSFs, injection strength, and rendered/post-optics images, providing an intuitive view of how the lens-correction condition depends on image position.

\paragraph{Robustness to Downstream-Model Mismatch.}
We evaluate robustness by training \proj with one canonical set of downstream-pipeline parameters and testing it under perturbed parameters that emulate simulation-to-hardware mismatch.
\Tbl{tab:downstream-robustness} reports the improvement of \proj over SS-Scaffold-GS under downstream-model perturbations, where each setting perturbs only one parameter.
Across variations in PSF spatial extent, RGB wavelengths, display gamma, and the display-power model, \proj consistently improves quality.
These results demonstrate its robustness to moderate inaccuracies in the simulation.

\subsection{Design Choice Analysis and Ablation Studies}
\label{sec:experiments:analysis}

\paragraph{Condition Dimension.}
A larger condition dimension gives the plug-in more capacity, but also increases overhead.
\Fig{fig:condition-dimension} studies this trade-off by varying the per-module condition dimension on the Garden and Lego scenes.
We therefore choose $N=8$ for each downstream module, giving a total condition dimension of $24$, as the default balance between quality and overhead.

\begin{table}[t]
  \centering
  \caption{Robustness to downstream-model mismatch on Mip-NeRF~360. Values report the quality improvement of \proj over SS-Scaffold-GS.}
  \label{tab:downstream-robustness}
  \setlength{\tabcolsep}{4pt}
  \resizebox{0.8\linewidth}{!}{%
  \begin{tabular}{lccc}
  \toprule
  Setting & $\Delta$PSNR$\uparrow$ & $\Delta$SSIM$\uparrow$ & $\Delta$LPIPS$\downarrow$ \\
  \midrule
  Canonical & +0.37 & +0.028 & -0.031 \\
  PSF $0.9\times$ & +0.38 & +0.028 & -0.032 \\
  PSF $1.1\times$ & +0.37 & +0.026 & -0.028 \\
  RGB wavelengths $-10$ nm & +0.37 & +0.028 & -0.030 \\
  RGB wavelengths $+10$ nm & +0.35 & +0.027 & -0.029 \\
  Gamma $2.3$ & +0.36 & +0.027 & -0.031 \\
  Gamma $2.5$ & +0.38 & +0.029 & -0.030 \\
  Blue power $0.9\times$ & +0.33 & +0.027 & -0.030 \\
  Blue power $0.8\times$ & +0.28 & +0.026 & -0.030 \\
  \bottomrule
  \end{tabular}%
  }
\end{table}

\begin{table}[t]
  \centering
  \caption{Ablation studies on Mip-NeRF~360 and NeRF Synthetic. Results are averaged over all scenes and downstream-pipeline settings.}
  \vspace{-10pt}
  \label{tab:condition_ablation}
  \setlength{\tabcolsep}{4pt}
  \resizebox{\linewidth}{!}{%
  \begin{tabular}{lcccccc}
  \toprule
  \multirow{2}{*}{Method} & \multicolumn{3}{c}{{Mip-NeRF~360}} & \multicolumn{3}{c}{NeRF Synthetic} \\
  \cmidrule(lr){2-4}\cmidrule(lr){5-7}
   & PSNR$\uparrow$ & SSIM$\uparrow$ & LPIPS$\downarrow$ & PSNR$\uparrow$ & SSIM$\uparrow$ & LPIPS$\downarrow$ \\
  \midrule
  ControlGS & 19.33 & 0.524 & 0.528 & 23.59 & 0.871 & 0.137 \\
  w/o Lens & 19.25 & 0.521 & 0.534 & 23.52 & 0.870 & 0.139 \\
  w/o Display Mapping & 19.07 & 0.511 & 0.531 & 23.42 & 0.870 & 0.138 \\
  w/o Resampling & 19.32 & 0.523 & 0.530 & 23.58 & 0.871 & 0.137 \\
  w/o State-Aware & 19.05 & 0.513 & 0.526 & 23.40 & 0.870 & 0.136 \\
  SS-Scaffold-GS & 18.96 & 0.496 & 0.558 & 23.35 & 0.869 & 0.147 \\
  \bottomrule
  \end{tabular}
  }
\end{table}

\paragraph{Per-module Condition Ablation.}
\Tbl{tab:condition_ablation} ablates the condition modules on Scaffold-GS over all scenes in Mip-NeRF~360 and NeRF Synthetic.
We remove the lens-correction, resampling, and display-mapping conditions one at a time while keeping the training and evaluation protocol unchanged. The results show that the module contributions are unequal: display mapping and lens correction provide larger gains, whereas removing the resampling condition causes only a small degradation, which is expected as supersampling already compensates for much of the quality loss caused by sampling-rate changes.
This suggests that future work could allocate more condition budget to more important downstream modules.

\paragraph{State-aware Injection Ablation.}
To verify that the gains do not merely result from additional conditioning capacity, we compare \proj against a state-blind baseline with the same $24$-dimensional per-feature conditions.
This baseline fixes the injection strength to $1$ during both training and inference, thereby removing run-time state awareness.
\Tbl{tab:condition_ablation} shows that state-aware injection achieves higher overall quality than this parameter-matched baseline.

\section{Discussion and Limitations}
\label{sec:discuss}

\paragraph{Sim-to-Real Gap.}
\proj evaluates the virtual image after a simulated display-optics path, as justified in \Apx{sec:appendix:optics-simulation}.
However, this path does not model much of the human visual system (HVS)\footnote{With the exception of the eye optics (e.g., cornea and pupil), which we must model in order to generate post-optics virtual images.}.
Integrating human vision into end-to-end XR evaluation remain an open challenge~\cite{chapiro2024ar, mantiuk2024colorvideovdp, zhao2022spatiotemporal, padmanaban2020perceptual}.
This sim-to-real gap is orthogonal to our main contribution of adapting rendering to downstream processing parameters.
We leave HVS-in-the-loop optimization and user study to future work.

\paragraph{Additional Downstream Modules.}
We do not model run-time reprojection (e.g., space/time warping) in XR~\cite{van2016asynchronous, evangelakos2016extended, mark1999postrendering}.
Such warping requires accurate depth and visibility, while reliable GS geometry remains an active research area~\cite{wolf2024gs2mesh,huang20242d,yu2024gaussian}.
We leave incorporating warping as an additional downstream module to future work.

Nevertheless, \proj is not limited to the downstream modules considered in this work.
Its conditional-injection framework can be extended to other modules through module-specific designs. For example, in foveated rendering, the injection strength could be conditioned on the projected eccentricity of each Gaussian feature.
Changes to the downstream modules or physical headset may therefore require retraining or fine-tuning.
Such system-specific calibration is common in XR; for example, lens correction requires knowledge of optical aberrations, while display mapping depends on the display power model.

\paragraph{Generalization to Asymmetric Optics.}
Our current injection-strength design targets radially symmetric lenses, with blur extent varying by distance from the optical center.
The framework can support asymmetric lenses by conditioning injection strength on each Gaussian’s projected image-plane coordinates $(x,y)$ rather than radius alone.

\paragraph{Gains with Compact Optics.}
The gains of \proj may appear modest because Google Cardboard, used for its open optical parameters, uses a bulky optical design with relatively low aberration, much of which is already corrected by post-processing.
As the optics in future XR devices become more compact, they will introduce stronger aberrations that might be be fully corrected by post-processing alone, increasing the value and benefits of \proj.

\section{Conclusion}
\label{sec:conclusion}

We introduced \proj, a downstream-aware condition-injection framework for XR neural Gaussian rendering.
\proj integrate the entire downstream processing pipeline, both the computational and the hardware components, into the optimization objective.
\proj dynamically generates Gaussian primitives conditioned upon the downstream processing parameters.
\proj consistently improves end-to-end quality with small overhead.

\begin{acks}
The work is partially supported by NSF
Award \#2225860, \#2126642, and \#2044963 and a Meta research grant.
\end{acks}

\clearpage

\clearpage
\bibliographystyle{ACM-Reference-Format}
\bibliography{references}

@inproceedings{zwicker2001surface,
  title={Surface splatting},
  author={Zwicker, Matthias and Pfister, Hanspeter and Van Baar, Jeroen and Gross, Markus},
  booktitle={Proceedings of the 28th annual conference on Computer graphics and interactive techniques},
  pages={371--378},
  year={2001}
}

@article{mantiuk2023hdr,
  title={HDR-VDP-3: A multi-metric for predicting image differences, quality and contrast distortions in high dynamic range and regular content},
  author={Mantiuk, Rafal K and Hammou, Dounia and Hanji, Param},
  journal={arXiv preprint arXiv:2304.13625},
  year={2023}
}

@article{Mantiuk2021FovVideoVDP,
  title={FovVideoVDP},
  author={Rafał K. Mantiuk and Alexandre Chapiro and Gizem Rufo and Trisha Lian and Rafał K. Mantiuk and Gyorgy Denes and Alexandre Chapiro and Anton Kaplanyan},
  journal={ACM Transactions on Graphics (TOG)},
  year={2021},
  volume={40},
  pages={1 - 19}
}

@incollection{andersson2021visualizing,
  title={Visualizing errors in rendered high dynamic range images},
  author={Andersson, Pontus and Nilsson, Jim and Shirley, Peter and Akenine-M{\"o}ller, Tomas},
  booktitle={Eurographics-Short Papers},
  pages={25--28},
  year={2021},
  publisher={Eurographics-European Association for Computer Graphics}
}

@inproceedings{zwicker2001ewa,
  title={Ewa volume splatting},
  author={Zwicker, Matthias and Pfister, Hanspeter and Van Baar, Jeroen and Gross, Markus},
  booktitle={Proceedings Visualization, 2001. VIS'01.},
  year={2001},
  organization={IEEE}
}

@article{kerbl20233dgs,
  title={3d gaussian splatting for real-time radiance field rendering.},
  author={Kerbl, Bernhard and Kopanas, Georgios and Leimk{\"u}hler, Thomas and Drettakis, George and others},
  journal={ACM Trans. Graph.},
  volume={42},
  number={4},
  pages={139--1},
  year={2023}
}

@inproceedings{zhang2023adding,
  title={Adding conditional control to text-to-image diffusion models},
  author={Zhang, Lvmin and Rao, Anyi and Agrawala, Maneesh},
  booktitle={Proceedings of the IEEE/CVF international conference on computer vision},
  pages={3836--3847},
  year={2023}
}

@inproceedings{lin2025powergs,
  title={PowerGS: Display-Rendering Power Co-Optimization for Neural Rendering in Power-Constrained XR Systems},
  author={Lin, Weikai and Kondguli, Sushant and Marshall, Carl and Zhu, Yuhao},
  booktitle={Proceedings of the SIGGRAPH Asia 2025 Conference Papers},
  pages={1--12},
  year={2025}
}

@misc{googlecardboard,
  title        = {Google Cardboard},
  author       = {{Google}},
  howpublished = {\url{https://arvr.google.com/cardboard/}},
  year         = {2014}
}

@misc{ansyszemaxopticstudio,
  title        = {{Ansys Zemax OpticStudio}},
  author       = {{Ansys}},
  year         = {2026},
  howpublished = {\url{https://www.ansys.com/products/optics/ansys-zemax-opticstudio}}
}

@inproceedings{zhang2018unreasonable,
  title={The unreasonable effectiveness of deep features as a perceptual metric},
  author={Zhang, Richard and Isola, Phillip and Efros, Alexei A and Shechtman, Eli and Wang, Oliver},
  booktitle={Proceedings of the IEEE conference on computer vision and pattern recognition},
  pages={586--595},
  year={2018}
}

@inproceedings{scaffoldgs,
  title={Scaffold-gs: Structured 3d gaussians for view-adaptive rendering},
  author={Lu, Tao and Yu, Mulin and Xu, Linning and Xiangli, Yuanbo and Wang, Limin and Lin, Dahua and Dai, Bo},
  booktitle={Proceedings of the IEEE/CVF conference on computer vision and pattern recognition},
  pages={20654--20664},
  year={2024}
}

@article{huynh2008scope,
  title={Scope of validity of PSNR in image/video quality assessment},
  author={Huynh-Thu, Quan and Ghanbari, Mohammed},
  journal={Electronics letters},
  volume={44},
  number={13},
  pages={800--801},
  year={2008},
  publisher={IET}
}

@inproceedings{botsch2005high,
  title={High-quality surface splatting on today's GPUs},
  author={Botsch, Mario and Hornung, Alexander and Zwicker, Matthias and Kobbelt, Leif},
  booktitle={Proceedings Eurographics/IEEE VGTC Symposium Point-Based Graphics, 2005.},
  pages={17--141},
  year={2005},
  organization={IEEE}
}

@article{padmanaban2020perceptual,
  author  = {Padmanaban, Nitish and Konrad, Robert and Wetzstein, Gordon},
  title   = {A Perceptual Eyebox for Near-Eye Displays},
  journal = {Optics Express},
  volume  = {28},
  number  = {25},
  pages   = {38008--38028},
  year    = {2020},
  doi     = {10.1364/OE.408764}
}

@article{zhao2022spatiotemporal,
  title={Spatiotemporal image quality of virtual reality head mounted displays},
  author={Zhao, Chumin and Kim, Andrea S and Beams, Ryan and Badano, Aldo},
  journal={Scientific Reports},
  volume={12},
  number={1},
  pages={20235},
  year={2022},
  publisher={Nature Publishing Group UK London}
}

@inproceedings{yu2024mipsplatting,
  title={Mip-splatting: Alias-free 3d gaussian splatting},
  author={Yu, Zehao and Chen, Anpei and Huang, Binbin and Sattler, Torsten and Geiger, Andreas},
  booktitle={Proceedings of the IEEE/CVF conference on computer vision and pattern recognition},
  pages={19447--19456},
  year={2024}
}

@article{ren2024octreegs,
  title={Octree-GS: Towards Consistent Real-time Rendering with LOD-Structured 3D Gaussians},
  author={Ren, Kerui and Jiang, Lihan and Lu, Tao and Yu, Mulin and Xu, Linning and Ni, Zhangkai and Dai, Bo},
  journal={IEEE Transactions on Pattern Analysis and Machine Intelligence},
  year={2025},
  publisher={IEEE}
}

@article{hacpp,
  title={Hac++: Towards 100x compression of 3d gaussian splatting},
  author={Chen, Yihang and Wu, Qianyi and Lin, Weiyao and Harandi, Mehrtash and Cai, Jianfei},
  journal={IEEE Transactions on Pattern Analysis and Machine Intelligence},
  year={2025},
  publisher={IEEE}
}

@article{contextgs,
  title={Contextgs: Compact 3d gaussian splatting with anchor level context model},
  author={Wang, Yufei and Li, Zhihao and Guo, Lanqing and Yang, Wenhan and Kot, Alex C and Wen, Bihan},
  journal={Advances in neural information processing systems},
  volume={37},
  pages={51532--51551},
  year={2024}
}

@inproceedings{compgs,
  title={Compgs: Efficient 3d scene representation via compressed gaussian splatting},
  author={Liu, Xiangrui and Wu, Xinju and Zhang, Pingping and Wang, Shiqi and Li, Zhu and Kwong, Sam},
  booktitle={Proceedings of the 32nd ACM International Conference on Multimedia},
  pages={2936--2944},
  year={2024}
}

@article{fan2023lightgaussian,
  title={Lightgaussian: Unbounded 3d gaussian compression with 15x reduction and 200+ fps},
  author={Fan, Zhiwen and Wang, Kevin and Wen, Kairun and Zhu, Zehao and Xu, Dejia and Wang, Zhangyang},
  journal={Advances in neural information processing systems},
  volume={37},
  pages={140138--140158},
  year={2024}
}

@inproceedings{lin2024metasapiens,
  title={Metasapiens: Real-time neural rendering with efficiency-aware pruning and accelerated foveated rendering},
  author={Lin, Weikai and Feng, Yu and Zhu, Yuhao},
  booktitle={Proceedings of the 30th ACM International Conference on Architectural Support for Programming Languages and Operating Systems, Volume 1},
  pages={669--682},
  year={2025}
}

@article{chapiro2024ar,
  title={Ar-david: Augmented reality display artifact video dataset},
  author={Chapiro, Alexandre and Kim, Dongyeon and Asano, Yuta and Mantiuk, Rafa{\l} K},
  journal={ACM Transactions on Graphics (TOG)},
  volume={43},
  number={6},
  pages={1--11},
  year={2024},
  publisher={ACM New York, NY, USA}
}

@article{duinkharjav2022color,
  title={Color-perception-guided display power reduction for virtual reality},
  author={Duinkharjav, Budmonde and Chen, Kenneth and Tyagi, Abhishek and He, Jiayi and Zhu, Yuhao and Sun, Qi},
  journal={ACM Transactions on Graphics (TOG)},
  volume={41},
  number={6},
  pages={1--16},
  year={2022},
  publisher={ACM New York, NY, USA}
}

@article{chen2024pea,
  title={Pea-pods: Perceptual evaluation of algorithms for power optimization in xr displays},
  author={Chen, Kenneth and Wan, Thomas and Matsuda, Nathan and Ninan, Ajit and Chapiro, Alexandre and Sun, Qi},
  journal={ACM Transactions on Graphics (TOG)},
  volume={43},
  number={4},
  pages={1--17},
  year={2024},
  publisher={ACM New York, NY, USA}
}

@article{mantiuk2024colorvideovdp,
  title={ColorVideoVDP: A visual difference predictor for image, video and display distortions},
  author={Mantiuk, Rafal K and Hanji, Param and Ashraf, Maliha and Asano, Yuta and Chapiro, Alexandre},
  journal={arXiv preprint arXiv:2401.11485},
  year={2024}
}

@inproceedings{huang20242d,
  title={2d gaussian splatting for geometrically accurate radiance fields},
  author={Huang, Binbin and Yu, Zehao and Chen, Anpei and Geiger, Andreas and Gao, Shenghua},
  booktitle={ACM SIGGRAPH 2024 conference papers},
  pages={1--11},
  year={2024}
}

@article{yu2024gaussian,
  title={Gaussian opacity fields: Efficient adaptive surface reconstruction in unbounded scenes},
  author={Yu, Zehao and Sattler, Torsten and Geiger, Andreas},
  journal={ACM Transactions on Graphics (ToG)},
  volume={43},
  number={6},
  pages={1--13},
  year={2024},
  publisher={ACM New York, NY, USA}
}

@inproceedings{wolf2024gs2mesh,
  title={Gs2mesh: Surface reconstruction from gaussian splatting via novel stereo views},
  author={Wolf, Yaniv and Bracha, Amit and Kimmel, Ron},
  booktitle={European Conference on Computer Vision},
  pages={207--224},
  year={2024},
  organization={Springer}
}

@book{mark1999postrendering,
  title={Postrendering 3D image warping: Visibility, reconstruction, and performance for depth-image warping},
  author={Mark, William R},
  year={1999},
  publisher={The University of North Carolina at Chapel Hill}
}

@inproceedings{van2016asynchronous,
  title={The asynchronous time warp for virtual reality on consumer hardware},
  author={Van Waveren, Johannes Marinus Paulus},
  booktitle={Proceedings of the 22nd ACM Conference on Virtual Reality Software and Technology},
  pages={37--46},
  year={2016}
}

@inproceedings{evangelakos2016extended,
  title={Extended timewarp latency compensation for virtual reality},
  author={Evangelakos, Daniel and Mara, Michael},
  booktitle={Proceedings of the 20th ACM SIGGRAPH Symposium on Interactive 3D Graphics and Games},
  pages={193--194},
  year={2016}
}

@article{vrsplatting,
  title={Vr-splatting: Foveated radiance field rendering via 3d gaussian splatting and neural points},
  author={Franke, Linus and Fink, Laura and Stamminger, Marc},
  journal={Proceedings of the ACM on Computer Graphics and Interactive Techniques},
  volume={8},
  number={1},
  pages={1--21},
  year={2025},
  publisher={ACM New York, NY}
}

@misc{waveshare55amoled,
  author       = {{Waveshare}},
  title        = {{5.5inch Capacitive Touch AMOLED Display Compatible with Raspberry Pi 4B, 1080$\times$1920 Resolution HDMI Toughened Glass Panel}},
  year         = {2026},
  howpublished = {\url{https://www.amazon.com/gp/product/B083BKSVNP/}},
}

@misc{openxr_distortion,
  title        = {{OpenXR} Specification},
  author       = {{Khronos OpenXR Working Group}},
  year         = {2026},
  note         = {Version 1.1.59},
  url          = {https://registry.khronos.org/OpenXR/specs/1.1/html/xrspec.html}
}

@inproceedings{pohl2013improved,
  title={Improved pre-warping for wide angle, head mounted displays},
  author={Pohl, Daniel and Johnson, Gregory S and Bolkart, Timo},
  booktitle={Proceedings of the 19th ACM symposium on Virtual Reality Software and Technology},
  pages={259--262},
  year={2013}
}

@inproceedings{martschinke2019gaze,
  title={Gaze-dependent distortion correction for thick lenses in hmds},
  author={Martschinke, Jonathan and Martschinke, Jana and Stamminger, Marc and Bauer, Frank},
  booktitle={2019 IEEE Conference on virtual reality and 3D user interfaces (VR)},
  pages={1848--1851},
  year={2019},
  organization={IEEE}
}

@article{hiroi2022neuraldistortion,
  title={Neural distortion fields for spatial calibration of wide field-of-view near-eye displays},
  author={Hiroi, Yuichi and Someya, Kiyosato and Itoh, Yuta},
  journal={Optics Express},
  volume={30},
  number={22},
  pages={40628--40644},
  year={2022},
  publisher={Optica Publishing Group}
}

@article{wang2004ssim,
  title={Image quality assessment: from error visibility to structural similarity},
  author={Wang, Zhou and Bovik, Alan C and Sheikh, Hamid R and Simoncelli, Eero P},
  journal={IEEE transactions on image processing},
  volume={13},
  number={4},
  pages={600--612},
  year={2004},
  publisher={IEEE}
}

@inproceedings{barron2022mipnerf360,
  title={Mip-nerf 360: Unbounded anti-aliased neural radiance fields},
  author={Barron, Jonathan T and Mildenhall, Ben and Verbin, Dor and Srinivasan, Pratul P and Hedman, Peter},
  booktitle={Proceedings of the IEEE/CVF conference on computer vision and pattern recognition},
  pages={5470--5479},
  year={2022}
}

@inproceedings{mildenhall2020nerf,
  title={NeRF: Representing Scenes as Neural Radiance Fields for View Synthesis},
  author={Ben Mildenhall and Pratul P. Srinivasan and Matthew Tancik and Jonathan T. Barron and Ravi Ramamoorthi and Ren Ng},
  year={2020},
  booktitle={ECCV},
}

@article{sun2021differentiable,
  title={Differentiable compound optics and processing pipeline optimization for end-to-end camera design},
  author={Tseng, Ethan and Mosleh, Ali and Mannan, Fahim and St-Arnaud, Karl and Sharma, Avinash and Peng, Yifan and Braun, Alexander and Nowrouzezahrai, Derek and Lalonde, Jean-Francois and Heide, Felix},
  journal={ACM Transactions on Graphics (TOG)},
  volume={40},
  number={2},
  pages={1--19},
  year={2021},
  publisher={ACM New York, NY}
}

@article{zhang2024end,
  title={End-to-end automatic lens design with a differentiable diffraction model},
  author={Zhang, Wenguan and Ren, Zheng and Zhou, Jingwen and Chen, Shiqi and Feng, Huajun and Li, Qi and Xu, Zhihai and Chen, Yueting},
  journal={Optics Express},
  volume={32},
  number={25},
  pages={44328--44345},
  year={2024},
  publisher={Optica Publishing Group}
}

@article{fan2024fovgs,
  title={Fov-gs: Foveated 3d gaussian splatting for dynamic scenes},
  author={Fan, Runze and Wu, Jian and Shi, Xuehuai and Zhao, Lizhi and Ma, Qixiang and Wang, Lili},
  journal={IEEE Transactions on Visualization and Computer Graphics},
  year={2025},
  publisher={IEEE}
}

@article{lin2026lowpowar,
  title={LowPowAR: Power-Constrained Tone Mapping for Augmented Reality},
  author={Lin, Weikai and Zhao, Sheng and Ross, Ian and Marshall, Carl and Kondguli, Sushant and Zhu, Yuhao},
  journal={IEEE Transactions on Visualization and Computer Graphics},
  year={2026},
  publisher={IEEE},
  note={To appear}
}

\clearpage
\twocolumn[
\begin{flushleft}
\vspace{0.25em}
{\Huge
\linespread{1.05}\selectfont
\proj:  Conditioning Neural Gaussians for\\[0.1em]
Downstream-Processing--Aware XR Rendering \\
}
\vspace{0.7em}
{\huge
Supplementary Material\\[0.3em]
}
\end{flushleft}
]

\appendix
\setcounter{figure}{0}
\setcounter{table}{0}
\setcounter{equation}{0}
\renewcommand{\thefigure}{S\arabic{figure}}
\renewcommand{\thetable}{S\arabic{table}}
\renewcommand{\theequation}{S\arabic{equation}}

This supplementary material provides implementation details, extended results, and additional diagnostics for \proj.
We first describe the full end-to-end XR evaluation protocol, including post-processing run-time state ranges, display-optics simulation, and metric computation after the XR pipeline.
We then report per-scene and per-setting results, additional qualitative comparisons.

\section{Experimental Protocol Details}
\label{sec:appendix:implementation}
These settings are used for all main experiments unless otherwise specified.
We train each condition-injection plug-in for $10{,}000$ iterations with a condition learning rate of $3\times10^{-3}$.
For each downstream module, we use an $8$-dimensional condition vector, giving a total condition dimension of $24$ across anti-aliasing resampling, lens correction, and display mapping.

During training, we randomly sample run-time  states from the supported range of each downstream module.
For anti-aliasing resampling, we sample the downsampling scale in log space over the supported range.
For display mapping, we uniformly sample the display-power-saving ratio from $0\%$ to $80\%$.
For lens correction, the injection strength is not sampled directly; it is determined by each Gaussian feature's screen-space position under the sampled camera pose.
All sampled states are continuous floating-point values, so the plug-in is not trained only on the discrete $4{\times}5$ evaluation grid.

Because the full differentiable PSF model is expensive, reaching only about one iteration per second on an RTX 5090, we use a distortion-only approximation for the first $9{,}000$ iterations and enable the full PSF model during the final $1{,}000$ iterations.
We use the same $10{,}000$-iteration schedule to train the lens-distortion MLP used for pre-warping, optimizing it with the original 3DGS loss.

\section{Dynamic Runtime State Examples}
\label{sec:appendix:runtime-state-examples}

This section illustrates why the effective rendering resolution of an object and the aberration state of each Gaussian feature should change at run time.
As shown in the top of \Fig{fig:supp:runtime-state-examples}, when the camera moves farther from an object, the object occupies a smaller region on the image plane.
This reduces its effective screen-space resolution, requiring a different anti-aliasing resampling state.

The bottom of \Fig{fig:supp:runtime-state-examples} illustrates the view-dependent aberration state.
The same Gaussian primitive can project to different screen-space locations under different camera poses.
Because lens aberrations are spatially varying, different screen-space locations experience different optical aberrations.
Therefore, the same Gaussian primitive should render differently under different camera views to adapt to location-dependent lens aberrations.
Thus, the aberration state of a Gaussian primitive naturally depends on the camera pose.
\section{Post-Rendering Module Details}
\label{sec:appendix:post-rendering-module-details}

This section documents the post-rendering modules used in the evaluation pipeline.
These modules include anti-aliasing resampling, lens correction (pre-warping), display mapping, the OLED display-power model, and physical display-optics simulation.
The goal is to make the end-to-end metric path reproducible.

\subsection{Lens Correction Through Pre-Warping}
\label{sec:appendix:lens-prewarping}

For lens correction, we learn an inverse distortion map to compensate for lens distortion with an MLP, following prior
work~\cite{openxr_distortion,pohl2013improved,martschinke2019gaze,hiroi2022neuraldistortion}.
The MLP compensates for end-to-end lens aberrations.
We train the MLP end-to-end with the original 3DGS loss for $10{,}000$ iterations using a learning rate of $10^{-3}$.
After training, the MLP can be baked into a pixel-space lookup table with little inference overhead.

In addition to compensating for lens distortion, the MLP also predicts a per-pixel, per-channel scaling map, which is multiplied element-wise with the pre-warped image.
This is needed because the display-optics path can introduce  color shifts/chromatic aberrations.
The scaling factor reduces systematic color bias and prevents pixel-wise metrics such as PSNR from being dominated by color mismatch.

\subsection{Anti-Aliasing Resampling}
\label{sec:appendix:display-resolution-matching}

For anti-aliasing resampling, we render the image at a supersampled resolution and then resample it to the target display resolution using average pooling.
This implements a simple low-pass filtering step before sampling, which reduces aliasing when the target display resolution is lower than the rendering resolution.
For the baseline renderers, we keep their default anti-aliasing mechanisms: {Mip-Splatting}~\cite{yu2024mipsplatting} uses its 3D-2D filtering scheme, while the other Gaussian baselines use approximate EWA filtering~\cite{botsch2005high}.
For the supersampled baselines, we render at the native resolution used to train each 3DGS model and apply average pooling to obtain the target resolution, following the same strategy as \proj, so the comparison isolates the effect of downstream-aware condition injection.

\subsection{Display Mapping}
\label{sec:appendix:display-mapping}

For display mapping, we use uniform dimming (UD) for display-power saving.
We use target power-saving ratios $\rho \in [0,0.8]$.
For each setting, we first compute the reference display power $P_{\mathrm{ref}}$ using the ground-truth image.
We then set the target power as $P_{\mathrm{target}}=(1-\rho)P_{\mathrm{ref}}$, where $\rho$ is the power-saving ratio.
UD scales the rendered image so that its display power matches $P_{\mathrm{target}}$.
This ensures that all methods are evaluated under the same target power for each run-time  setting.

\subsection{Physical Display}
\label{sec:appendix:physical-display}

We model the OLED panel as a three-primary display in which each primary is represented by a single wavelength.
The input image is stored in RGB space and treated as sRGB for the display-optics simulation.
Before optical propagation, we first decode the image to linear RGB and convert each RGB vector to three monochromatic display-primary intensities:
\begin{equation}
  \mathbf{s}(p)
  =
  \mathbf{M}_{\mathrm{RGB}\rightarrow\lambda}\,
  \mathbf{i}_{\mathrm{lin}}(p),
  \qquad
  \mathbf{s}(p)
  =
  \big[
    s_{656}(p),\,
    s_{550}(p),\,
    s_{486}(p)
  \big]^{\top}.
  \label{eq:supp:rgb-to-spectral-display}
\end{equation}
Here $\mathbf{i}_{\mathrm{lin}}(p)$ is the linear RGB value at display pixel $p$, and the three wavelengths are $656$ nm, $550$ nm, and $486$ nm for the red, green, and blue display primaries, respectively.
The conversion matrix $\mathbf{M}_{\mathrm{RGB}\rightarrow\lambda}$ is obtained from the CIE color matching functions at these three wavelengths together with the standard RGB-to-XYZ transform.

For spatial emission, each display pixel is modeled as a box filter.
When the display-resolution module outputs a lower-resolution image,
we use nearest-neighbor upsampling to match the optics-simulation resolution.
This makes each low-resolution sample drive a block of display pixels,
which effectively implements the box-filter footprint of display pixels.

For display power, we follow prior XR display-power models~\cite{duinkharjav2022color,chen2024pea,lin2025powergs}
and model the image-dependent dynamic OLED power as
\begin{equation}
  P_{\mathrm{disp}}(I)
  = \sum_{x,y} \big(\alpha I_R(x,y) + \beta I_G(x,y) + \gamma I_B(x,y)\big),
  \label{eq:supp:disp-power}
\end{equation}
where $(\alpha,\beta,\gamma)$ are panel-dependent RGB power coefficients.
We use $\alpha=228.52$, $\beta=205.66$, and $\gamma=567.64$ in our experiments.
This model captures dynamic display power only; static panel power is excluded.
Because the blue-channel coefficient $\gamma$ is substantially larger than the red and green coefficients, reducing blue intensity is especially effective for lowering display power; this makes yellowish color shifts power-efficient under this panel model.
Power is computed from the RGB image values, while the monochromatic display-primary representation is used only for optical propagation.

\paragraph{Sim-to-real gap.}
The main simplification in this display model is spectral.
Real OLED primaries are not delta functions at a single wavelength; each pixel emits a wavelength-dependent spectrum.
Modeling the full spectrum would reduce this approximation, but it has two practical costs.
First, the spectra must be measured with calibrated optical instruments for the target display.
Second, the optics simulation would need to evaluate and integrate PSFs over many wavelength samples.
Because PSF simulation is already the most expensive part of the end-to-end path ($\sim$ 1 FPS on RTX 5090), this dense spectral integration would make training and evaluation substantially slower.
We therefore use three representative monochromatic primaries as a tractable display model for differentiable optimization.

\subsection{Optics Simulation}
\label{sec:appendix:optics-simulation}

\begin{table}[t]
\centering
\caption{Optical parameters used for the Google Cardboard lens simulation.
Most parameters are taken from the public Google Cardboard specification, while the lens surface radii $R_1$ and $R_2$ are optimized in Zemax.}
\label{tab:supp:cardboard-optics}
\setlength{\tabcolsep}{4pt}
\renewcommand{\arraystretch}{1.08}
\begin{tabular}{@{}ll@{}}
\toprule
Parameter & Value \\
\midrule
Object distance & $39.07$ mm \\
Lens thickness & $8.795$ mm \\
Lens material & PMMA $(n_d=1.492,\; v=57.4)$ \\
Lens diameter & $34$ mm \\
Eye relief & $18$ mm \\
Projected virtual image distance & $667$ mm \\
Display size & $24 \times 24$ mm \\
Lens surface radius $R_1$ & $290.444$ mm \; (optimized) \\
Lens surface radius $R_2$ & $-25.498$ mm \; (optimized) \\
Pupil diameter & $4$ mm \\
Retina diameter & $5.822$ mm \\
Retina curvature radius $R_{\mathrm{retina}}$ & $7.8$ mm \\
\bottomrule
\end{tabular}
\end{table}

\begin{figure*}[t]
  \centering
  \includegraphics[width=0.85\linewidth]{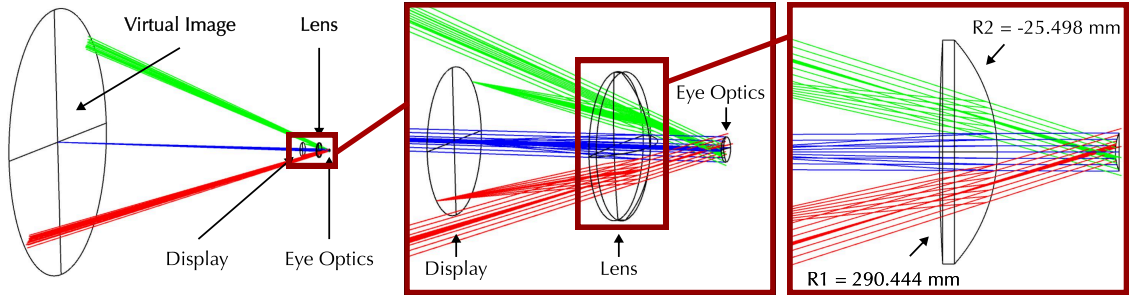}
  \caption{
  Zemax setup for our optics simulation.
  Left: overview of the full optical system.
  Middle: zoom-in view of the spatial layout of the display, lens, and eye model.
  Right: lens parameter settings used in the simulation.  }
  \label{fig:zemax}
\end{figure*}

\begin{figure*}[t]
\centering
\includegraphics[width=\linewidth]{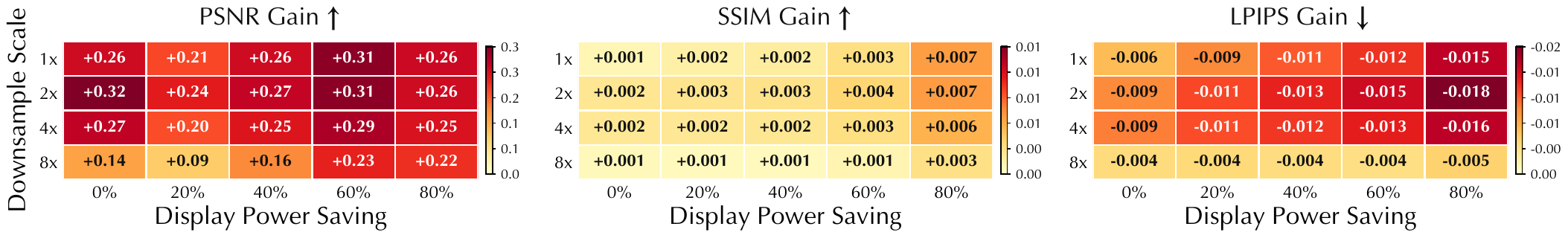}
\caption{Average quality improvement of \proj over SS-Scaffold-GS on NeRF Synthetic.
Heatmaps show PSNR, SSIM, and LPIPS gains across downsample scale and display-power-saving settings, averaged over all scenes.
\proj consistently improves end-to-end quality.}
\label{fig:supp:heatmap}
\end{figure*}

We use a Google-Cardboard single-convex lens design as the reference optics, because Google Cardboard is a canonical low-cost XR headset with publicly available specifications~\citep{googlecardboard}.
Although most of its optical parameters are publicly available, the surface radii $R_1$ and $R_2$ are not specified.
We therefore complete the lens model in Zemax OpticStudio~\citep{ansyszemaxopticstudio} by optimizing $R_1$ and $R_2$ to minimize blur on the virtual image plane at the target virtual distance.
The optimized values are $R_1=290.444$ mm and $R_2=-25.498$ mm.
\Tbl{tab:supp:cardboard-optics} reports the final lens parameters, and \Fig{fig:zemax} shows the Zemax setup.
We use a $4$ mm pupil and a curved retina with radius $7.8$ mm to determine which rays enter the eye and which retinal region is evaluated.
We then use backward ray tracing to obtain the virtual image plane used for optics simulation.
We use the image formed on this virtual image plane as the post-optics target image, and compute all end-to-end quality metrics on this plane.

Using this lens model, we export field-dependent PSFs from Zemax.
The exported PSF data contains a $30{\times}30$ field grid and three wavelength channels at $486$ nm, $550$ nm, and $656$ nm.
Each field sample stores a local PSF and the corresponding chief-ray location.

Given the spectral display image, the optics simulator forms the virtual  image independently for each wavelength:
\begin{equation}
  I_{\mathrm{virtual}}^{\lambda}(q)
  =
  \sum_{p}
  I_{\mathrm{disp}}^{\lambda}(p)\,
  K_{\lambda}\!\left(q;\,p\right),
  \qquad
  \lambda \in \{486,550,656\}\,\mathrm{nm}.
  \label{eq:supp:psf-splat}
\end{equation}
Here $p$ indexes display pixels, $q$ indexes output pixels on the simulated virtual image plane, and $K_{\lambda}(q;p)$ is the shift-variant PSF contribution from display pixel $p$ to output pixel $q$.
For each display pixel $p$, we bilinearly interpolate the four neighboring Zemax field samples to obtain its local PSF and chief ray.
The simulator then splats the display pixel $p$ intensity into a local output window centered at the wavelength-dependent chief ray and samples the interpolated PSF at the exact output-pixel $q$ centers as $K_{\lambda}(q;p)$.
After PSF propagation, we convert the three simulated single-wavelength virtual images back to linear RGB:
\begin{equation}
  \mathbf{I}_{\mathrm{virt,lin}}(q)
  =
  \mathbf{M}_{\lambda\rightarrow\mathrm{RGB}}\,
  \big[
    I_{\mathrm{virtual}}^{656}(q),\,
    I_{\mathrm{virtual}}^{550}(q),\,
    I_{\mathrm{virtual}}^{486}(q)
  \big]^{\top},
  \label{eq:supp:spectral-to-rgb}
\end{equation}
and then apply the sRGB transfer function to obtain the final sRGB image used for metrics.
We implement this optics simulation with a highly-optimized custom CUDA simulator.
The CUDA path performs PSF accumulation once per wavelength channel and supports a custom backward pass with gradients to the input intensities.
This differentiability allows the PSF-based display-optics path to be included in the end-to-end training loss.
The final evaluation also uses same PSF process, so PSNR, SSIM, and LPIPS are computed on the image formed after display emission and optical propagation.

\paragraph{Sim-to-real gap.}
Our optics simulation uses two practical approximations.
For each scene, we run the simulation at the dataset's native high-resolution image resolution, i.e., the $1{\times}$ evaluation resolution, rather than an infinitely dense optical sampling grid.
We also interpolate shift-variant PSFs from a finite set of 30x30 sampled field positions.
Denser image and PSF sampling would reduce the sim-to-real gap, but would make the already expensive PSF simulation substantially slower.
We therefore use the native image resolution and sampled PSF grid as a practical trade-off for differentiable end-to-end optimization.

The larger gap is that we model only the virtual image formed by the display-optics system, not the full human visual process.
Our simulator does not include eye optics and photoreceptor responses.
Thus, the virtual image is an optics-simulated proxy for user-visible image quality rather than a complete model of human perception.
Calibrated eye models for end-to-end XR evaluation remain an open challenge and important future work.

\subsection{Simulation Validation}

\begin{figure*}[t]
  \centering
  \includegraphics[width=\linewidth]{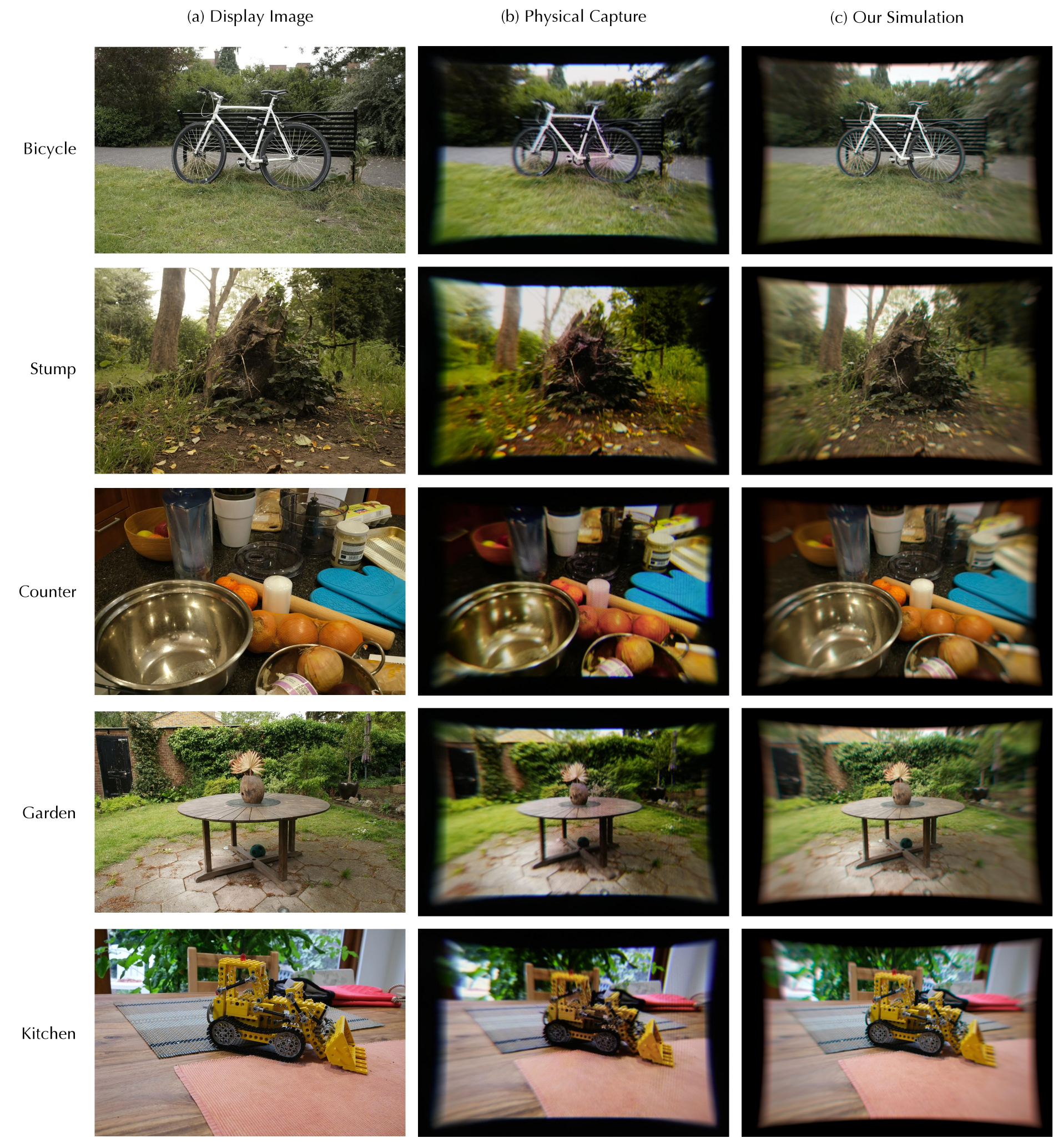}
\caption{
Camera-based verification of the display--optics simulation against real captures.
We display the input image on an AMOLED display and capture the result through a Google Cardboard lens using a calibrated Sony FE 24--70mm F2.8 GM II camera.
Left: input image shown on the display.
Middle: calibrated camera capture through the real lens.
Right: simulated result from our display-optics model.
Remaining color differences may come from the simplified emission spectrum model of the display, camera sensor response functions, and potential Image Signal Processor (ISP) post-processing, which are not fully accounted for in our simulation.
}
  \label{fig:camera}
\end{figure*}

Despite the sim-to-real gaps discussed above, our optics simulation captures key characteristics of the XR headset.
\Fig{fig:camera} compares the input image, a camera capture through a real Google Cardboard lens, and our simulated result.

We use a Waveshare 5.5-inch AMOLED display~\cite{waveshare55amoled} and a SONY ILCE-7RM5 with SONY FE 24-70mm F2.8 GM II as the camera.
The camera is positioned at the eye-viewing location to capture the image formed through the lens of the Google Cardboard.
Because the camera output can be affected by the display emission spectrum, camera sensor response functions, and possible Image Signal Processor (ISP) post-processing, we color-calibrate the captured image by fitting a $3{\times}3$ color-correction matrix.
Our simulation produces a result visually close to the camera capture, with similar lens-induced spatially varying blur and distortion.
Remaining differences mainly arise from (i) residual errors in calibration and (ii) discrepancies between the optimized lens surface radii $R_1,R_2$ and the actual Google Cardboard lens.

\section{Additional Results}
\label{sec:appendix:additional-results}

This section provides additional quantitative and qualitative results for \proj.

\subsection{\proj Generalizes to Different Neural Gaussian Backbones}
\label{sec:appendix:gen}
\proj also generalizes to other neural Gaussian backbones, with complete results provided in \Tbl{tab:supp:additional_bb}.

\subsection{Per-Scene End-to-End Quality}
\label{sec:appendix:per-scene-main-results}

\Tbl{tab:supp:xrgs-e2e-quality-per-scene-mip360} and \Tbl{tab:supp:xrgs-e2e-quality-per-scene-nerf-synthetic} report per-scene end-to-end quality.
Each row averages over all test views and the same $20$ run-time  settings used in the main paper.
The deltas compare \proj with the corresponding SS-enhanced baseline after the full XR pipeline.
These results show that the improvements reported in the main paper are consistent across scenes and are not dominated by a few examples.

\subsection{Run-Time State Quality Heatmaps}
\label{sec:appendix:run-time-state-heatmaps}

The main paper reports per-setting heatmaps on Mip-NeRF~360.
Here, we provide the same analysis on NeRF Synthetic to verify the consistency of \proj on object-centric scenes.
Using Scaffold-GS as the backbone, \Fig{fig:supp:heatmap} compares \proj with the SS-enhanced baseline on the full $4{\times}5$ run-time  grid.
Each cell reports the quality difference under one resolution downsampling factor and one display-power-saving ratio.
For all metrics, the difference is computed as \proj minus the baseline.
Thus, positive values indicate improvement for PSNR and SSIM, while negative values indicate improvement for LPIPS.
The heatmaps show that \proj consistently improves end-to-end quality across run-time  settings beyond the Mip-NeRF~360 results in the main paper.

\begin{table}[t]
\centering
\caption{Supersampling impact. SS: Supersampling.
No-SS: direct low-resolution rendering.}
\label{tab:xrgs-ss-overhead}
\setlength{\tabcolsep}{4.0pt}
\renewcommand{\arraystretch}{1.08}
\resizebox{\columnwidth}{!}{%
\begin{tabular}{lcccc}
\toprule
Method / Dataset & \multicolumn{2}{c}{Mip-NeRF~360} & \multicolumn{2}{c}{NeRF Synthetic} \\
\cmidrule(lr){2-3}\cmidrule(lr){4-5}
& FPS$\uparrow$ & Storage (MB)$\downarrow$ & FPS$\uparrow$ & Storage (MB)$\downarrow$ \\
\midrule
Scaffold-GS (No SS, 2x) & 369.7 & 180.53 & 744.6 & 15.13 \\
Scaffold-GS (No SS, 4x) & 278.2 & 180.53 & 478.0 & 15.13 \\
Scaffold-GS (No SS, 8x) & 157.8 & 180.53 & 250.7 & 15.13 \\
\bottomrule
\end{tabular}%
}
\end{table}

\subsection{Impact of Supersampling}
\label{sec:appendix:ss-overhead}
\Tbl{tab:xrgs-ss-overhead} also reports the run-time  impact of supersampling (SS).
Compared with direct Scaffold-GS rendering at $2{\times}$, $4{\times}$, and $8{\times}$ lower resolutions, SS is only slightly slower at $2{\times}$ on Mip-NeRF~360 and up to $2.8{\times}$ faster at $8{\times}$ downsampling.
Thus, SS can improve both quality and speed.
The reason is two-fold.
First, the Gaussian rasterization kernel uses a tile size optimized for high-resolution outputs, so very low-resolution outputs can lead to GPU resource underutilization.
Second, without resolution-aware level-of-detail (LOD) modeling, the renderer still processes almost the same number of Gaussian primitives even when the output resolution is reduced.

\subsection{Additional Qualitative Results}
\label{sec:appendix:additional-qualitative}

\begin{figure*}[t]
\centering
\includegraphics[width=\linewidth]{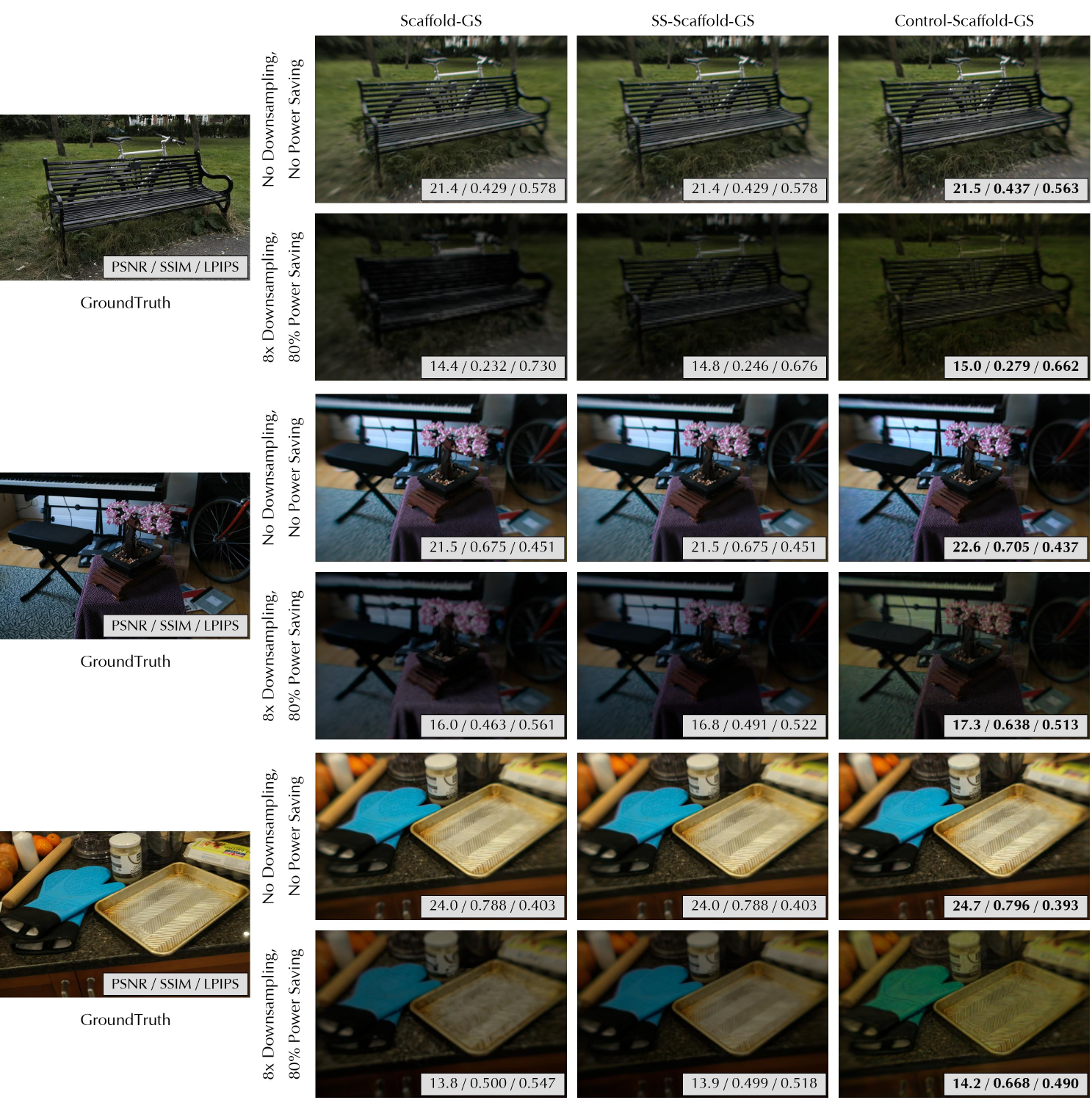}
\caption{
Additional qualitative comparisons on Bicycle, Bonsai, and Counter.
}
\Description{Additional qualitative comparisons for Bicycle, Bonsai, and Counter using Scaffold-GS, SS-Scaffold-GS, and \proj under two XR run-time  settings.}
\label{fig:supp:additional-qualitative}
\end{figure*}

\begin{figure*}[t]
\centering
\includegraphics[width=\linewidth]{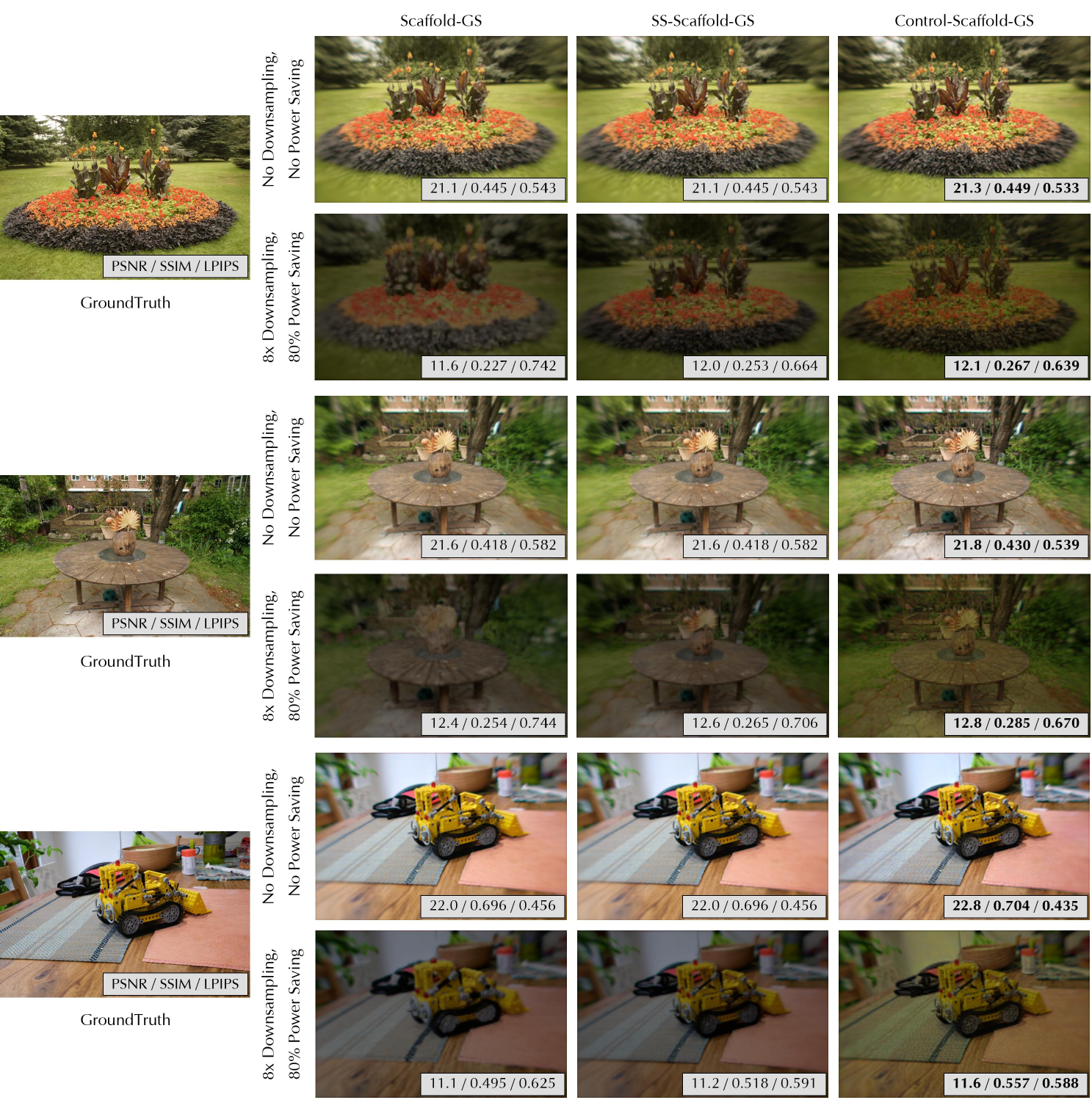}
\caption{
Additional qualitative comparisons on Flowers, Garden, and Kitchen.
}
\Description{Additional qualitative comparisons for Flowers, Garden, and Kitchen using Scaffold-GS, SS-Scaffold-GS, and \proj under two XR run-time  settings.}
\label{fig:supp:additional-qualitative-2}
\end{figure*}

\begin{figure*}[t]
\centering
\includegraphics[width=\linewidth]{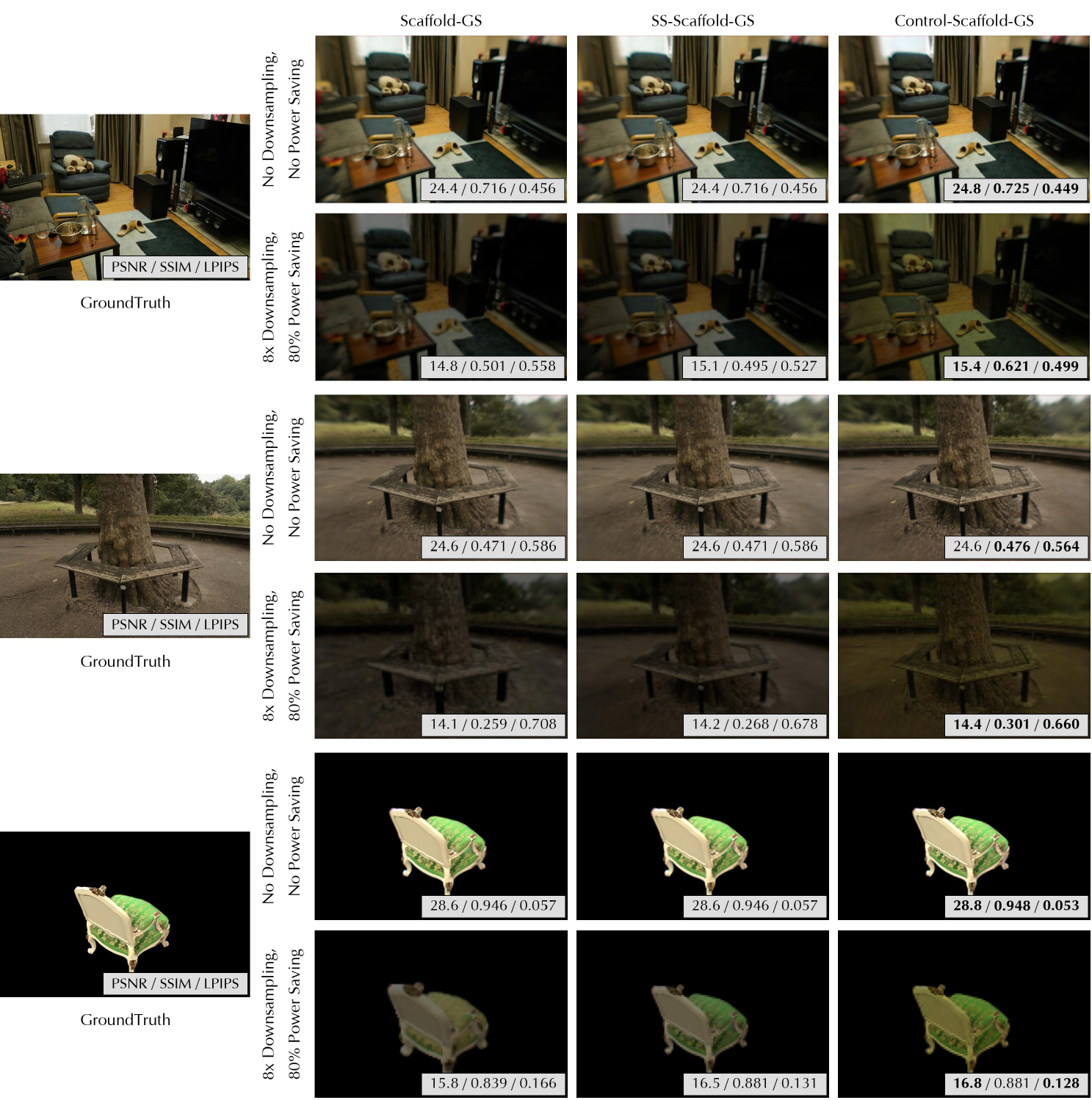}
\caption{
Additional qualitative comparisons on Room, Treehill, and Chair.
}
\Description{Additional qualitative comparisons for Room, Treehill, and Chair using Scaffold-GS, SS-Scaffold-GS, and \proj under two XR run-time  settings.}
\label{fig:supp:additional-qualitative-3}
\end{figure*}

\begin{figure*}[t]
\centering
\includegraphics[width=\linewidth]{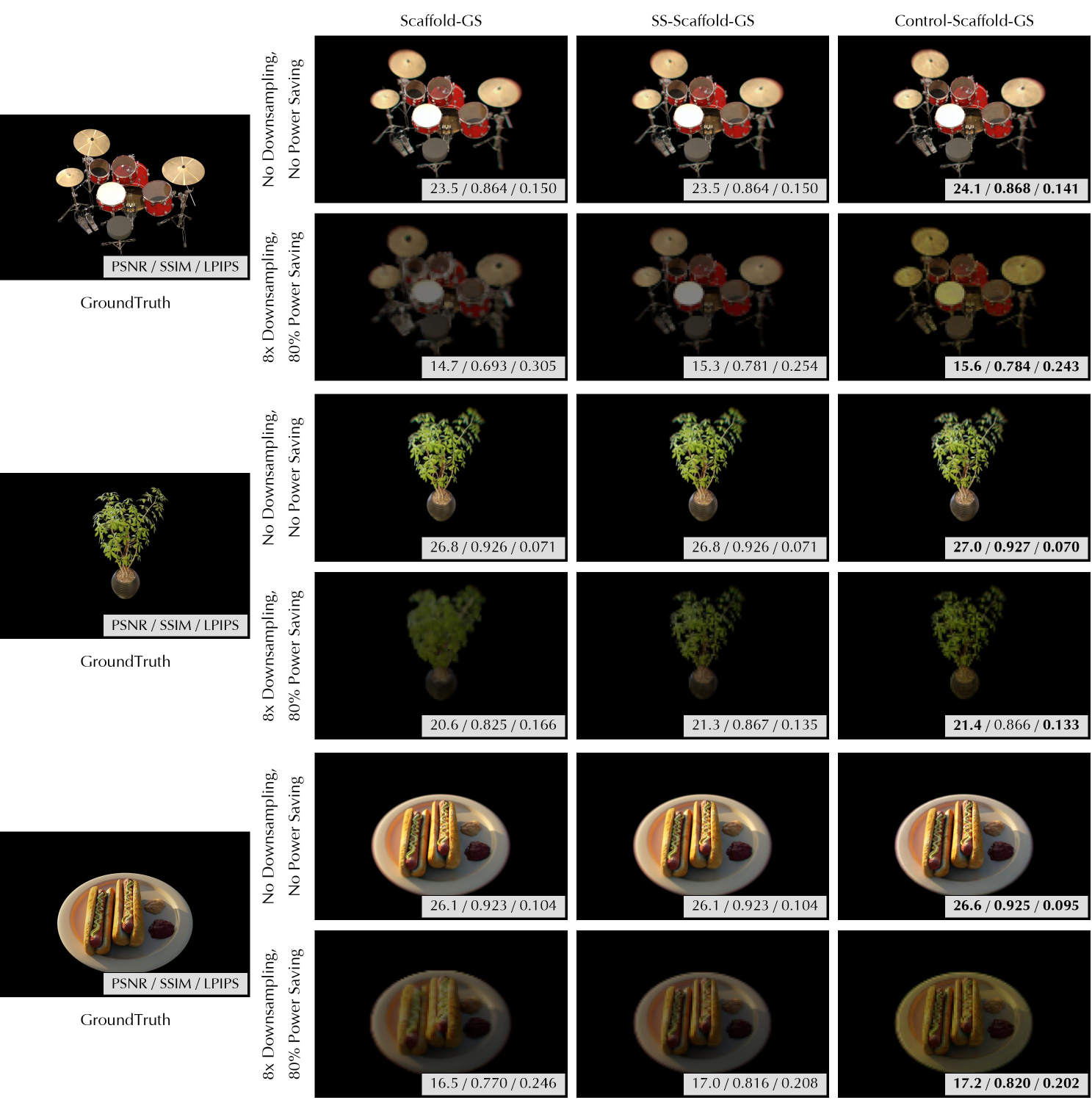}
\caption{
Additional qualitative comparisons on Drums, Ficus, and Hotdog.
}
\Description{Additional qualitative comparisons for Drums, Ficus, and Hotdog using Scaffold-GS, SS-Scaffold-GS, and \proj under two XR run-time  settings.}
\label{fig:supp:additional-qualitative-4}
\end{figure*}

\begin{figure*}[t]
\centering
\includegraphics[width=\linewidth]{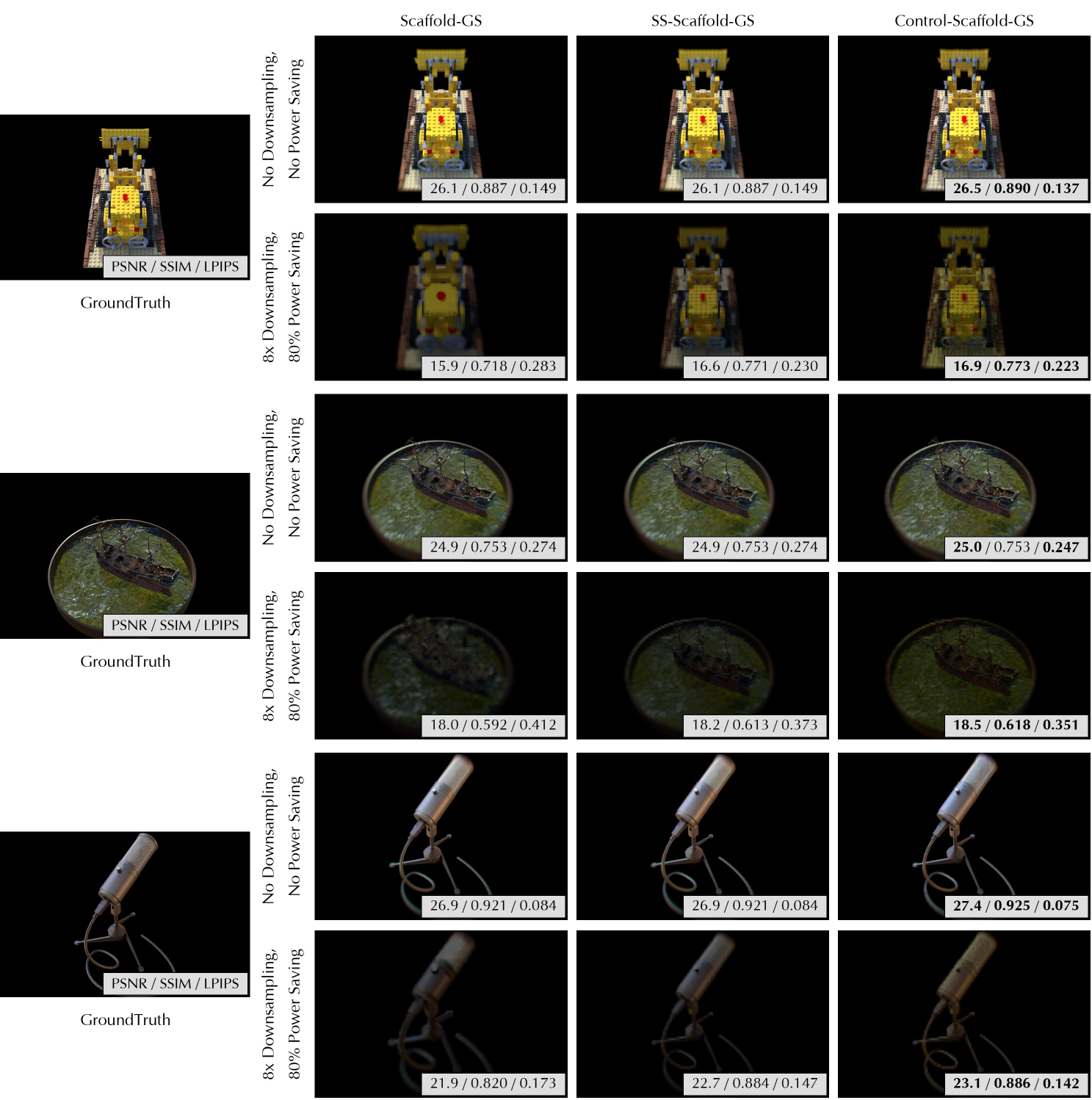}
\caption{
Additional qualitative comparisons on Lego, Ship, and Mic.
}
\Description{Additional qualitative comparisons for Lego, Ship, and Mic using Scaffold-GS, SS-Scaffold-GS, and \proj under two XR run-time  settings.}
\label{fig:supp:additional-qualitative-5}
\end{figure*}

\Fig{fig:supp:additional-qualitative}--\Fig{fig:supp:additional-qualitative-5}
provide additional qualitative comparisons across scenes and run-time  settings.
Following the main paper, each example compares the original baseline, the SS-enhanced baseline, \proj after the display-optics pipeline.
At high resolution without power saving, \proj preserves sharper peripheral details by compensating for lens-induced blur.
Under stronger downsampling and power saving, \proj better preserves image structures while shifting colors toward lower-power yellowish tones, retaining details that would otherwise be suppressed by downsampling/dimming.

\subsection{Effect of Injection Strength}
\label{sec:appendix:injection-strength}

\begin{figure*}[t]
\centering
\includegraphics[width=\linewidth,height=0.94\textheight,keepaspectratio]{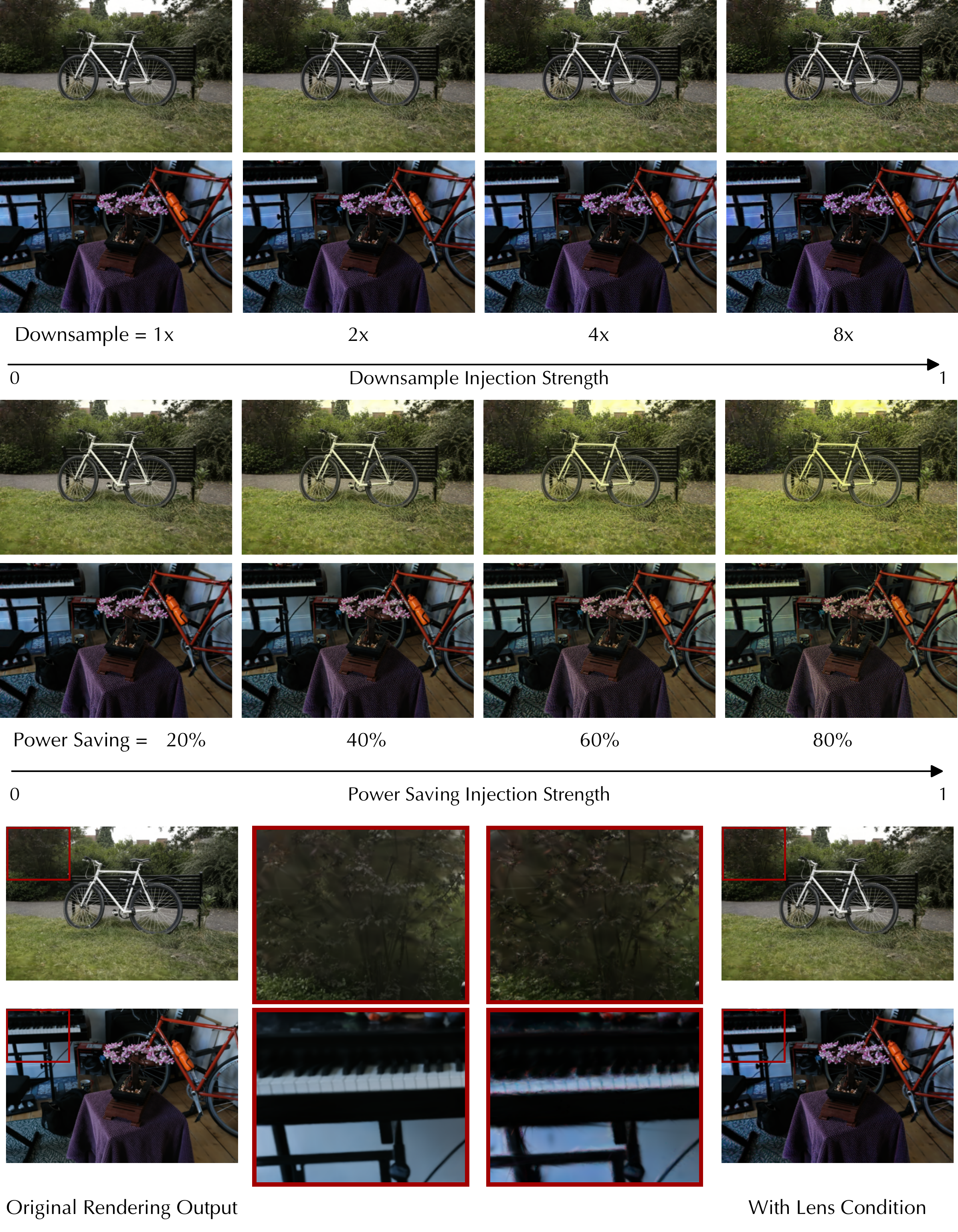}
\caption{
Additional condition-response visualization on Bicycle and Bonsai.
}
\Description{Additional condition-response visualization combining Bicycle and Bonsai scenes.}
\label{fig:supp:condition-response}
\end{figure*}

\begin{figure*}[t]
\centering
\includegraphics[width=\linewidth,height=0.94\textheight,keepaspectratio]{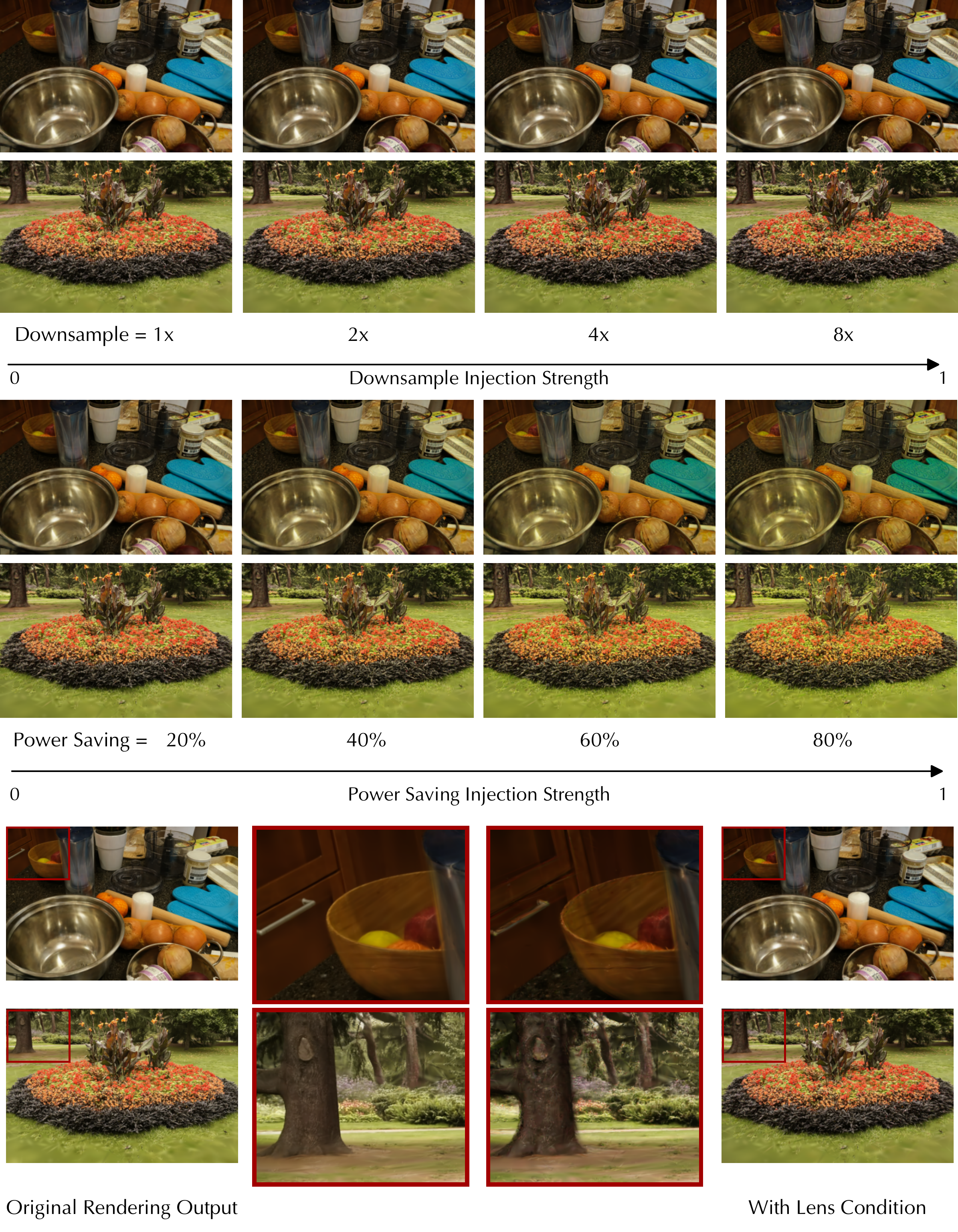}
\caption{
Additional condition-response visualization on Counter and Flowers.
}
\Description{Additional condition-response visualization combining Counter and Flowers scenes.}
\label{fig:supp:condition-response-counter-flowers}
\end{figure*}

\begin{figure*}[t]
\centering
\includegraphics[width=\linewidth,height=0.94\textheight,keepaspectratio]{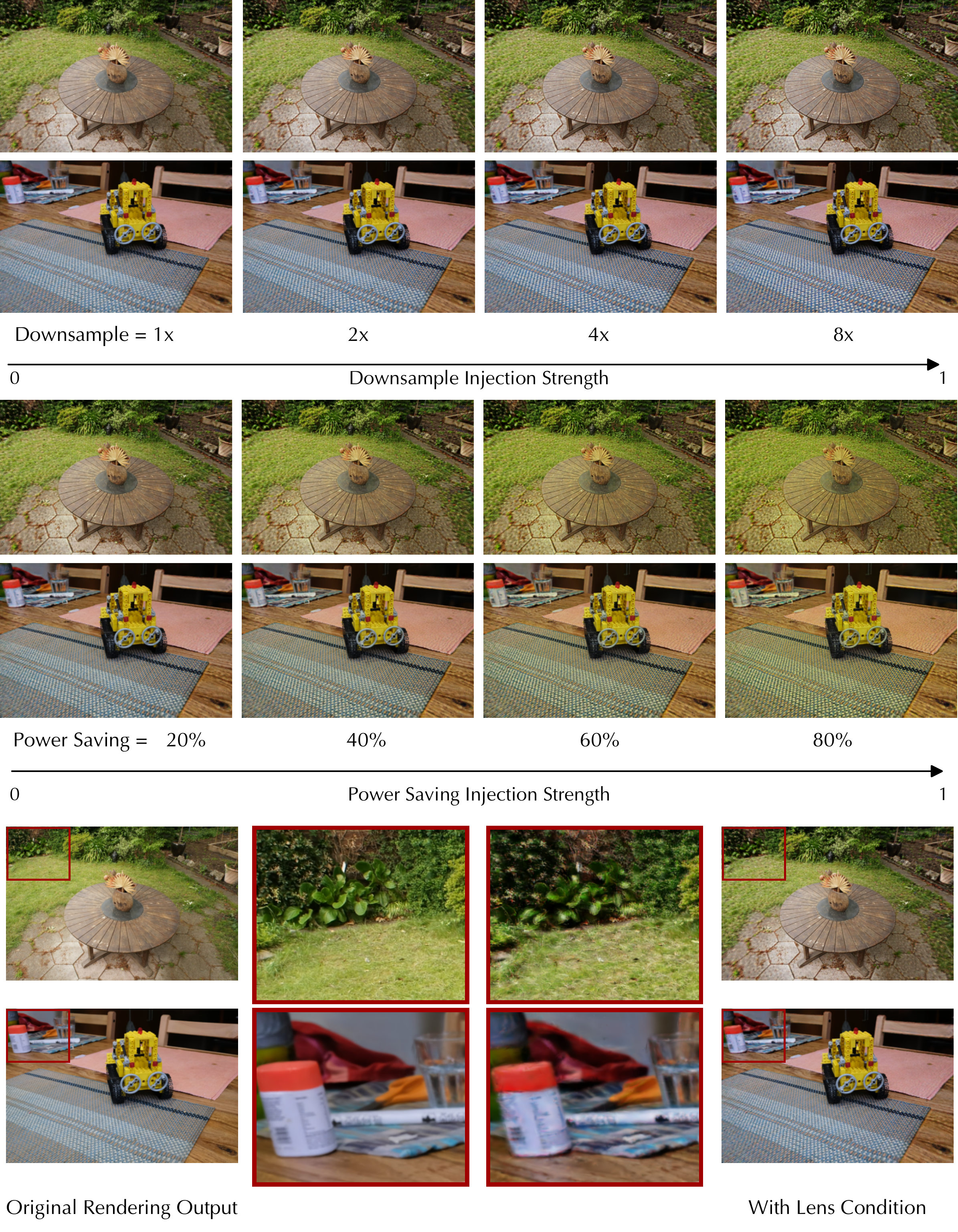}
\caption{
Additional condition-response visualization on Garden and Kitchen.
}
\Description{Additional condition-response visualization combining Garden and Kitchen scenes.}
\label{fig:supp:condition-response-garden-kitchen}
\end{figure*}

\begin{figure*}[t]
\centering
\includegraphics[width=\linewidth,height=0.94\textheight,keepaspectratio]{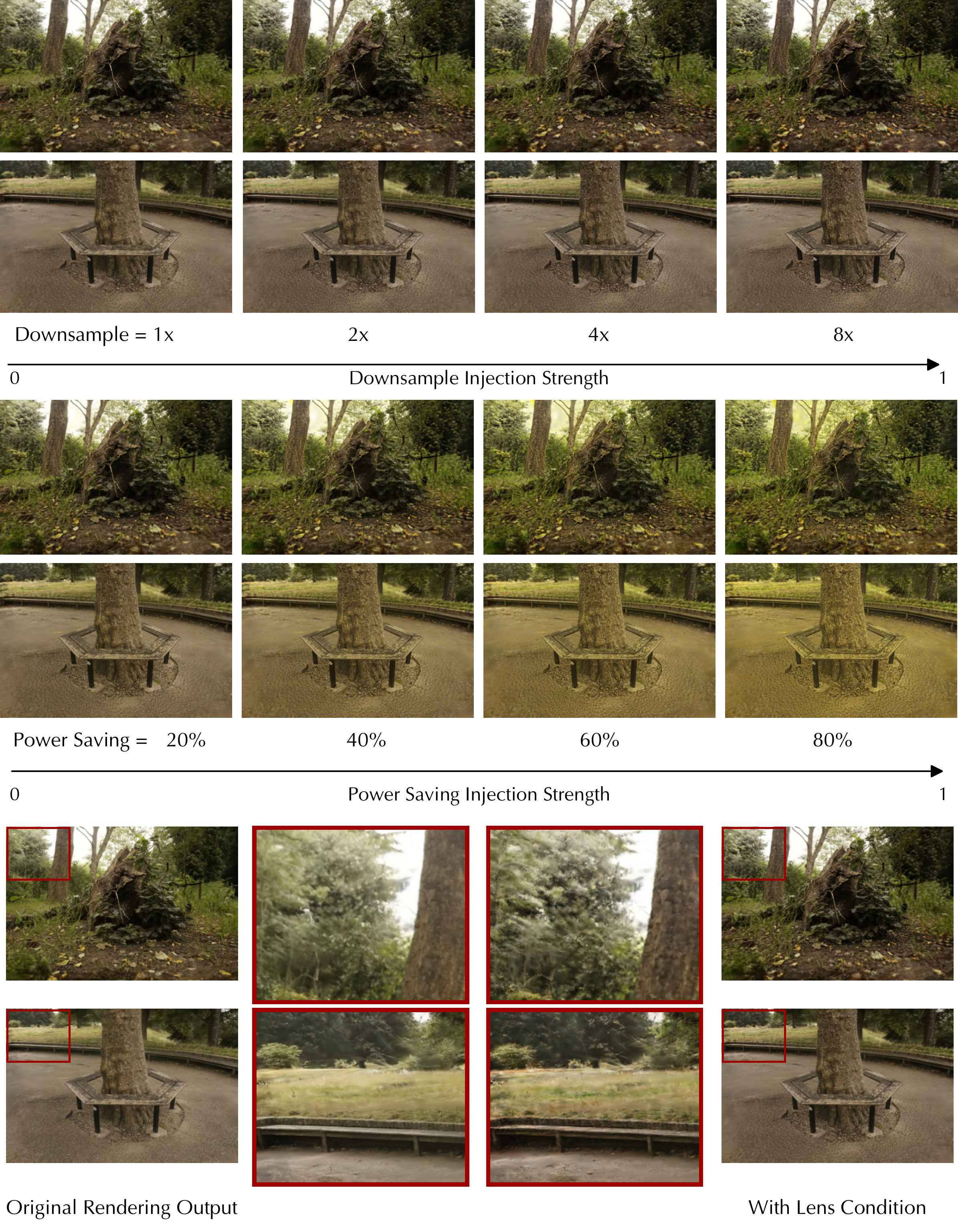}
\caption{
Additional condition-response visualization on Stump and Treehill.
}
\Description{Additional condition-response visualization combining Stump and Treehill scenes.}
\label{fig:supp:condition-response-stump-treehill}
\end{figure*}

\begin{figure*}[t]
\centering
\includegraphics[width=\linewidth,height=0.94\textheight,keepaspectratio]{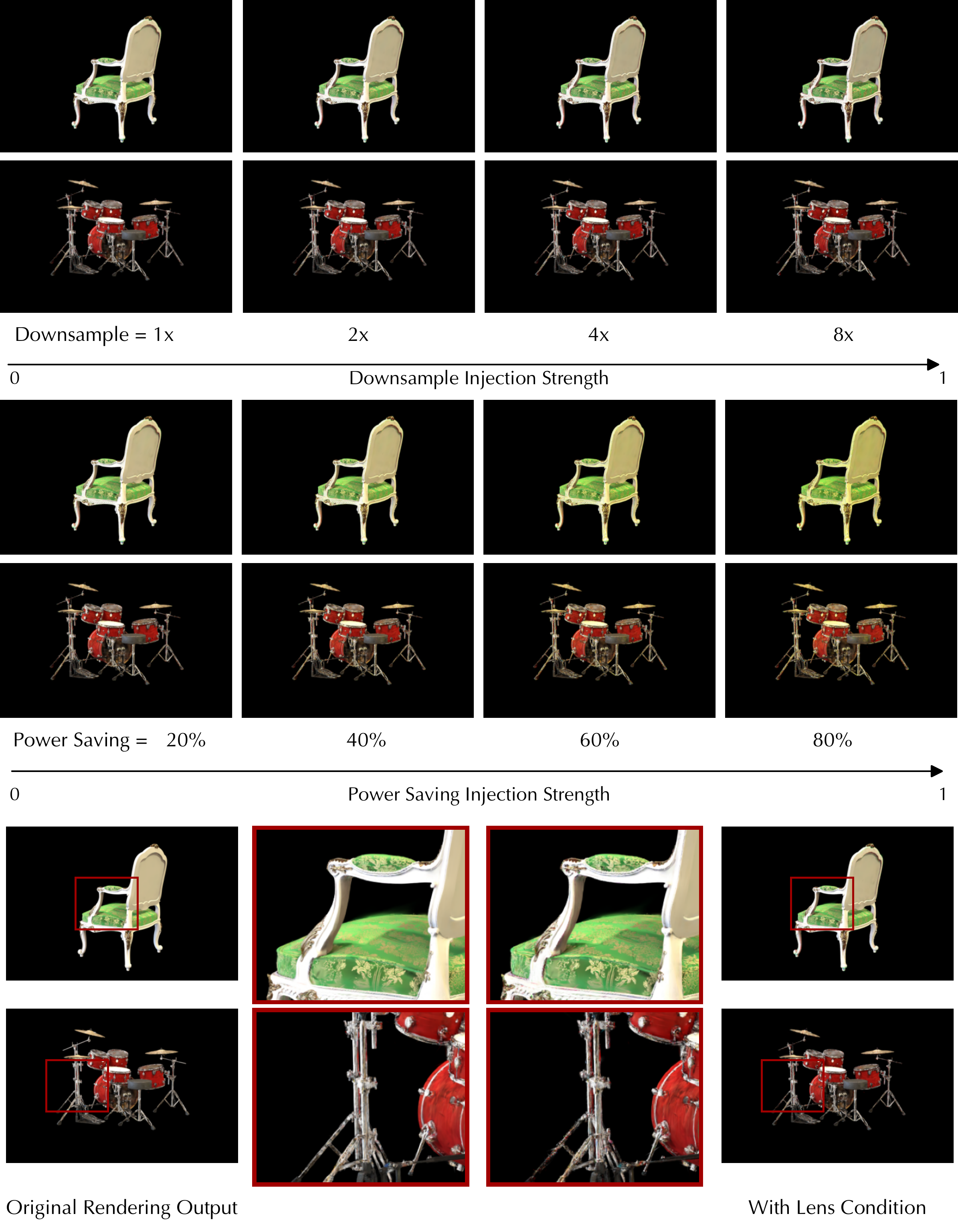}
\caption{
Additional condition-response visualization on Chair and Drums.
}
\Description{Additional condition-response visualization combining Chair and Drums scenes.}
\label{fig:supp:condition-response-chair-drums}
\end{figure*}

\begin{figure*}[t]
\centering
\includegraphics[width=\linewidth,height=0.94\textheight,keepaspectratio]{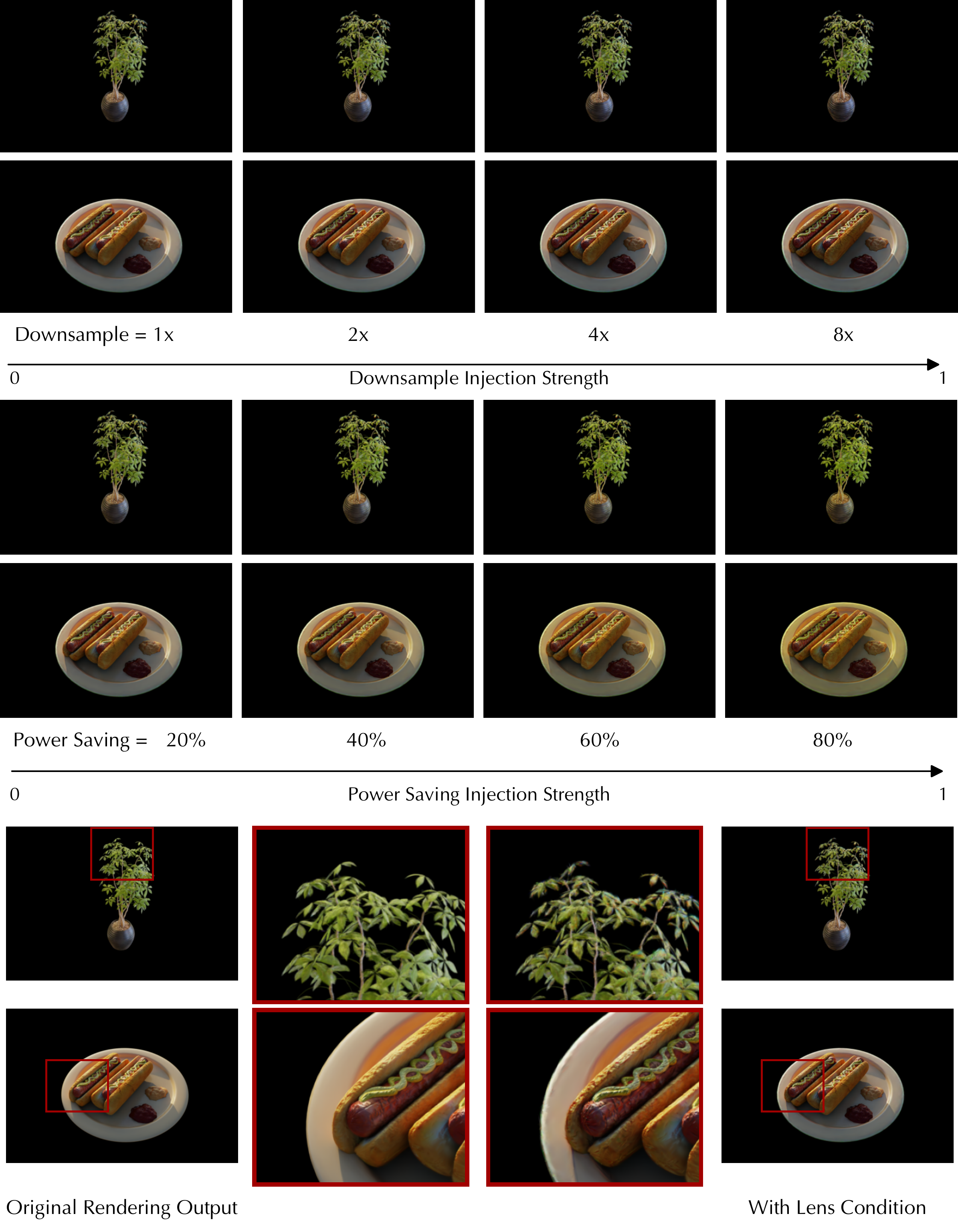}
\caption{
Additional condition-response visualization on Ficus and Hotdog.
}
\Description{Additional condition-response visualization combining Ficus and Hotdog scenes.}
\label{fig:supp:condition-response-ficus-hotdog}
\end{figure*}

\begin{figure*}[t]
\centering
\includegraphics[width=\linewidth,height=0.94\textheight,keepaspectratio]{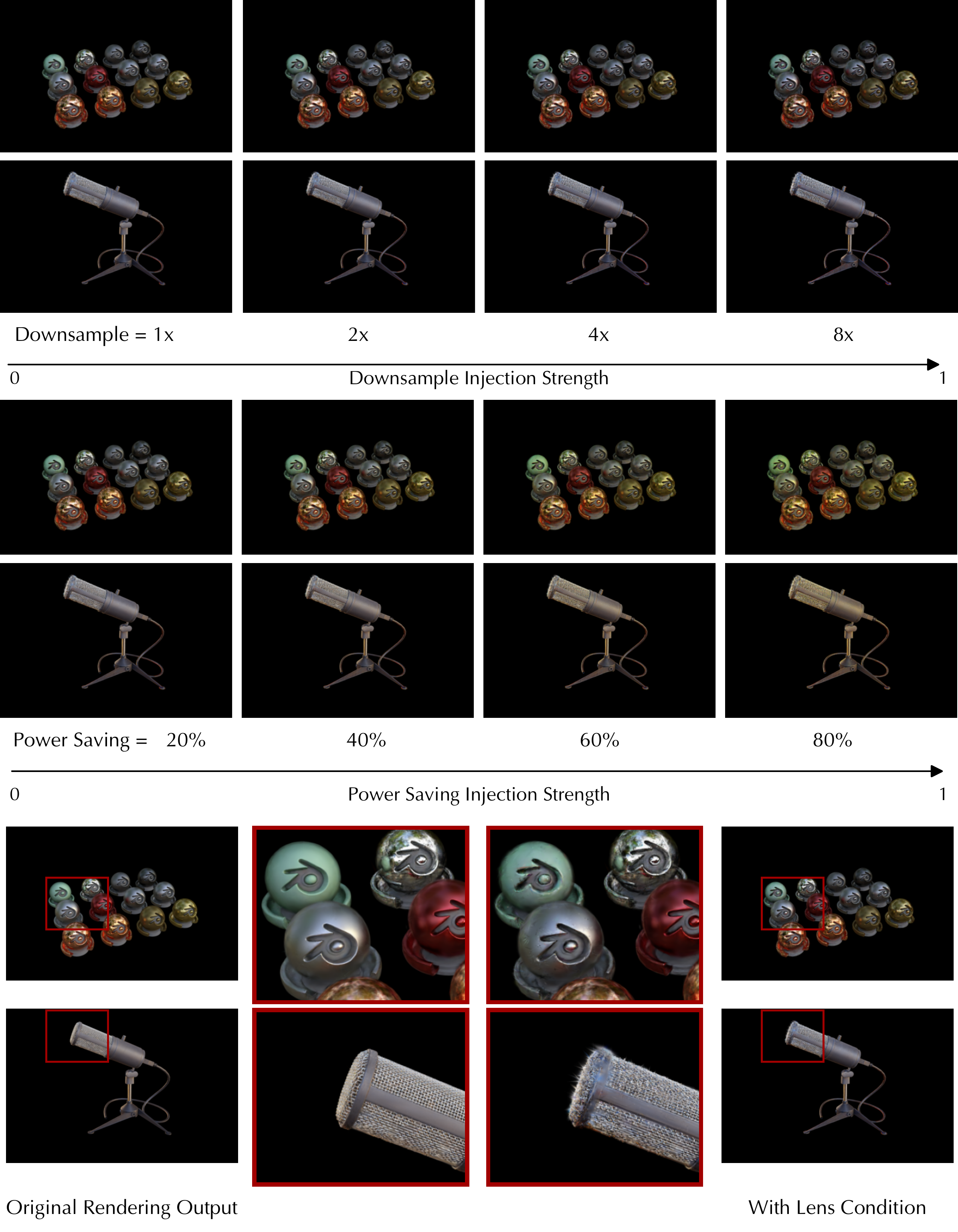}
\caption{
Additional condition-response visualization on Materials and Mic.
}
\Description{Additional condition-response visualization combining Materials and Mic scenes.}
\label{fig:supp:condition-response-materials-mic}
\end{figure*}

\Fig{fig:supp:condition-response}--\Fig{fig:supp:condition-response-materials-mic}
provide additional condition-response visualizations across more scenes.
Each figure shows raw rendered outputs before the downstream pipeline.
From top to bottom, the groups correspond to anti-aliasing resampling, display mapping, and lens correction.
Within each group, we vary the selected condition while disabling the others.
The annotations indicate the corresponding run-time  state.
For lens correction, the highlighted peripheral crops show that stronger lens injection enhances high-frequency details to compensate for the low-pass blur introduced by the optics.

\newpage

\newpage

\begin{figure*}[t]
  \centering
  \includegraphics[width=\linewidth]{sections/figure/dynamics.pdf}
\caption{
Examples of dynamic runtime states.
Top: as the camera moves farther from an object, the object occupies fewer pixels on the image plane, reducing its effective screen-space sampling rate.
Bottom: the same Gaussian primitives can project to different screen-space locations under different camera poses.
Because lens aberrations vary spatially, the same primitives should render differently at different projected locations to compensate for location-dependent optical aberrations.
This requires dynamically changing the aberration-related state during rendering.
}
  \label{fig:supp:runtime-state-examples}
\end{figure*}

\newpage

\begin{table*}[t]
\centering
\caption{
End-to-end quality across neural Gaussian backbones after the XR post-processing pipeline.
Each cell averages all test views and all $20$ resolution-power run-time  settings.
Rows without a prefix report the original backbone baseline, \textsc{SS} rows use the supersampling-enhanced baseline, and \textsc{Control} rows further add learned condition-injection adapters to the same frozen backbone.
\proj consistently improves end-to-end quality across backbones.
}
\label{tab:supp:additional_bb}
\definecolor{xrgsGroupShade}{RGB}{235,241,248}
\definecolor{xrgsSSShade}{RGB}{255,249,224}
\definecolor{xrgsControlShade}{RGB}{255,239,224}
\setlength{\tabcolsep}{3.5pt}
\renewcommand{\arraystretch}{1.08}
\resizebox{0.5\linewidth}{!}{%
\begin{tabular}{lcccccc}
\toprule
Method / Dataset & \multicolumn{3}{c}{{Mip-NeRF~360}} & \multicolumn{3}{c}{{NeRF Synthetic}} \\
\cmidrule(lr){2-4}\cmidrule(lr){5-7}
& PSNR$\uparrow$ & SSIM$\uparrow$ & LPIPS$\downarrow$ & PSNR$\uparrow$ & SSIM$\uparrow$ & LPIPS$\downarrow$ \\
\midrule
\rowcolor{xrgsGroupShade}
\multicolumn{7}{l}{\textbf{\textsc{run-time -fixed Gaussian Splatting}}} \\
3DGS & 18.15 & 0.478 & 0.579 & 21.96 & 0.840 & 0.166 \\
\rowcolor{xrgsSSShade}
SS-3DGS & \textbf{18.93} & \textbf{0.496} & \textbf{0.558} & \textbf{23.36} & \textbf{0.869} & \textbf{0.146} \\
\specialrule{0.25pt}{1.0pt}{1.0pt}
Mip-Splatting & 18.87 & 0.493 & 0.564 & 23.17 & 0.862 & 0.152 \\
\rowcolor{xrgsSSShade}
SS-Mip-Splatting & \textbf{18.93} & \textbf{0.497} & \textbf{0.557} & \textbf{23.37} & \textbf{0.870} & \textbf{0.145} \\
\midrule
\rowcolor{xrgsGroupShade}
\multicolumn{7}{l}{\textbf{\textsc{Neural Gaussian Splatting}}} \\
CompGS & 18.26 & 0.479 & 0.577 & 21.47 & 0.836 & 0.173 \\
\rowcolor{xrgsSSShade}
SS-CompGS & 18.93 & 0.495 & 0.559 & 22.60 & 0.861 & 0.156 \\
\rowcolor{xrgsControlShade}
Control-CompGS & \textbf{19.29} & \textbf{0.522} & \textbf{0.532} & \textbf{22.95} & \textbf{0.864} & \textbf{0.147} \\
\specialrule{0.25pt}{1.0pt}{1.0pt}
ContextGS & 18.28 & 0.480 & 0.576 & 22.03 & 0.840 & 0.165 \\
\rowcolor{xrgsSSShade}
SS-ContextGS & 18.96 & 0.496 & 0.558 & 23.37 & 0.869 & 0.146 \\
\rowcolor{xrgsControlShade}
Control-ContextGS & \textbf{19.33} & \textbf{0.524} & \textbf{0.529} & \textbf{23.60} & \textbf{0.872} & \textbf{0.137} \\
\specialrule{0.25pt}{1.0pt}{1.0pt}
HAC{++} & 18.28 & 0.480 & 0.575 & 22.03 & 0.841 & 0.165 \\
\rowcolor{xrgsSSShade}
SS-HAC{++} & 18.96 & 0.496 & 0.558 & 23.36 & 0.869 & 0.146 \\
\rowcolor{xrgsControlShade}
Control-HAC{++} & \textbf{19.32} & \textbf{0.523} & \textbf{0.530} & \textbf{23.60} & \textbf{0.872} & \textbf{0.136} \\
\specialrule{0.25pt}{1.0pt}{1.0pt}
Octree-GS & 18.04 & 0.477 & 0.581 & 21.98 & 0.841 & 0.168 \\
\rowcolor{xrgsSSShade}
SS-Octree-GS & 18.96 & 0.496 & 0.558 & 23.29 & 0.868 & 0.147 \\
\rowcolor{xrgsControlShade}
Control-Octree-GS & \textbf{19.32} & \textbf{0.524} & \textbf{0.529} & \textbf{23.52} & \textbf{0.870} & \textbf{0.138} \\
\specialrule{0.25pt}{1.0pt}{1.0pt}
Scaffold-GS & 18.19 & 0.479 & 0.578 & 21.94 & 0.839 & 0.167 \\
\rowcolor{xrgsSSShade}
SS-Scaffold-GS & 18.96 & 0.496 & 0.558 & 23.35 & 0.869 & 0.147 \\
\rowcolor{xrgsControlShade}
Control-Scaffold-GS & \textbf{19.33} & \textbf{0.524} & \textbf{0.528} & \textbf{23.59} & \textbf{0.871} & \textbf{0.137} \\
\bottomrule
\end{tabular}%
}
\end{table*}

\begin{table*}[t]
\centering
\caption{Per-scene end-to-end PSNR and SSIM on Mip-NeRF~360.
Each scene cell averages all test views and all 20 resolution-power settings.
Rows without prefix report the original backbone baseline without supersampling.
\textsc{SS} rows use the enhanced supersampling baseline.
\textsc{Control} rows add learned condition-injection adapters.}
\label{tab:supp:xrgs-e2e-quality-per-scene-mip360}
\definecolor{xrgsSSShade}{RGB}{255,249,224}
\definecolor{xrgsControlShade}{RGB}{255,239,224}
\tiny
\setlength{\tabcolsep}{1.7pt}
\renewcommand{\arraystretch}{0.78}
\resizebox{0.85\textwidth}{!}{%
\begin{tabular}{llcccccccccc}
\toprule
\multirow{2}{*}{Metric} & \multirow{2}{*}{Method} & \multicolumn{9}{c}{Mip-NeRF~360} & \multirow{2}{*}{Avg} \\
\cmidrule(lr){3-11}
& & Bicycle & Bonsai & Counter & Flowers & Garden & Kitchen & Room & Stump & Treehill & \\
\midrule
\multirow{19}{*}{PSNR$\uparrow$} & 3DGS & 17.32 & 19.79 & 19.06 & 15.51 & 17.08 & 17.93 & 20.09 & 19.46 & 17.11 & 18.15 \\
 & \cellcolor{xrgsSSShade}SS-3DGS & \cellcolor{xrgsSSShade}\textbf{18.14} & \cellcolor{xrgsSSShade}\textbf{20.86} & \cellcolor{xrgsSSShade}\textbf{19.72} & \cellcolor{xrgsSSShade}\textbf{16.48} & \cellcolor{xrgsSSShade}\textbf{17.93} & \cellcolor{xrgsSSShade}\textbf{18.50} & \cellcolor{xrgsSSShade}\textbf{20.74} & \cellcolor{xrgsSSShade}\textbf{20.27} & \cellcolor{xrgsSSShade}\textbf{17.77} & \cellcolor{xrgsSSShade}\textbf{18.93} \\
\cmidrule(lr){2-12}
 & Mip-Splatting & \textbf{18.17} & 20.68 & 19.62 & 16.42 & 17.88 & 18.41 & 20.58 & 20.25 & \textbf{17.81} & 18.87 \\
 & \cellcolor{xrgsSSShade}SS-Mip-Splatting & \cellcolor{xrgsSSShade}18.15 & \cellcolor{xrgsSSShade}\textbf{20.84} & \cellcolor{xrgsSSShade}\textbf{19.73} & \cellcolor{xrgsSSShade}\textbf{16.51} & \cellcolor{xrgsSSShade}\textbf{17.93} & \cellcolor{xrgsSSShade}\textbf{18.51} & \cellcolor{xrgsSSShade}\textbf{20.69} & \cellcolor{xrgsSSShade}\textbf{20.29} & \cellcolor{xrgsSSShade}17.75 & \cellcolor{xrgsSSShade}\textbf{18.93} \\
\cmidrule(lr){2-12}
 & CompGS & 17.40 & 19.96 & 19.11 & 15.69 & 17.15 & 17.93 & 20.10 & 19.57 & 17.42 & 18.26 \\
 & \cellcolor{xrgsSSShade}SS-CompGS & \cellcolor{xrgsSSShade}18.13 & \cellcolor{xrgsSSShade}20.90 & \cellcolor{xrgsSSShade}19.72 & \cellcolor{xrgsSSShade}16.47 & \cellcolor{xrgsSSShade}17.95 & \cellcolor{xrgsSSShade}18.48 & \cellcolor{xrgsSSShade}20.69 & \cellcolor{xrgsSSShade}20.22 & \cellcolor{xrgsSSShade}17.78 & \cellcolor{xrgsSSShade}18.93 \\
 & \cellcolor{xrgsControlShade}Control-CompGS & \cellcolor{xrgsControlShade}\textbf{18.36} & \cellcolor{xrgsControlShade}\textbf{21.54} & \cellcolor{xrgsControlShade}\textbf{20.18} & \cellcolor{xrgsControlShade}\textbf{16.62} & \cellcolor{xrgsControlShade}\textbf{18.30} & \cellcolor{xrgsControlShade}\textbf{18.97} & \cellcolor{xrgsControlShade}\textbf{21.18} & \cellcolor{xrgsControlShade}\textbf{20.38} & \cellcolor{xrgsControlShade}\textbf{18.07} & \cellcolor{xrgsControlShade}\textbf{19.29} \\
\cmidrule(lr){2-12}
 & ContextGS & 17.39 & 19.88 & 19.10 & 15.81 & 17.20 & 17.97 & 20.17 & 19.52 & 17.43 & 18.28 \\
 & \cellcolor{xrgsSSShade}SS-ContextGS & \cellcolor{xrgsSSShade}18.14 & \cellcolor{xrgsSSShade}20.96 & \cellcolor{xrgsSSShade}19.75 & \cellcolor{xrgsSSShade}16.51 & \cellcolor{xrgsSSShade}17.95 & \cellcolor{xrgsSSShade}18.50 & \cellcolor{xrgsSSShade}20.74 & \cellcolor{xrgsSSShade}20.25 & \cellcolor{xrgsSSShade}17.83 & \cellcolor{xrgsSSShade}18.96 \\
 & \cellcolor{xrgsControlShade}Control-ContextGS & \cellcolor{xrgsControlShade}\textbf{18.38} & \cellcolor{xrgsControlShade}\textbf{21.63} & \cellcolor{xrgsControlShade}\textbf{20.22} & \cellcolor{xrgsControlShade}\textbf{16.65} & \cellcolor{xrgsControlShade}\textbf{18.30} & \cellcolor{xrgsControlShade}\textbf{19.01} & \cellcolor{xrgsControlShade}\textbf{21.22} & \cellcolor{xrgsControlShade}\textbf{20.42} & \cellcolor{xrgsControlShade}\textbf{18.10} & \cellcolor{xrgsControlShade}\textbf{19.33} \\
\cmidrule(lr){2-12}
 & HAC{++} & 17.45 & 19.95 & 19.14 & 15.79 & 17.10 & 17.93 & 20.22 & 19.58 & 17.41 & 18.28 \\
 & \cellcolor{xrgsSSShade}SS-HAC{++} & \cellcolor{xrgsSSShade}18.15 & \cellcolor{xrgsSSShade}20.95 & \cellcolor{xrgsSSShade}19.73 & \cellcolor{xrgsSSShade}16.52 & \cellcolor{xrgsSSShade}17.94 & \cellcolor{xrgsSSShade}18.50 & \cellcolor{xrgsSSShade}20.76 & \cellcolor{xrgsSSShade}20.27 & \cellcolor{xrgsSSShade}17.81 & \cellcolor{xrgsSSShade}18.96 \\
 & \cellcolor{xrgsControlShade}Control-HAC{++} & \cellcolor{xrgsControlShade}\textbf{18.39} & \cellcolor{xrgsControlShade}\textbf{21.64} & \cellcolor{xrgsControlShade}\textbf{20.21} & \cellcolor{xrgsControlShade}\textbf{16.64} & \cellcolor{xrgsControlShade}\textbf{18.30} & \cellcolor{xrgsControlShade}\textbf{19.00} & \cellcolor{xrgsControlShade}\textbf{21.24} & \cellcolor{xrgsControlShade}\textbf{20.42} & \cellcolor{xrgsControlShade}\textbf{18.09} & \cellcolor{xrgsControlShade}\textbf{19.32} \\
\cmidrule(lr){2-12}
 & Octree-GS & 17.44 & 19.61 & 18.74 & 15.67 & 16.82 & 17.49 & 19.66 & 19.57 & 17.35 & 18.04 \\
 & \cellcolor{xrgsSSShade}SS-Octree-GS & \cellcolor{xrgsSSShade}18.16 & \cellcolor{xrgsSSShade}20.92 & \cellcolor{xrgsSSShade}19.79 & \cellcolor{xrgsSSShade}16.50 & \cellcolor{xrgsSSShade}17.95 & \cellcolor{xrgsSSShade}18.50 & \cellcolor{xrgsSSShade}20.77 & \cellcolor{xrgsSSShade}20.26 & \cellcolor{xrgsSSShade}17.81 & \cellcolor{xrgsSSShade}18.96 \\
 & \cellcolor{xrgsControlShade}Control-Octree-GS & \cellcolor{xrgsControlShade}\textbf{18.39} & \cellcolor{xrgsControlShade}\textbf{21.58} & \cellcolor{xrgsControlShade}\textbf{20.23} & \cellcolor{xrgsControlShade}\textbf{16.63} & \cellcolor{xrgsControlShade}\textbf{18.31} & \cellcolor{xrgsControlShade}\textbf{19.00} & \cellcolor{xrgsControlShade}\textbf{21.26} & \cellcolor{xrgsControlShade}\textbf{20.42} & \cellcolor{xrgsControlShade}\textbf{18.08} & \cellcolor{xrgsControlShade}\textbf{19.32} \\
\cmidrule(lr){2-12}
 & Scaffold-GS & 17.43 & 19.77 & 19.09 & 15.61 & 16.96 & 17.88 & 20.05 & 19.57 & 17.32 & 18.19 \\
 & \cellcolor{xrgsSSShade}SS-Scaffold-GS & \cellcolor{xrgsSSShade}18.14 & \cellcolor{xrgsSSShade}20.94 & \cellcolor{xrgsSSShade}19.76 & \cellcolor{xrgsSSShade}16.51 & \cellcolor{xrgsSSShade}17.93 & \cellcolor{xrgsSSShade}18.51 & \cellcolor{xrgsSSShade}20.78 & \cellcolor{xrgsSSShade}20.26 & \cellcolor{xrgsSSShade}17.82 & \cellcolor{xrgsSSShade}18.96 \\
 & \cellcolor{xrgsControlShade}Control-Scaffold-GS & \cellcolor{xrgsControlShade}\textbf{18.38} & \cellcolor{xrgsControlShade}\textbf{21.63} & \cellcolor{xrgsControlShade}\textbf{20.23} & \cellcolor{xrgsControlShade}\textbf{16.64} & \cellcolor{xrgsControlShade}\textbf{18.29} & \cellcolor{xrgsControlShade}\textbf{19.01} & \cellcolor{xrgsControlShade}\textbf{21.27} & \cellcolor{xrgsControlShade}\textbf{20.43} & \cellcolor{xrgsControlShade}\textbf{18.09} & \cellcolor{xrgsControlShade}\textbf{19.33} \\
\cmidrule(lr){1-12}
\multirow{19}{*}{SSIM$\uparrow$} & 3DGS & 0.340 & 0.665 & 0.646 & 0.283 & 0.346 & 0.531 & 0.713 & 0.393 & 0.382 & 0.478 \\
 & \cellcolor{xrgsSSShade}SS-3DGS & \cellcolor{xrgsSSShade}\textbf{0.360} & \cellcolor{xrgsSSShade}\textbf{0.684} & \cellcolor{xrgsSSShade}\textbf{0.660} & \cellcolor{xrgsSSShade}\textbf{0.312} & \cellcolor{xrgsSSShade}\textbf{0.366} & \cellcolor{xrgsSSShade}\textbf{0.548} & \cellcolor{xrgsSSShade}\textbf{0.723} & \cellcolor{xrgsSSShade}\textbf{0.420} & \cellcolor{xrgsSSShade}\textbf{0.393} & \cellcolor{xrgsSSShade}\textbf{0.496} \\
\cmidrule(lr){2-12}
 & Mip-Splatting & 0.361 & 0.679 & 0.656 & 0.305 & 0.361 & 0.543 & 0.720 & 0.416 & 0.391 & 0.493 \\
 & \cellcolor{xrgsSSShade}SS-Mip-Splatting & \cellcolor{xrgsSSShade}\textbf{0.361} & \cellcolor{xrgsSSShade}\textbf{0.684} & \cellcolor{xrgsSSShade}\textbf{0.660} & \cellcolor{xrgsSSShade}\textbf{0.316} & \cellcolor{xrgsSSShade}\textbf{0.366} & \cellcolor{xrgsSSShade}\textbf{0.549} & \cellcolor{xrgsSSShade}\textbf{0.723} & \cellcolor{xrgsSSShade}\textbf{0.422} & \cellcolor{xrgsSSShade}\textbf{0.393} & \cellcolor{xrgsSSShade}\textbf{0.497} \\
\cmidrule(lr){2-12}
 & CompGS & 0.341 & 0.666 & 0.648 & 0.284 & 0.349 & 0.531 & 0.711 & 0.395 & 0.384 & 0.479 \\
 & \cellcolor{xrgsSSShade}SS-CompGS & \cellcolor{xrgsSSShade}0.360 & \cellcolor{xrgsSSShade}0.683 & \cellcolor{xrgsSSShade}0.660 & \cellcolor{xrgsSSShade}0.310 & \cellcolor{xrgsSSShade}0.366 & \cellcolor{xrgsSSShade}0.548 & \cellcolor{xrgsSSShade}0.719 & \cellcolor{xrgsSSShade}0.418 & \cellcolor{xrgsSSShade}0.392 & \cellcolor{xrgsSSShade}0.495 \\
 & \cellcolor{xrgsControlShade}Control-CompGS & \cellcolor{xrgsControlShade}\textbf{0.378} & \cellcolor{xrgsControlShade}\textbf{0.727} & \cellcolor{xrgsControlShade}\textbf{0.703} & \cellcolor{xrgsControlShade}\textbf{0.321} & \cellcolor{xrgsControlShade}\textbf{0.389} & \cellcolor{xrgsControlShade}\textbf{0.573} & \cellcolor{xrgsControlShade}\textbf{0.760} & \cellcolor{xrgsControlShade}\textbf{0.433} & \cellcolor{xrgsControlShade}\textbf{0.413} & \cellcolor{xrgsControlShade}\textbf{0.522} \\
\cmidrule(lr){2-12}
 & ContextGS & 0.339 & 0.668 & 0.648 & 0.288 & 0.350 & 0.532 & 0.712 & 0.396 & 0.386 & 0.480 \\
 & \cellcolor{xrgsSSShade}SS-ContextGS & \cellcolor{xrgsSSShade}0.358 & \cellcolor{xrgsSSShade}0.684 & \cellcolor{xrgsSSShade}0.660 & \cellcolor{xrgsSSShade}0.311 & \cellcolor{xrgsSSShade}0.366 & \cellcolor{xrgsSSShade}0.548 & \cellcolor{xrgsSSShade}0.721 & \cellcolor{xrgsSSShade}0.420 & \cellcolor{xrgsSSShade}0.393 & \cellcolor{xrgsSSShade}0.496 \\
 & \cellcolor{xrgsControlShade}Control-ContextGS & \cellcolor{xrgsControlShade}\textbf{0.380} & \cellcolor{xrgsControlShade}\textbf{0.730} & \cellcolor{xrgsControlShade}\textbf{0.704} & \cellcolor{xrgsControlShade}\textbf{0.322} & \cellcolor{xrgsControlShade}\textbf{0.390} & \cellcolor{xrgsControlShade}\textbf{0.574} & \cellcolor{xrgsControlShade}\textbf{0.761} & \cellcolor{xrgsControlShade}\textbf{0.436} & \cellcolor{xrgsControlShade}\textbf{0.416} & \cellcolor{xrgsControlShade}\textbf{0.524} \\
\cmidrule(lr){2-12}
 & HAC{++} & 0.344 & 0.668 & 0.648 & 0.288 & 0.349 & 0.532 & 0.712 & 0.397 & 0.386 & 0.480 \\
 & \cellcolor{xrgsSSShade}SS-HAC{++} & \cellcolor{xrgsSSShade}0.361 & \cellcolor{xrgsSSShade}0.684 & \cellcolor{xrgsSSShade}0.660 & \cellcolor{xrgsSSShade}0.311 & \cellcolor{xrgsSSShade}0.366 & \cellcolor{xrgsSSShade}0.548 & \cellcolor{xrgsSSShade}0.722 & \cellcolor{xrgsSSShade}0.420 & \cellcolor{xrgsSSShade}0.392 & \cellcolor{xrgsSSShade}0.496 \\
 & \cellcolor{xrgsControlShade}Control-HAC{++} & \cellcolor{xrgsControlShade}\textbf{0.380} & \cellcolor{xrgsControlShade}\textbf{0.730} & \cellcolor{xrgsControlShade}\textbf{0.704} & \cellcolor{xrgsControlShade}\textbf{0.322} & \cellcolor{xrgsControlShade}\textbf{0.389} & \cellcolor{xrgsControlShade}\textbf{0.574} & \cellcolor{xrgsControlShade}\textbf{0.761} & \cellcolor{xrgsControlShade}\textbf{0.436} & \cellcolor{xrgsControlShade}\textbf{0.415} & \cellcolor{xrgsControlShade}\textbf{0.523} \\
\cmidrule(lr){2-12}
 & Octree-GS & 0.345 & 0.661 & 0.638 & 0.287 & 0.345 & 0.522 & 0.708 & 0.398 & 0.386 & 0.477 \\
 & \cellcolor{xrgsSSShade}SS-Octree-GS & \cellcolor{xrgsSSShade}0.361 & \cellcolor{xrgsSSShade}0.684 & \cellcolor{xrgsSSShade}0.661 & \cellcolor{xrgsSSShade}0.312 & \cellcolor{xrgsSSShade}0.366 & \cellcolor{xrgsSSShade}0.548 & \cellcolor{xrgsSSShade}0.723 & \cellcolor{xrgsSSShade}0.420 & \cellcolor{xrgsSSShade}0.393 & \cellcolor{xrgsSSShade}0.496 \\
 & \cellcolor{xrgsControlShade}Control-Octree-GS & \cellcolor{xrgsControlShade}\textbf{0.381} & \cellcolor{xrgsControlShade}\textbf{0.727} & \cellcolor{xrgsControlShade}\textbf{0.705} & \cellcolor{xrgsControlShade}\textbf{0.324} & \cellcolor{xrgsControlShade}\textbf{0.390} & \cellcolor{xrgsControlShade}\textbf{0.573} & \cellcolor{xrgsControlShade}\textbf{0.762} & \cellcolor{xrgsControlShade}\textbf{0.436} & \cellcolor{xrgsControlShade}\textbf{0.416} & \cellcolor{xrgsControlShade}\textbf{0.524} \\
\cmidrule(lr){2-12}
 & Scaffold-GS & 0.342 & 0.665 & 0.648 & 0.285 & 0.346 & 0.531 & 0.714 & 0.397 & 0.385 & 0.479 \\
 & \cellcolor{xrgsSSShade}SS-Scaffold-GS & \cellcolor{xrgsSSShade}0.361 & \cellcolor{xrgsSSShade}0.684 & \cellcolor{xrgsSSShade}0.661 & \cellcolor{xrgsSSShade}0.311 & \cellcolor{xrgsSSShade}0.366 & \cellcolor{xrgsSSShade}0.548 & \cellcolor{xrgsSSShade}0.723 & \cellcolor{xrgsSSShade}0.420 & \cellcolor{xrgsSSShade}0.393 & \cellcolor{xrgsSSShade}0.496 \\
 & \cellcolor{xrgsControlShade}Control-Scaffold-GS & \cellcolor{xrgsControlShade}\textbf{0.380} & \cellcolor{xrgsControlShade}\textbf{0.730} & \cellcolor{xrgsControlShade}\textbf{0.705} & \cellcolor{xrgsControlShade}\textbf{0.323} & \cellcolor{xrgsControlShade}\textbf{0.390} & \cellcolor{xrgsControlShade}\textbf{0.574} & \cellcolor{xrgsControlShade}\textbf{0.762} & \cellcolor{xrgsControlShade}\textbf{0.436} & \cellcolor{xrgsControlShade}\textbf{0.415} & \cellcolor{xrgsControlShade}\textbf{0.524} \\
\bottomrule
\end{tabular}%
}
\end{table*}

\begin{table*}[p]
\centering
\caption{Per-scene end-to-end LPIPS on Mip-NeRF~360.
Each scene cell averages all test views and all 20 resolution-power settings.
Rows without prefix report the original backbone baseline without supersampling.
\textsc{SS} rows use the enhanced supersampling baseline.
\textsc{Control} rows add learned condition-injection adapters.}
\label{tab:supp:xrgs-e2e-quality-per-scene-mip360-lpips}
\definecolor{xrgsSSShade}{RGB}{255,249,224}
\definecolor{xrgsControlShade}{RGB}{255,239,224}
\tiny
\setlength{\tabcolsep}{1.7pt}
\renewcommand{\arraystretch}{0.78}
\resizebox{0.85\textwidth}{!}{%
\begin{tabular}{llcccccccccc}
\toprule
\multirow{2}{*}{Metric} & \multirow{2}{*}{Method} & \multicolumn{9}{c}{Mip-NeRF~360} & \multirow{2}{*}{Avg} \\
\cmidrule(lr){3-11}
& & Bicycle & Bonsai & Counter & Flowers & Garden & Kitchen & Room & Stump & Treehill & \\
\midrule
\multirow{19}{*}{LPIPS$\downarrow$} & 3DGS & 0.636 & 0.494 & 0.500 & 0.655 & 0.648 & 0.543 & 0.479 & 0.613 & 0.643 & 0.579 \\
 & \cellcolor{xrgsSSShade}SS-3DGS & \cellcolor{xrgsSSShade}\textbf{0.612} & \cellcolor{xrgsSSShade}\textbf{0.479} & \cellcolor{xrgsSSShade}\textbf{0.484} & \cellcolor{xrgsSSShade}\textbf{0.621} & \cellcolor{xrgsSSShade}\textbf{0.625} & \cellcolor{xrgsSSShade}\textbf{0.525} & \cellcolor{xrgsSSShade}\textbf{0.467} & \cellcolor{xrgsSSShade}\textbf{0.583} & \cellcolor{xrgsSSShade}\textbf{0.621} & \cellcolor{xrgsSSShade}\textbf{0.558} \\
\cmidrule(lr){2-12}
 & Mip-Splatting & 0.616 & 0.484 & 0.489 & 0.633 & 0.632 & 0.532 & 0.471 & 0.590 & 0.626 & 0.564 \\
 & \cellcolor{xrgsSSShade}SS-Mip-Splatting & \cellcolor{xrgsSSShade}\textbf{0.611} & \cellcolor{xrgsSSShade}\textbf{0.479} & \cellcolor{xrgsSSShade}\textbf{0.484} & \cellcolor{xrgsSSShade}\textbf{0.619} & \cellcolor{xrgsSSShade}\textbf{0.625} & \cellcolor{xrgsSSShade}\textbf{0.524} & \cellcolor{xrgsSSShade}\textbf{0.468} & \cellcolor{xrgsSSShade}\textbf{0.581} & \cellcolor{xrgsSSShade}\textbf{0.621} & \cellcolor{xrgsSSShade}\textbf{0.557} \\
\cmidrule(lr){2-12}
 & CompGS & 0.634 & 0.495 & 0.501 & 0.652 & 0.644 & 0.544 & 0.480 & 0.610 & 0.636 & 0.577 \\
 & \cellcolor{xrgsSSShade}SS-CompGS & \cellcolor{xrgsSSShade}0.614 & \cellcolor{xrgsSSShade}0.481 & \cellcolor{xrgsSSShade}0.487 & \cellcolor{xrgsSSShade}0.623 & \cellcolor{xrgsSSShade}0.626 & \cellcolor{xrgsSSShade}0.526 & \cellcolor{xrgsSSShade}0.469 & \cellcolor{xrgsSSShade}0.587 & \cellcolor{xrgsSSShade}0.621 & \cellcolor{xrgsSSShade}0.559 \\
 & \cellcolor{xrgsControlShade}Control-CompGS & \cellcolor{xrgsControlShade}\textbf{0.595} & \cellcolor{xrgsControlShade}\textbf{0.465} & \cellcolor{xrgsControlShade}\textbf{0.463} & \cellcolor{xrgsControlShade}\textbf{0.607} & \cellcolor{xrgsControlShade}\textbf{0.569} & \cellcolor{xrgsControlShade}\textbf{0.488} & \cellcolor{xrgsControlShade}\textbf{0.441} & \cellcolor{xrgsControlShade}\textbf{0.557} & \cellcolor{xrgsControlShade}\textbf{0.601} & \cellcolor{xrgsControlShade}\textbf{0.532} \\
\cmidrule(lr){2-12}
 & ContextGS & 0.633 & 0.492 & 0.498 & 0.650 & 0.643 & 0.542 & 0.478 & 0.610 & 0.635 & 0.576 \\
 & \cellcolor{xrgsSSShade}SS-ContextGS & \cellcolor{xrgsSSShade}0.612 & \cellcolor{xrgsSSShade}0.480 & \cellcolor{xrgsSSShade}0.485 & \cellcolor{xrgsSSShade}0.623 & \cellcolor{xrgsSSShade}0.625 & \cellcolor{xrgsSSShade}0.526 & \cellcolor{xrgsSSShade}0.468 & \cellcolor{xrgsSSShade}0.585 & \cellcolor{xrgsSSShade}0.621 & \cellcolor{xrgsSSShade}0.558 \\
 & \cellcolor{xrgsControlShade}Control-ContextGS & \cellcolor{xrgsControlShade}\textbf{0.588} & \cellcolor{xrgsControlShade}\textbf{0.463} & \cellcolor{xrgsControlShade}\textbf{0.460} & \cellcolor{xrgsControlShade}\textbf{0.607} & \cellcolor{xrgsControlShade}\textbf{0.569} & \cellcolor{xrgsControlShade}\textbf{0.486} & \cellcolor{xrgsControlShade}\textbf{0.439} & \cellcolor{xrgsControlShade}\textbf{0.551} & \cellcolor{xrgsControlShade}\textbf{0.599} & \cellcolor{xrgsControlShade}\textbf{0.529} \\
\cmidrule(lr){2-12}
 & HAC{++} & 0.633 & 0.492 & 0.499 & 0.649 & 0.644 & 0.542 & 0.477 & 0.607 & 0.635 & 0.575 \\
 & \cellcolor{xrgsSSShade}SS-HAC{++} & \cellcolor{xrgsSSShade}0.614 & \cellcolor{xrgsSSShade}0.480 & \cellcolor{xrgsSSShade}0.485 & \cellcolor{xrgsSSShade}0.623 & \cellcolor{xrgsSSShade}0.626 & \cellcolor{xrgsSSShade}0.526 & \cellcolor{xrgsSSShade}0.468 & \cellcolor{xrgsSSShade}0.584 & \cellcolor{xrgsSSShade}0.621 & \cellcolor{xrgsSSShade}0.558 \\
 & \cellcolor{xrgsControlShade}Control-HAC{++} & \cellcolor{xrgsControlShade}\textbf{0.589} & \cellcolor{xrgsControlShade}\textbf{0.463} & \cellcolor{xrgsControlShade}\textbf{0.461} & \cellcolor{xrgsControlShade}\textbf{0.607} & \cellcolor{xrgsControlShade}\textbf{0.570} & \cellcolor{xrgsControlShade}\textbf{0.486} & \cellcolor{xrgsControlShade}\textbf{0.440} & \cellcolor{xrgsControlShade}\textbf{0.551} & \cellcolor{xrgsControlShade}\textbf{0.601} & \cellcolor{xrgsControlShade}\textbf{0.530} \\
\cmidrule(lr){2-12}
 & Octree-GS & 0.633 & 0.500 & 0.509 & 0.654 & 0.648 & 0.555 & 0.487 & 0.609 & 0.639 & 0.581 \\
 & \cellcolor{xrgsSSShade}SS-Octree-GS & \cellcolor{xrgsSSShade}0.613 & \cellcolor{xrgsSSShade}0.480 & \cellcolor{xrgsSSShade}0.484 & \cellcolor{xrgsSSShade}0.624 & \cellcolor{xrgsSSShade}0.625 & \cellcolor{xrgsSSShade}0.526 & \cellcolor{xrgsSSShade}0.467 & \cellcolor{xrgsSSShade}0.585 & \cellcolor{xrgsSSShade}0.621 & \cellcolor{xrgsSSShade}0.558 \\
 & \cellcolor{xrgsControlShade}Control-Octree-GS & \cellcolor{xrgsControlShade}\textbf{0.589} & \cellcolor{xrgsControlShade}\textbf{0.465} & \cellcolor{xrgsControlShade}\textbf{0.459} & \cellcolor{xrgsControlShade}\textbf{0.607} & \cellcolor{xrgsControlShade}\textbf{0.566} & \cellcolor{xrgsControlShade}\textbf{0.489} & \cellcolor{xrgsControlShade}\textbf{0.435} & \cellcolor{xrgsControlShade}\textbf{0.553} & \cellcolor{xrgsControlShade}\textbf{0.598} & \cellcolor{xrgsControlShade}\textbf{0.529} \\
\cmidrule(lr){2-12}
 & Scaffold-GS & 0.635 & 0.494 & 0.499 & 0.655 & 0.646 & 0.543 & 0.479 & 0.611 & 0.637 & 0.578 \\
 & \cellcolor{xrgsSSShade}SS-Scaffold-GS & \cellcolor{xrgsSSShade}0.614 & \cellcolor{xrgsSSShade}0.479 & \cellcolor{xrgsSSShade}0.484 & \cellcolor{xrgsSSShade}0.624 & \cellcolor{xrgsSSShade}0.626 & \cellcolor{xrgsSSShade}0.525 & \cellcolor{xrgsSSShade}0.467 & \cellcolor{xrgsSSShade}0.585 & \cellcolor{xrgsSSShade}0.621 & \cellcolor{xrgsSSShade}0.558 \\
 & \cellcolor{xrgsControlShade}Control-Scaffold-GS & \cellcolor{xrgsControlShade}\textbf{0.589} & \cellcolor{xrgsControlShade}\textbf{0.460} & \cellcolor{xrgsControlShade}\textbf{0.458} & \cellcolor{xrgsControlShade}\textbf{0.607} & \cellcolor{xrgsControlShade}\textbf{0.567} & \cellcolor{xrgsControlShade}\textbf{0.483} & \cellcolor{xrgsControlShade}\textbf{0.436} & \cellcolor{xrgsControlShade}\textbf{0.551} & \cellcolor{xrgsControlShade}\textbf{0.598} & \cellcolor{xrgsControlShade}\textbf{0.528} \\
\bottomrule
\end{tabular}%
}
\end{table*}

\begin{table*}[t]
\centering
\caption{Per-scene end-to-end PSNR and SSIM on NeRF Synthetic.
Each scene cell averages all test views and all 20 resolution-power settings.
Rows without prefix report the original backbone baseline without supersampling.
\textsc{SS} rows use the enhanced supersampling baseline.
\textsc{Control} rows add learned condition-injection adapters.}
\label{tab:supp:xrgs-e2e-quality-per-scene-nerf-synthetic}
\definecolor{xrgsSSShade}{RGB}{255,249,224}
\definecolor{xrgsControlShade}{RGB}{255,239,224}
\tiny
\setlength{\tabcolsep}{1.7pt}
\renewcommand{\arraystretch}{0.78}
\resizebox{0.75\textwidth}{!}{%
\begin{tabular}{llccccccccc}
\toprule
\multirow{2}{*}{Metric} & \multirow{2}{*}{Method} & \multicolumn{8}{c}{NeRF Synthetic} & \multirow{2}{*}{Avg} \\
\cmidrule(lr){3-10}
& & Chair & Drums & Ficus & Hotdog & Lego & Mater. & Mic & Ship & \\
\midrule
\multirow{19}{*}{PSNR$\uparrow$} & 3DGS & 19.86 & 19.03 & 23.43 & 22.95 & 20.47 & 22.58 & 25.14 & 22.20 & 21.96 \\
 & \cellcolor{xrgsSSShade}SS-3DGS & \cellcolor{xrgsSSShade}\textbf{21.54} & \cellcolor{xrgsSSShade}\textbf{20.32} & \cellcolor{xrgsSSShade}\textbf{24.94} & \cellcolor{xrgsSSShade}\textbf{24.46} & \cellcolor{xrgsSSShade}\textbf{21.90} & \cellcolor{xrgsSSShade}\textbf{23.96} & \cellcolor{xrgsSSShade}\textbf{26.51} & \cellcolor{xrgsSSShade}\textbf{23.26} & \cellcolor{xrgsSSShade}\textbf{23.36} \\
\cmidrule(lr){2-11}
 & Mip-Splatting & 21.29 & 20.17 & 24.66 & 24.21 & 21.71 & 23.85 & 26.24 & 23.25 & 23.17 \\
 & \cellcolor{xrgsSSShade}SS-Mip-Splatting & \cellcolor{xrgsSSShade}\textbf{21.54} & \cellcolor{xrgsSSShade}\textbf{20.32} & \cellcolor{xrgsSSShade}\textbf{24.94} & \cellcolor{xrgsSSShade}\textbf{24.47} & \cellcolor{xrgsSSShade}\textbf{21.90} & \cellcolor{xrgsSSShade}\textbf{23.97} & \cellcolor{xrgsSSShade}\textbf{26.51} & \cellcolor{xrgsSSShade}\textbf{23.28} & \cellcolor{xrgsSSShade}\textbf{23.37} \\
\cmidrule(lr){2-11}
 & CompGS & 19.91 & 18.68 & 21.01 & 22.99 & 20.43 & 21.71 & 25.11 & 21.94 & 21.47 \\
 & \cellcolor{xrgsSSShade}SS-CompGS & \cellcolor{xrgsSSShade}21.47 & \cellcolor{xrgsSSShade}19.76 & \cellcolor{xrgsSSShade}21.47 & \cellcolor{xrgsSSShade}24.35 & \cellcolor{xrgsSSShade}21.89 & \cellcolor{xrgsSSShade}22.32 & \cellcolor{xrgsSSShade}26.39 & \cellcolor{xrgsSSShade}23.17 & \cellcolor{xrgsSSShade}22.60 \\
 & \cellcolor{xrgsControlShade}Control-CompGS & \cellcolor{xrgsControlShade}\textbf{21.82} & \cellcolor{xrgsControlShade}\textbf{20.08} & \cellcolor{xrgsControlShade}\textbf{21.93} & \cellcolor{xrgsControlShade}\textbf{24.64} & \cellcolor{xrgsControlShade}\textbf{22.09} & \cellcolor{xrgsControlShade}\textbf{22.91} & \cellcolor{xrgsControlShade}\textbf{26.73} & \cellcolor{xrgsControlShade}\textbf{23.40} & \cellcolor{xrgsControlShade}\textbf{22.95} \\
\cmidrule(lr){2-11}
 & ContextGS & 19.96 & 19.08 & 23.45 & 23.17 & 20.52 & 22.83 & 25.24 & 22.03 & 22.03 \\
 & \cellcolor{xrgsSSShade}SS-ContextGS & \cellcolor{xrgsSSShade}21.54 & \cellcolor{xrgsSSShade}20.36 & \cellcolor{xrgsSSShade}24.91 & \cellcolor{xrgsSSShade}24.45 & \cellcolor{xrgsSSShade}21.90 & \cellcolor{xrgsSSShade}23.98 & \cellcolor{xrgsSSShade}26.53 & \cellcolor{xrgsSSShade}23.26 & \cellcolor{xrgsSSShade}23.37 \\
 & \cellcolor{xrgsControlShade}Control-ContextGS & \cellcolor{xrgsControlShade}\textbf{21.87} & \cellcolor{xrgsControlShade}\textbf{20.59} & \cellcolor{xrgsControlShade}\textbf{24.99} & \cellcolor{xrgsControlShade}\textbf{24.76} & \cellcolor{xrgsControlShade}\textbf{22.11} & \cellcolor{xrgsControlShade}\textbf{24.27} & \cellcolor{xrgsControlShade}\textbf{26.79} & \cellcolor{xrgsControlShade}\textbf{23.45} & \cellcolor{xrgsControlShade}\textbf{23.60} \\
\cmidrule(lr){2-11}
 & HAC{++} & 19.97 & 19.06 & 23.43 & 23.18 & 20.51 & 22.82 & 25.21 & 22.03 & 22.03 \\
 & \cellcolor{xrgsSSShade}SS-HAC{++} & \cellcolor{xrgsSSShade}21.53 & \cellcolor{xrgsSSShade}20.37 & \cellcolor{xrgsSSShade}24.92 & \cellcolor{xrgsSSShade}24.47 & \cellcolor{xrgsSSShade}21.90 & \cellcolor{xrgsSSShade}23.98 & \cellcolor{xrgsSSShade}26.52 & \cellcolor{xrgsSSShade}23.23 & \cellcolor{xrgsSSShade}23.36 \\
 & \cellcolor{xrgsControlShade}Control-HAC{++} & \cellcolor{xrgsControlShade}\textbf{21.87} & \cellcolor{xrgsControlShade}\textbf{20.59} & \cellcolor{xrgsControlShade}\textbf{24.98} & \cellcolor{xrgsControlShade}\textbf{24.76} & \cellcolor{xrgsControlShade}\textbf{22.12} & \cellcolor{xrgsControlShade}\textbf{24.26} & \cellcolor{xrgsControlShade}\textbf{26.77} & \cellcolor{xrgsControlShade}\textbf{23.42} & \cellcolor{xrgsControlShade}\textbf{23.60} \\
\cmidrule(lr){2-11}
 & Octree-GS & 19.85 & 19.01 & 23.47 & 23.03 & 20.38 & 22.79 & 25.29 & 22.00 & 21.98 \\
 & \cellcolor{xrgsSSShade}SS-Octree-GS & \cellcolor{xrgsSSShade}21.52 & \cellcolor{xrgsSSShade}20.22 & \cellcolor{xrgsSSShade}24.65 & \cellcolor{xrgsSSShade}24.41 & \cellcolor{xrgsSSShade}21.89 & \cellcolor{xrgsSSShade}23.95 & \cellcolor{xrgsSSShade}26.46 & \cellcolor{xrgsSSShade}23.24 & \cellcolor{xrgsSSShade}23.29 \\
 & \cellcolor{xrgsControlShade}Control-Octree-GS & \cellcolor{xrgsControlShade}\textbf{21.85} & \cellcolor{xrgsControlShade}\textbf{20.39} & \cellcolor{xrgsControlShade}\textbf{24.71} & \cellcolor{xrgsControlShade}\textbf{24.74} & \cellcolor{xrgsControlShade}\textbf{22.13} & \cellcolor{xrgsControlShade}\textbf{24.22} & \cellcolor{xrgsControlShade}\textbf{26.71} & \cellcolor{xrgsControlShade}\textbf{23.43} & \cellcolor{xrgsControlShade}\textbf{23.52} \\
\cmidrule(lr){2-11}
 & Scaffold-GS & 19.83 & 18.97 & 23.37 & 22.97 & 20.39 & 22.76 & 25.14 & 22.04 & 21.94 \\
 & \cellcolor{xrgsSSShade}SS-Scaffold-GS & \cellcolor{xrgsSSShade}21.52 & \cellcolor{xrgsSSShade}20.33 & \cellcolor{xrgsSSShade}24.89 & \cellcolor{xrgsSSShade}24.44 & \cellcolor{xrgsSSShade}21.89 & \cellcolor{xrgsSSShade}23.95 & \cellcolor{xrgsSSShade}26.51 & \cellcolor{xrgsSSShade}23.26 & \cellcolor{xrgsSSShade}23.35 \\
 & \cellcolor{xrgsControlShade}Control-Scaffold-GS & \cellcolor{xrgsControlShade}\textbf{21.86} & \cellcolor{xrgsControlShade}\textbf{20.55} & \cellcolor{xrgsControlShade}\textbf{24.98} & \cellcolor{xrgsControlShade}\textbf{24.77} & \cellcolor{xrgsControlShade}\textbf{22.12} & \cellcolor{xrgsControlShade}\textbf{24.21} & \cellcolor{xrgsControlShade}\textbf{26.80} & \cellcolor{xrgsControlShade}\textbf{23.45} & \cellcolor{xrgsControlShade}\textbf{23.59} \\
\cmidrule(lr){1-11}
\multirow{19}{*}{SSIM$\uparrow$} & 3DGS & 0.850 & 0.801 & 0.875 & 0.897 & 0.806 & 0.849 & 0.893 & 0.745 & 0.840 \\
 & \cellcolor{xrgsSSShade}SS-3DGS & \cellcolor{xrgsSSShade}\textbf{0.879} & \cellcolor{xrgsSSShade}\textbf{0.844} & \cellcolor{xrgsSSShade}\textbf{0.902} & \cellcolor{xrgsSSShade}\textbf{0.916} & \cellcolor{xrgsSSShade}\textbf{0.842} & \cellcolor{xrgsSSShade}\textbf{0.880} & \cellcolor{xrgsSSShade}\textbf{0.921} & \cellcolor{xrgsSSShade}\textbf{0.772} & \cellcolor{xrgsSSShade}\textbf{0.869} \\
\cmidrule(lr){2-11}
 & Mip-Splatting & 0.871 & 0.833 & 0.895 & 0.911 & 0.832 & 0.873 & 0.912 & 0.769 & 0.862 \\
 & \cellcolor{xrgsSSShade}SS-Mip-Splatting & \cellcolor{xrgsSSShade}\textbf{0.879} & \cellcolor{xrgsSSShade}\textbf{0.844} & \cellcolor{xrgsSSShade}\textbf{0.902} & \cellcolor{xrgsSSShade}\textbf{0.916} & \cellcolor{xrgsSSShade}\textbf{0.842} & \cellcolor{xrgsSSShade}\textbf{0.880} & \cellcolor{xrgsSSShade}\textbf{0.921} & \cellcolor{xrgsSSShade}\textbf{0.773} & \cellcolor{xrgsSSShade}\textbf{0.870} \\
\cmidrule(lr){2-11}
 & CompGS & 0.850 & 0.799 & 0.864 & 0.896 & 0.807 & 0.840 & 0.894 & 0.742 & 0.836 \\
 & \cellcolor{xrgsSSShade}SS-CompGS & \cellcolor{xrgsSSShade}0.878 & \cellcolor{xrgsSSShade}0.834 & \cellcolor{xrgsSSShade}\textbf{0.877} & \cellcolor{xrgsSSShade}0.914 & \cellcolor{xrgsSSShade}0.841 & \cellcolor{xrgsSSShade}0.859 & \cellcolor{xrgsSSShade}0.919 & \cellcolor{xrgsSSShade}0.770 & \cellcolor{xrgsSSShade}0.861 \\
 & \cellcolor{xrgsControlShade}Control-CompGS & \cellcolor{xrgsControlShade}\textbf{0.882} & \cellcolor{xrgsControlShade}\textbf{0.838} & \cellcolor{xrgsControlShade}0.876 & \cellcolor{xrgsControlShade}\textbf{0.915} & \cellcolor{xrgsControlShade}\textbf{0.846} & \cellcolor{xrgsControlShade}\textbf{0.862} & \cellcolor{xrgsControlShade}\textbf{0.924} & \cellcolor{xrgsControlShade}\textbf{0.771} & \cellcolor{xrgsControlShade}\textbf{0.864} \\
\cmidrule(lr){2-11}
 & ContextGS & 0.852 & 0.802 & 0.875 & 0.899 & 0.806 & 0.852 & 0.896 & 0.741 & 0.840 \\
 & \cellcolor{xrgsSSShade}SS-ContextGS & \cellcolor{xrgsSSShade}0.879 & \cellcolor{xrgsSSShade}0.843 & \cellcolor{xrgsSSShade}0.902 & \cellcolor{xrgsSSShade}0.916 & \cellcolor{xrgsSSShade}0.841 & \cellcolor{xrgsSSShade}0.880 & \cellcolor{xrgsSSShade}0.921 & \cellcolor{xrgsSSShade}0.770 & \cellcolor{xrgsSSShade}0.869 \\
 & \cellcolor{xrgsControlShade}Control-ContextGS & \cellcolor{xrgsControlShade}\textbf{0.882} & \cellcolor{xrgsControlShade}\textbf{0.847} & \cellcolor{xrgsControlShade}\textbf{0.902} & \cellcolor{xrgsControlShade}\textbf{0.916} & \cellcolor{xrgsControlShade}\textbf{0.846} & \cellcolor{xrgsControlShade}\textbf{0.885} & \cellcolor{xrgsControlShade}\textbf{0.923} & \cellcolor{xrgsControlShade}\textbf{0.771} & \cellcolor{xrgsControlShade}\textbf{0.872} \\
\cmidrule(lr){2-11}
 & HAC{++} & 0.852 & 0.802 & 0.875 & 0.899 & 0.806 & 0.852 & 0.895 & 0.743 & 0.841 \\
 & \cellcolor{xrgsSSShade}SS-HAC{++} & \cellcolor{xrgsSSShade}0.879 & \cellcolor{xrgsSSShade}0.843 & \cellcolor{xrgsSSShade}0.902 & \cellcolor{xrgsSSShade}0.915 & \cellcolor{xrgsSSShade}0.841 & \cellcolor{xrgsSSShade}0.880 & \cellcolor{xrgsSSShade}0.921 & \cellcolor{xrgsSSShade}0.771 & \cellcolor{xrgsSSShade}0.869 \\
 & \cellcolor{xrgsControlShade}Control-HAC{++} & \cellcolor{xrgsControlShade}\textbf{0.883} & \cellcolor{xrgsControlShade}\textbf{0.847} & \cellcolor{xrgsControlShade}\textbf{0.902} & \cellcolor{xrgsControlShade}\textbf{0.916} & \cellcolor{xrgsControlShade}\textbf{0.846} & \cellcolor{xrgsControlShade}\textbf{0.885} & \cellcolor{xrgsControlShade}\textbf{0.923} & \cellcolor{xrgsControlShade}\textbf{0.772} & \cellcolor{xrgsControlShade}\textbf{0.872} \\
\cmidrule(lr){2-11}
 & Octree-GS & 0.850 & 0.805 & 0.879 & 0.897 & 0.804 & 0.854 & 0.899 & 0.742 & 0.841 \\
 & \cellcolor{xrgsSSShade}SS-Octree-GS & \cellcolor{xrgsSSShade}0.878 & \cellcolor{xrgsSSShade}0.841 & \cellcolor{xrgsSSShade}0.899 & \cellcolor{xrgsSSShade}0.915 & \cellcolor{xrgsSSShade}0.841 & \cellcolor{xrgsSSShade}0.879 & \cellcolor{xrgsSSShade}0.920 & \cellcolor{xrgsSSShade}0.770 & \cellcolor{xrgsSSShade}0.868 \\
 & \cellcolor{xrgsControlShade}Control-Octree-GS & \cellcolor{xrgsControlShade}\textbf{0.882} & \cellcolor{xrgsControlShade}\textbf{0.844} & \cellcolor{xrgsControlShade}\textbf{0.899} & \cellcolor{xrgsControlShade}\textbf{0.915} & \cellcolor{xrgsControlShade}\textbf{0.846} & \cellcolor{xrgsControlShade}\textbf{0.884} & \cellcolor{xrgsControlShade}\textbf{0.921} & \cellcolor{xrgsControlShade}\textbf{0.771} & \cellcolor{xrgsControlShade}\textbf{0.870} \\
\cmidrule(lr){2-11}
 & Scaffold-GS & 0.850 & 0.799 & 0.874 & 0.897 & 0.805 & 0.851 & 0.893 & 0.741 & 0.839 \\
 & \cellcolor{xrgsSSShade}SS-Scaffold-GS & \cellcolor{xrgsSSShade}0.878 & \cellcolor{xrgsSSShade}0.843 & \cellcolor{xrgsSSShade}0.901 & \cellcolor{xrgsSSShade}0.915 & \cellcolor{xrgsSSShade}0.841 & \cellcolor{xrgsSSShade}0.880 & \cellcolor{xrgsSSShade}0.920 & \cellcolor{xrgsSSShade}0.770 & \cellcolor{xrgsSSShade}0.869 \\
 & \cellcolor{xrgsControlShade}Control-Scaffold-GS & \cellcolor{xrgsControlShade}\textbf{0.882} & \cellcolor{xrgsControlShade}\textbf{0.846} & \cellcolor{xrgsControlShade}\textbf{0.902} & \cellcolor{xrgsControlShade}\textbf{0.915} & \cellcolor{xrgsControlShade}\textbf{0.846} & \cellcolor{xrgsControlShade}\textbf{0.884} & \cellcolor{xrgsControlShade}\textbf{0.923} & \cellcolor{xrgsControlShade}\textbf{0.771} & \cellcolor{xrgsControlShade}\textbf{0.871} \\
\bottomrule
\end{tabular}%
}
\end{table*}

\begin{table*}[p]
\centering
\caption{Per-scene end-to-end LPIPS on NeRF Synthetic.
Each scene cell averages all test views and all 20 resolution-power settings.
Rows without prefix report the original backbone baseline without supersampling.
\textsc{SS} rows use the enhanced supersampling baseline.
\textsc{Control} rows add learned condition-injection adapters.}
\label{tab:supp:xrgs-e2e-quality-per-scene-nerf-synthetic-lpips}
\definecolor{xrgsSSShade}{RGB}{255,249,224}
\definecolor{xrgsControlShade}{RGB}{255,239,224}
\tiny
\setlength{\tabcolsep}{1.7pt}
\renewcommand{\arraystretch}{0.78}
\resizebox{0.75\textwidth}{!}{%
\begin{tabular}{llccccccccc}
\toprule
\multirow{2}{*}{Metric} & \multirow{2}{*}{Method} & \multicolumn{8}{c}{NeRF Synthetic} & \multirow{2}{*}{Avg} \\
\cmidrule(lr){3-10}
& & Chair & Drums & Ficus & Hotdog & Lego & Mater. & Mic & Ship & \\
\midrule
\multirow{19}{*}{LPIPS$\downarrow$} & 3DGS & 0.139 & 0.188 & 0.116 & 0.130 & 0.203 & 0.157 & 0.110 & 0.283 & 0.166 \\
 & \cellcolor{xrgsSSShade}SS-3DGS & \cellcolor{xrgsSSShade}\textbf{0.118} & \cellcolor{xrgsSSShade}\textbf{0.161} & \cellcolor{xrgsSSShade}\textbf{0.097} & \cellcolor{xrgsSSShade}\textbf{0.114} & \cellcolor{xrgsSSShade}\textbf{0.175} & \cellcolor{xrgsSSShade}\textbf{0.138} & \cellcolor{xrgsSSShade}\textbf{0.096} & \cellcolor{xrgsSSShade}\textbf{0.265} & \cellcolor{xrgsSSShade}\textbf{0.146} \\
\cmidrule(lr){2-11}
 & Mip-Splatting & 0.124 & 0.169 & 0.102 & 0.120 & 0.184 & 0.143 & 0.101 & 0.271 & 0.152 \\
 & \cellcolor{xrgsSSShade}SS-Mip-Splatting & \cellcolor{xrgsSSShade}\textbf{0.118} & \cellcolor{xrgsSSShade}\textbf{0.161} & \cellcolor{xrgsSSShade}\textbf{0.097} & \cellcolor{xrgsSSShade}\textbf{0.114} & \cellcolor{xrgsSSShade}\textbf{0.175} & \cellcolor{xrgsSSShade}\textbf{0.138} & \cellcolor{xrgsSSShade}\textbf{0.096} & \cellcolor{xrgsSSShade}\textbf{0.264} & \cellcolor{xrgsSSShade}\textbf{0.145} \\
\cmidrule(lr){2-11}
 & CompGS & 0.141 & 0.196 & 0.135 & 0.131 & 0.202 & 0.179 & 0.111 & 0.285 & 0.173 \\
 & \cellcolor{xrgsSSShade}SS-CompGS & \cellcolor{xrgsSSShade}0.119 & \cellcolor{xrgsSSShade}0.174 & \cellcolor{xrgsSSShade}0.127 & \cellcolor{xrgsSSShade}0.117 & \cellcolor{xrgsSSShade}0.175 & \cellcolor{xrgsSSShade}0.168 & \cellcolor{xrgsSSShade}0.098 & \cellcolor{xrgsSSShade}0.269 & \cellcolor{xrgsSSShade}0.156 \\
 & \cellcolor{xrgsControlShade}Control-CompGS & \cellcolor{xrgsControlShade}\textbf{0.110} & \cellcolor{xrgsControlShade}\textbf{0.166} & \cellcolor{xrgsControlShade}\textbf{0.123} & \cellcolor{xrgsControlShade}\textbf{0.108} & \cellcolor{xrgsControlShade}\textbf{0.164} & \cellcolor{xrgsControlShade}\textbf{0.162} & \cellcolor{xrgsControlShade}\textbf{0.090} & \cellcolor{xrgsControlShade}\textbf{0.251} & \cellcolor{xrgsControlShade}\textbf{0.147} \\
\cmidrule(lr){2-11}
 & ContextGS & 0.139 & 0.188 & 0.116 & 0.128 & 0.203 & 0.155 & 0.109 & 0.284 & 0.165 \\
 & \cellcolor{xrgsSSShade}SS-ContextGS & \cellcolor{xrgsSSShade}0.118 & \cellcolor{xrgsSSShade}0.163 & \cellcolor{xrgsSSShade}0.098 & \cellcolor{xrgsSSShade}0.115 & \cellcolor{xrgsSSShade}0.175 & \cellcolor{xrgsSSShade}0.139 & \cellcolor{xrgsSSShade}0.096 & \cellcolor{xrgsSSShade}0.267 & \cellcolor{xrgsSSShade}0.146 \\
 & \cellcolor{xrgsControlShade}Control-ContextGS & \cellcolor{xrgsControlShade}\textbf{0.109} & \cellcolor{xrgsControlShade}\textbf{0.153} & \cellcolor{xrgsControlShade}\textbf{0.094} & \cellcolor{xrgsControlShade}\textbf{0.106} & \cellcolor{xrgsControlShade}\textbf{0.164} & \cellcolor{xrgsControlShade}\textbf{0.133} & \cellcolor{xrgsControlShade}\textbf{0.088} & \cellcolor{xrgsControlShade}\textbf{0.246} & \cellcolor{xrgsControlShade}\textbf{0.137} \\
\cmidrule(lr){2-11}
 & HAC{++} & 0.140 & 0.187 & 0.116 & 0.128 & 0.203 & 0.155 & 0.109 & 0.284 & 0.165 \\
 & \cellcolor{xrgsSSShade}SS-HAC{++} & \cellcolor{xrgsSSShade}0.118 & \cellcolor{xrgsSSShade}0.163 & \cellcolor{xrgsSSShade}0.097 & \cellcolor{xrgsSSShade}0.115 & \cellcolor{xrgsSSShade}0.175 & \cellcolor{xrgsSSShade}0.140 & \cellcolor{xrgsSSShade}0.096 & \cellcolor{xrgsSSShade}0.267 & \cellcolor{xrgsSSShade}0.146 \\
 & \cellcolor{xrgsControlShade}Control-HAC{++} & \cellcolor{xrgsControlShade}\textbf{0.108} & \cellcolor{xrgsControlShade}\textbf{0.153} & \cellcolor{xrgsControlShade}\textbf{0.093} & \cellcolor{xrgsControlShade}\textbf{0.106} & \cellcolor{xrgsControlShade}\textbf{0.163} & \cellcolor{xrgsControlShade}\textbf{0.133} & \cellcolor{xrgsControlShade}\textbf{0.088} & \cellcolor{xrgsControlShade}\textbf{0.246} & \cellcolor{xrgsControlShade}\textbf{0.136} \\
\cmidrule(lr){2-11}
 & Octree-GS & 0.144 & 0.190 & 0.115 & 0.132 & 0.206 & 0.159 & 0.111 & 0.284 & 0.168 \\
 & \cellcolor{xrgsSSShade}SS-Octree-GS & \cellcolor{xrgsSSShade}0.118 & \cellcolor{xrgsSSShade}0.166 & \cellcolor{xrgsSSShade}0.100 & \cellcolor{xrgsSSShade}0.116 & \cellcolor{xrgsSSShade}0.175 & \cellcolor{xrgsSSShade}0.140 & \cellcolor{xrgsSSShade}0.098 & \cellcolor{xrgsSSShade}0.267 & \cellcolor{xrgsSSShade}0.147 \\
 & \cellcolor{xrgsControlShade}Control-Octree-GS & \cellcolor{xrgsControlShade}\textbf{0.109} & \cellcolor{xrgsControlShade}\textbf{0.158} & \cellcolor{xrgsControlShade}\textbf{0.097} & \cellcolor{xrgsControlShade}\textbf{0.106} & \cellcolor{xrgsControlShade}\textbf{0.163} & \cellcolor{xrgsControlShade}\textbf{0.134} & \cellcolor{xrgsControlShade}\textbf{0.092} & \cellcolor{xrgsControlShade}\textbf{0.246} & \cellcolor{xrgsControlShade}\textbf{0.138} \\
\cmidrule(lr){2-11}
 & Scaffold-GS & 0.141 & 0.191 & 0.117 & 0.130 & 0.205 & 0.157 & 0.110 & 0.284 & 0.167 \\
 & \cellcolor{xrgsSSShade}SS-Scaffold-GS & \cellcolor{xrgsSSShade}0.118 & \cellcolor{xrgsSSShade}0.164 & \cellcolor{xrgsSSShade}0.098 & \cellcolor{xrgsSSShade}0.115 & \cellcolor{xrgsSSShade}0.175 & \cellcolor{xrgsSSShade}0.140 & \cellcolor{xrgsSSShade}0.096 & \cellcolor{xrgsSSShade}0.267 & \cellcolor{xrgsSSShade}0.147 \\
 & \cellcolor{xrgsControlShade}Control-Scaffold-GS & \cellcolor{xrgsControlShade}\textbf{0.108} & \cellcolor{xrgsControlShade}\textbf{0.154} & \cellcolor{xrgsControlShade}\textbf{0.094} & \cellcolor{xrgsControlShade}\textbf{0.106} & \cellcolor{xrgsControlShade}\textbf{0.163} & \cellcolor{xrgsControlShade}\textbf{0.134} & \cellcolor{xrgsControlShade}\textbf{0.088} & \cellcolor{xrgsControlShade}\textbf{0.245} & \cellcolor{xrgsControlShade}\textbf{0.137} \\
\bottomrule
\end{tabular}%
}
\end{table*}

\end{document}